\documentclass{article}
\usepackage{arxiv}
\usepackage[utf8]{inputenc}
\usepackage[T1]{fontenc}   
\usepackage{hyperref}       
\usepackage{url}            
\usepackage{booktabs}       
\usepackage{amsmath}        
\usepackage{amsfonts}       
\usepackage{nicefrac}       
\usepackage{microtype}     
\usepackage{graphicx}
\usepackage{natbib}
\usepackage{doi}
\usepackage{bm}
\usepackage{multirow}
\usepackage{caption}
\usepackage{subcaption}
\usepackage{float}          
\graphicspath{{./figs/}}    
\title{When Does Frequency Decomposition Benefit Physics-Informed Neural Networks? A Preliminary Ablation Study}
\date{August 14, 2026}
\author{ Shubham Rai \thanks{Corresponding author.} \\
	\texttt{shubham.rai@bibha.ai} \\
	\textbf{bibha.ai} \\}
\renewcommand{\shorttitle}{Frequency Decomposition in PINNs}
\hypersetup{
	pdftitle={When Does Frequency Decomposition Benefit Physics-Informed Neural Networks? A Preliminary Ablation Study},
	pdfsubject={cs.LG, math.NA},
	pdfauthor={Shubham Rai},
	pdfkeywords={Physics-Informed Neural Networks, spectral bias, frequency decomposition, ablation study, PDE},}
\begin{document}
	\maketitle
	\hypersetup{colorlinks=true,linkcolor=blue,citecolor=blue,urlcolor=blue}
	
	\begin{abstract}
		Partial differential equations (PDEs) often have high-frequency and multi-scale features that neural networks struggle to approximate. Physics-Informed Neural Networks (PINNs) build the governing equations directly into training, but suffer from \emph{spectral bias}: they learn low-frequency components faster than high-frequency ones. Techniques such as Fourier feature embeddings and sinusoidal activations address this, but most studies assume they help across the board without checking which spectral regimes actually benefit. We introduce a dual-branch, spectrally-gated architecture (DBSG-PINN) that splits low- and high-frequency components into separate subnetworks joined by an adaptive gate, and use it to run a partially controlled ablation of frequency decomposition and spectral routing. We test this on five one-dimensional benchmark PDEs, ranging from smooth, single-scale problems to oscillatory, multi-scale ones. Frequency decomposition helps most on the spectrally complex benchmarks, cutting relative $L_2$ error by up to $59.2\%$ on a multimodal wave problem, but gives little benefit on smoother PDEs. On one benchmark (1D Wave), it performs substantially worse than a simpler fixed-combination variant. The gate's benefit scales with how spectrally rich the target solution is: the full model's advantage over the ablations is largest on multi-scale benchmarks and smallest (or negative) on single-scale ones, consistent with the gate exploiting frequency structure rather than acting as noise, though we do not directly visualize or quantify its spatial activations in this study. All results come from a single training seed across five 1D benchmarks, so we present this as an exploratory study meant to raise questions rather than answer them, and outline the additional seeds and benchmarks needed to test whether the pattern holds.
	\end{abstract}
	\keywords{Physics Informed Neural Networks \and Spectral Bias \and Frequency Decomposition \and Ablation Study \and Partial Differential Equations}
	\section{Introduction}
	
	Partial differential equations (PDEs) are the mathematical backbone for modelling physical, biological, and engineering systems, including fluid flow, heat transfer, wave propagation, diffusion, reaction kinetics, and electromagnetic phenomena. Most of these equations come from conservation laws and first principles, yet closed-form solutions exist only for a narrow class of idealized cases. Once nonlinear dynamics, complex geometries, heterogeneous materials, or high-dimensional parameter spaces enter the picture, an analytical solution is usually out of reach. That is why numerical methods -- finite differences, finite elements, spectral solvers -- carry most of the load in practice. They work, but not cheaply: mesh generation, repeated simulation runs, and heavy compute are the price, and that price gets steeper for inverse problems, uncertainty quantification, and parameter estimation.
	
	Scientific Machine Learning (SciML) takes a different route by folding the governing physical laws directly into a data-driven model, rather than relying on observed data alone -- the network is pushed toward physically consistent solutions without needing as much labelled data. Physics-Informed Neural Networks (PINNs) \citep{raissi2019physics} are probably the most influential framework to come out of this line of work for forward and inverse PDE problems: governing equations, boundary conditions, and initial conditions all get folded into the training objective through automatic differentiation, so whatever the network learns has to satisfy the underlying physics. PINNs have since shown up in computational fluid dynamics, inverse modelling, biomedical engineering, and geophysical simulation, and are more or less a default tool in SciML at this point~\citep{karniadakis2021physics,cuomo2022scientific,luo2025physics}.
	
	That said, conventional PINNs are not without problems: optimization pathologies, imbalance between competing loss terms, and poor convergence on stiff or highly nonlinear PDEs have all been reported~\citep{krishnapriyan2021characterizing,wang2021understanding,wang2022and}. \emph{Spectral bias} is probably the most studied of these -- networks trained with gradient descent pick up low-frequency components of a target function well before high-frequency ones. \citet{rahaman2019spectral} first showed this in deep networks, and the Neural Tangent Kernel (NTK) later gave a reason why: low-frequency functions line up with the dominant kernel eigenmodes and so converge faster~\citep{jacot2018neural}. In practice this means PINNs often struggle with localized discontinuities, sharp gradients, oscillatory behaviour, or several interacting spatial or temporal scales -- wave propagation~\citep{moseley2020solving}, reaction--diffusion systems, and other high-frequency or multi-scale PDEs are where this tends to bite hardest~\citep{mustajab2024physics}.
	
	A number of frequency-aware fixes have been proposed for spectral bias, and they roughly split into four camps. One enriches the input itself: Fourier feature embeddings help a network approximate high-frequency functions~\citep{tancik2020fourier}, and the periodic activations in Sinusoidal Representation Networks (SIREN) do something similar for oscillatory signals and their derivatives~\citep{sitzmann2020implicit}. A second camp works on the optimization process directly, through adaptive activation functions, adaptive loss balancing, and learning-rate annealing~\citep{jagtap2020adaptive,wang2021understanding}. A third tackles scalability with domain decomposition -- Conservative PINNs (cPINNs), Extended PINNs (XPINNs), and Finite Basis PINNs (FBPINNs) all split the domain into subregions to make optimization and local accuracy more tractable~\citep{jagtap2020conservative,jagtap2020extended,moseley2023finite}. And a fourth leans on gating mechanisms or deeper residual architectures for scalability and stability~\citep{stiller2020large,he2016deep}. Between them, these advances have pushed the range of problems PINNs can handle considerably further.
	
	Despite this progress, most existing work focuses on building more sophisticated architectures to fight spectral bias, and pays less attention to \emph{when} these mechanisms actually help. Most frequency-aware methods assume that richer spectral representations are broadly useful across PDE types. But physical systems differ a lot in their spectral character: some PDEs are smooth and low-frequency, while others involve localized oscillations, several interacting frequency modes, or complex multi-scale dynamics. This raises a natural question: does explicit frequency decomposition consistently improve accuracy, or does its value depend on the spectral character of the PDE? As far as we know, few studies have directly compared frequency-decomposed architectures against matched-capacity ablations across PDEs of varying spectral complexity. That gap motivates the exploratory study we present here.
	
	To study this question, we build the \textbf{Dual-Branch Spectral-Gated Physics-Informed Neural Network (DBSG-PINN)}, a controlled experimental framework for examining the role of frequency decomposition in PINNs. It consists of two subnetworks that each specialize in a different frequency regime: a low-frequency branch using hyperbolic tangent activations to capture smooth solution components, and a high-frequency branch using sinusoidal activations to model oscillatory behaviour. Their outputs are combined by an adaptive gate that learns spatially varying mixing weights, letting the model balance the two representations according to local solution behaviour. Each component can be removed without changing the optimization procedure or (roughly) the network's capacity, which is what makes ablation studies possible: we can isolate the individual contributions of frequency decomposition and adaptive spectral routing.
	
	We test the framework on five benchmark PDEs spanning a range of spectral complexity: the Burgers equation, Reaction--Diffusion equation, Allen--Cahn equation, Wave equation, and a Multimodal Wave equation, ranging from smooth, single-scale solutions to oscillatory, multi-scale dynamics. For each benchmark, we compare the full DBSG-PINN against three ablation variants, made by removing the high-frequency branch, the low-frequency branch, or the adaptive gate on its own, while keeping network capacity and optimization settings matched as closely as the architecture allows. This is meant to reduce, though not remove, the risk that any performance difference comes from model complexity or training strategy rather than frequency decomposition itself; we discuss what this control does and does not achieve in Section~\ref{sec:discussion}.
	
	Across the five benchmarks, frequency decomposition clearly pays off on the spectrally complex ones: up to 59.2\% lower relative $L_2$ error on Multimodal Wave, plus better spectral recovery there more broadly. On the smoother, single-scale problems the benefit shrinks or disappears -- a simpler fixed-combination variant edges out the full gated model on Burgers, and beats it outright on 1D Wave. We want to be careful about what this does and does not show: these are observations from one seed across five benchmarks, not a general claim about frequency-decomposed PINNs as a class.
	
	Beyond accuracy, we also looked at what the learned gate contributes across benchmarks. Its benefit is largest on the benchmarks with the richest spectral content and smallest (or negative) on the simplest ones -- a pattern consistent with the gate exploiting frequency structure rather than acting as noise, though we infer this from benchmark-level comparisons rather than from a direct visualization of the gate's spatial activations. That said, this comes from one trained model per benchmark, and it needs independent verification -- ideally including an explicit visualization of $g(x,t)$ against local spectral content -- before we can call it a general property of the architecture.
	
	\paragraph{Our Contributions}
	\begin{itemize}
		\item We build DBSG-PINN, a dual-branch architecture combining low- and high-frequency subnetworks through an adaptive spectral gate.
		\item We design an ablation framework that removes the low-frequency branch, high-frequency branch, or gate individually, keeping training settings matched exactly and per-branch capacity matched only approximately -- depth and width are not identical between the low- and high-frequency branches on most benchmarks (Section~\ref{sec:ablation-variants}).
		\item We report an initial comparison across five 1D benchmark PDEs of varying spectral complexity -- meant to surface a pattern worth investigating further, not to establish a general rule.
		\item We show a preliminary, benchmark-level pattern in which the gate's contribution scales with the spectral complexity of the target solution, though we do not directly visualize or quantify its spatial activations here.
	\end{itemize}
	\section{Methodology}
	
	\subsection{Physics-Informed Neural Networks}
	
	PINNs belong to the broader class of Scientific Machine Learning (SciML) models that fold the governing physical laws directly into the training objective rather than relying only on labelled data \citep{raissi2019physics}. Concretely, this means minimizing the residual of the governing PDE alongside the initial and boundary conditions, pushing the network toward solutions consistent with the underlying physics rather than merely fitting data points.
	
	Consider a general nonlinear PDE
	
	\begin{equation}
		\frac{\partial u(\mathbf{x},t)}{\partial t}+\mathcal{N}[u(\mathbf{x},t)]=0,
		\label{eq:pde}
	\end{equation}
	
	where $u(\mathbf{x},t)$ is the unknown solution, $\mathbf{x}$ is the spatial coordinate, and $\mathcal{N}(\cdot)$ is a nonlinear differential operator.
	
	The solution is approximated by a neural network
	
	\begin{equation}
		u(\mathbf{x},t)\approx u_\theta(\mathbf{x},t),
		\label{eq:network}
	\end{equation}
	
	where $\theta$ are the trainable network parameters.
	
	Using automatic differentiation, the PDE residual is computed as
	
	\begin{equation}
		r_\theta(\mathbf{x},t)=\frac{\partial u_\theta(\mathbf{x},t)}{\partial t}+
		\mathcal{N}[u_\theta(\mathbf{x},t)].
		\label{eq:residual}
	\end{equation}
	
	All the spatial and temporal derivatives needed to evaluate $r_\theta(\mathbf{x},t)$ come from automatic differentiation~\citep{baydin2018automatic}, which computes exact derivatives through the network graph instead of relying on finite-difference approximations.
	
	The overall PINN objective is a weighted sum of the governing-equation residual and the initial and boundary condition losses,
	
	\begin{equation}
		\mathcal{L}=\lambda_r\mathcal{L}_{PDE}+\lambda_{IC}\mathcal{L}_{IC}+\lambda_{BC}\mathcal{L}_{BC},
		\label{eq:loss}
	\end{equation}
	
	The scalar weights $\lambda_r$, $\lambda_{IC}$, and $\lambda_{BC}$ balance these terms. Poorly chosen weights are a known cause of the gradient imbalance and stiffness problems reported in PINN training, which is why prior work has explored adaptive or self-weighting schemes~\citep{mcclenny2020self,wang2022and}. Here we fix the weights per benchmark (Table~\ref{tab:hyperparameters}) so that all ablation variants share the same loss landscape; we leave adaptive weighting for future work.
	
	where
	
	\begin{equation}
		\mathcal{L}_{PDE}=\frac{1}{N_r}\sum_{i=1}^{N_r}\left|r_\theta(\mathbf{x}_i,t_i)\right|^2,
		\label{eq:pdeloss}
	\end{equation}
	
	and
	
	\begin{equation}
		\mathcal{L}_{IC}=\frac{1}{N_{IC}}\sum_{i=1}^{N_{IC}}\left|u_\theta(\mathbf{x}_i,0)-u_0(\mathbf{x}_i)
		\right|^2,
	\end{equation}
	
	Equation~\ref{eq:pde} is written as a first-order-in-time PDE for generality, in which case the value term above is the only initial condition needed. For the second-order-in-time (wave-type) benchmarks in Section~\ref{sec:results} (Multimodal Wave and 1D Wave), the governing equation also specifies an initial velocity $u_t(\mathbf{x},0)=v_0(\mathbf{x})$ alongside the initial value $u(\mathbf{x},0)=u_0(\mathbf{x})$; we enforce both terms in $\mathcal{L}_{IC}$,
	\begin{equation}
		\mathcal{L}_{IC}=\frac{1}{N_{IC}}\sum_{i=1}^{N_{IC}}\left[\left|u_\theta(\mathbf{x}_i,0)-u_0(\mathbf{x}_i)\right|^2+\left|\partial_t u_\theta(\mathbf{x}_i,0)-v_0(\mathbf{x}_i)\right|^2\right],
	\end{equation}
	with $\partial_t u_\theta(\mathbf{x}_i,0)$ again obtained via automatic differentiation and both terms weighted equally within $\mathcal{L}_{IC}$. For the first-order-in-time benchmarks (Burgers, Allen--Cahn, Reaction--Diffusion), only the value term applies, as in the equation above.
	
	\begin{equation}
		\mathcal{L}_{BC}=\frac{1}{N_{BC}}\sum_{i=1}^{N_{BC}}\left|u_\theta(\mathbf{x}_i,t_i)-
		g(\mathbf{x}_i,t_i)
		\right|^2.
	\end{equation}
	
	Following common practice in the PINN literature, we minimize $\mathcal{L}$ in two stages: a first-order Adam optimizer~\citep{kingma2014adam} handles rapid initial descent, then a second-order L-BFGS optimizer~\citep{liu1989limited} refines convergence near the minimum. Table~\ref{tab:hyperparameters} reports the iteration budgets used for each benchmark.
	
	
	\subsection{Proposed DBSG-PINN Architecture}
	\label{sec:architecture}
	
	The Dual-Branch Spectral-Gated Physics-Informed Neural Network (DBSG-PINN) splits the approximation of low- and high-frequency solution components into two subnetworks, combined by a learned gate. The input coordinates $(x,t)$ go in parallel to all three subnetworks — the low-frequency branch, the high-frequency branch, and the gate network — which are trained jointly, end to end, by backpropagating the composite loss (Eq.~\ref{eq:loss}) through the combined output. No branch is pretrained or frozen on its own.
	
	The low-frequency branch is represented as
	
	\begin{equation}
		u_{\mathrm{lo}}
		=
		f_{\mathrm{lo}}(\mathbf{x},t;\theta_{\mathrm{lo}}),
	\end{equation}
	
	where $f_{\mathrm{lo}}(\cdot)$ uses hyperbolic tangent activations — a standard smooth nonlinearity backed by classical universal approximation results for feedforward networks~\citep{hornik1989multilayer} — to model smooth solution components.
	
	Similarly, the high-frequency branch is defined as
	
	\begin{equation}
		u_{\mathrm{hi}}=f_{\mathrm{hi}}(\mathbf{x},t;\theta_{\mathrm{hi}}),
	\end{equation}
	
	where $f_{\mathrm{hi}}(\cdot)$ uses sinusoidal activations to better represent oscillatory and multi-scale features, following the periodic-activation design of SIREN-style networks~\citep{sitzmann2020implicit}.
	
	We use ``low-frequency'' and ``high-frequency branch'' as shorthand for each branch's intended inductive bias -- smooth $\tanh$ nonlinearities versus oscillatory $\sin(\omega x)$ activations at a fixed frequency $\omega$ -- not as a formal spectral decomposition. Nothing in the architecture constrains $u_{\mathrm{lo}}$ to contain only low-frequency content or $u_{\mathrm{hi}}$ to contain only high-frequency content: either branch is, in principle, free to represent any function its activation function and depth allow, and the gate is what determines how much each contributes at a given $(x,t)$. Throughout this paper we use ``frequency decomposition'' to mean this soft, activation-driven inductive bias, not a guarantee of exact spectral separation.
	
	To combine the two representations, a lightweight gating network predicts an adaptive weighting coefficient, in the spirit of mixture-of-experts architectures that route between subnetworks with a trained gate~\citep{jacobs1991adaptive,shazeer2017outrageously}. Those architectures usually gate more than two experts and so use a softmax; since DBSG-PINN mixes exactly two branches, we use a scalar sigmoid gate instead,
	
	\begin{equation}
		g(\mathbf{x},t)=\sigma\left(f_g(\mathbf{x},t;\theta_g)\right),
		\label{eq:gate}
	\end{equation}
	
	where $\sigma(\cdot)$ is the sigmoid activation function, so that
	
	\begin{equation}
		0
		\le
		g(\mathbf{x},t)
		\le
		1.
	\end{equation}
	
	The final prediction is a convex combination of the two branches,
	
	\begin{equation}
		u_\theta(\mathbf{x},t)
		=
		g(\mathbf{x},t)
		u_{\mathrm{hi}}(\mathbf{x},t)
		+
		\left(
		1-
		g(\mathbf{x},t)
		\right)
		u_{\mathrm{lo}}(\mathbf{x},t).
		\label{eq:dbsg}
	\end{equation}
	
\begin{figure}[t]
		\centering
		\includegraphics[width=\textwidth]{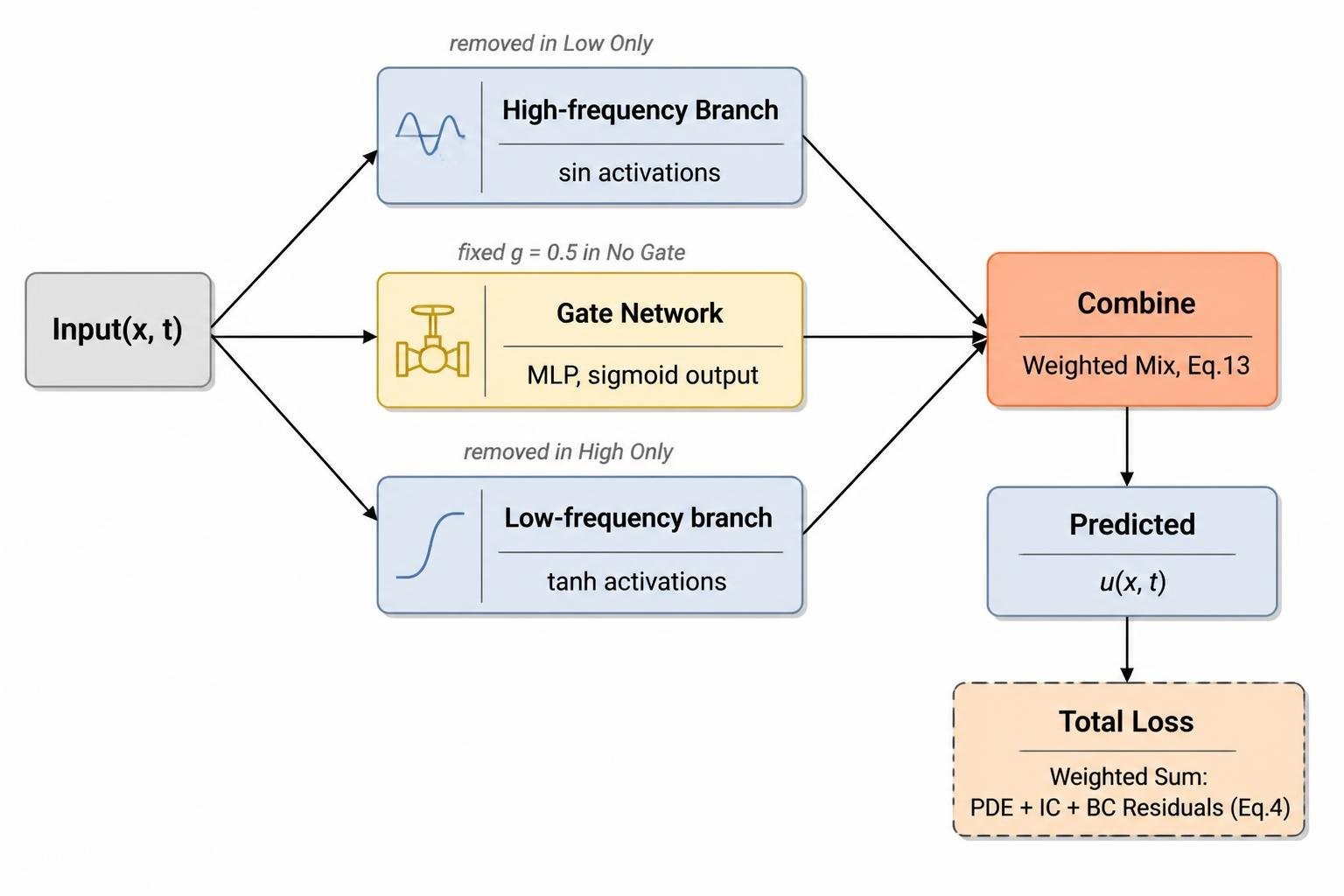}
		\caption{DBSG-PINN architecture}
		\label{fig:dbsg-pinn-architecture}
	\end{figure}

	The PDE residual for this architecture is therefore
	
	\begin{equation}
		r_\theta(\mathbf{x},t)
		=
		\frac{\partial u_\theta}{\partial t}
		+
		\mathcal{N}[u_\theta],
	\end{equation}
	
	where, again, all derivatives come from automatic differentiation. Figure~\ref{fig:dbsg-pinn-architecture} shows the overall framework.
	
	\subsection{Ablation Variants}
	\label{sec:ablation-variants}
	
	Each ablation variant is trained separately, from a fresh random initialization, on its own copy of the computational graph above — built by structurally removing or fixing the relevant component (the high-frequency branch, the low-frequency branch, or the gate), not by defining a different architecture. Every variant uses the same optimizer, iteration budget, collocation strategy, and loss weights as the full model for that benchmark (Table~\ref{tab:hyperparameters}); only the active architectural components change. This reduces the risk that any performance difference comes from optimizer settings or training procedure rather than frequency decomposition itself. It does not, on its own, control for seed variance, which we discuss as a limitation in Section~\ref{sec:discussion}.
	
	\textbf{The low- and high-frequency branches are not strictly capacity-matched.} We state this up front because it bears directly on how LowOnly and HighOnly should be read: depth and width differ between the two branches on most benchmarks (details below), so a LowOnly-vs-HighOnly comparison is not a clean like-for-like test of the two activation functions alone -- some of the difference could come from capacity rather than frequency bias.
	\begin{itemize}
		\item \textbf{Full (DBSG-PINN):} both branches active, combined via the learned gate $g(x,t)$ as in
		Eq.~\ref{eq:dbsg}.
		\item \textbf{NoGate:} both branches active, but $g(x,t)$ in Eq.~\ref{eq:dbsg} is fixed at $0.5$ instead of
		learned, giving a simple average of the two branches.
		\item \textbf{LowOnly:} the high-frequency branch is dropped ($u_\theta = u_{\mathrm{lo}}$); the gate and
		high-frequency branch are not used at all.
		\item \textbf{HighOnly:} the low-frequency branch is dropped in the same way ($u_\theta = u_{\mathrm{hi}}$).
	\end{itemize}
	For each PDE, the low- and high-frequency branches use a roughly similar number of hidden layers and parameters (Table~\ref{tab:hyperparameters}), but depth and width are not identical between the two branches: on most benchmarks the high-frequency branch is wider (e.g.\ 64 vs.\ 32 units on Multimodal Wave, 128 vs.\ 64 on 1D Wave), and depth differs by one or two layers on Burgers, Reaction--Diffusion, and Allen--Cahn. We chose these settings empirically per benchmark rather than by a strict matching rule, so LowOnly and HighOnly are not exactly capacity-matched ablations — they are only roughly comparable in scale, with activation function ($\tanh$ vs.\ $\sin(\omega x)$) as the main intended difference. Even when depth and width do match, that does not fully equate capacity in a functional sense: $\tanh$ and $\sin(\omega x)$ networks of the same size can still differ in expressivity, so any parameter matching here is a partial control, not a complete one. With that caveat in mind, the comparison is meant to highlight the inductive bias from the activation function rather than gross differences in network size, though the per-benchmark width and depth choices remain an uncontrolled source of variation. Together, the four variants probe two questions: does explicit frequency separation with a learned gate help (Full vs.\ NoGate), and is either single-frequency regime alone enough for the target solution (LowOnly / HighOnly vs.\ Full)?
	
	\subsection{Evaluation Metrics}
	\label{sec:metrics}
	
	We report the following metrics throughout Section~\ref{sec:results}, evaluated on a uniform $100 \times 100$ ($N_x \times N_t$) grid over the problem domain (Table~\ref{tab:hyperparameters} gives the collocation spacing used during training; the evaluation grid is separate from, and coarser-to-finer than, the training resolution depending on benchmark). Let $u_\theta$ be the trained network's prediction and $u_{\mathrm{exact}}$ the reference solution, both evaluated on this grid. For Multimodal Wave, Reaction--Diffusion, and 1D Wave, $u_{\mathrm{exact}}$ is the closed-form analytical solution given in Section~\ref{sec:results}. Allen--Cahn and Burgers admit no simple closed form on this domain, so for these two benchmarks $u_{\mathrm{exact}}$ denotes a high-resolution numerical reference solution rather than an analytical one; we say more about how each is obtained in the corresponding subsection of Section~\ref{sec:results}.
	
	\paragraph{Relative $L_2$ error.} Following the evaluation convention \citet{raissi2019physics} introduced for PINNs (used there on, among others, the Burgers and Allen--Cahn benchmarks studied here too) and adopted as a default reporting metric in the DeepXDE library~\citep{lu2021deepxde}, the global relative $L_2$ error over the evaluation grid is
	\begin{equation}
		\mathrm{Rel.L_2} = \frac{\left(\sum_{i,j} \left(u_\theta(x_i,t_j) - u_{\mathrm{exact}}(x_i,t_j)\right)^2\right)^{1/2}}{\left(\sum_{i,j} u_{\mathrm{exact}}(x_i,t_j)^2\right)^{1/2}}.
		\label{eq:rel-l2}
	\end{equation}
	
	\paragraph{$L_\infty$ error.} Reported alongside relative $L_2$ error in the same references~\citep{raissi2019physics,lu2021deepxde}, this is the pointwise maximum absolute error over the evaluation grid,
	\begin{equation}
		L_\infty = \max_{i,j} \left| u_\theta(x_i,t_j) - u_{\mathrm{exact}}(x_i,t_j) \right|.
		\label{eq:linf}
	\end{equation}
	
	\paragraph{Mean PDE residual.} Reporting the governing-equation residual on the evaluation grid as a post-training diagnostic, alongside pointwise accuracy metrics, follows practice established for PINN software such as DeepXDE~\citep{lu2021deepxde}. Unlike the metrics above, the PDE residual reported in Tables~\ref{tab:multimodal-wave}--\ref{tab:1d-wave} is not the autodiff-based residual $r_\theta$ minimized during training (Eq.~\ref{eq:residual}, following \citealp{raissi2019physics}). It is a post-hoc diagnostic: after training, we apply second-order central finite differences to the trained prediction $u_\theta$ on the evaluation grid, approximating each benchmark's own governing-equation operator. For the wave-type benchmarks (Eqs.~\ref{eq:multiwave} and~\ref{eq:wave}), for example, this is $\hat{r}_\theta(x_i,t_j) = \hat{u}_{tt}(x_i,t_j) - c^2\hat{u}_{xx}(x_i,t_j)$, with $\hat{u}_{xx}$ and $\hat{u}_{tt}$ evaluated by central differences at spacing $\Delta x$, $\Delta t$ over the interior evaluation grid; the other benchmarks use the equivalent finite-difference form of their own PDE (Eqs.~\ref{eq:allencahn},~\ref{eq:burgers},~\ref{eq:reaction}). We report the mean absolute residual over interior grid points,
	\begin{equation}
		\overline{|R|} = \frac{1}{(N_x-2)(N_t-2)} \sum_{i=2}^{N_x-1}\sum_{j=2}^{N_t-1} \left| \hat{r}_\theta(x_i,t_j) \right|.
		\label{eq:mean-residual}
	\end{equation}
	Because this is a finite-difference approximation computed separately from training, rather than the exact autodiff residual $r_\theta$, it should be read as a post-hoc consistency check on the trained solution, not as a direct measure of the quantity minimized during optimization.
	
	\paragraph{Spectral error, HF recovery, and LF recovery.} These diagnostics adapt, for a post-training evaluation setting, the Fourier-domain approach \citet{rahaman2019spectral} and \citet{xu2020frequency} used to characterize spectral bias during training: comparing the amplitude spectra of a network's output against the target function across frequency bands to show that low frequencies are learned first (\emph{spectral bias}/\emph{F-Principle}). We apply the same comparison to the trained solution instead of to training dynamics: for each time slice $t_j$, we compute the discrete Fourier amplitude spectra of the predicted and exact solutions along $x$, $\widehat{u}_\theta(k,t_j) = \left|\mathrm{FFT}_x[u_\theta(\cdot,t_j)](k)\right|$ and $\widehat{u}_{\mathrm{exact}}(k,t_j)$ defined the same way, for wavenumbers $k=1,\dots,K_{\max}$ (we use $K_{\max}=20$). We then average over the $N_t$ time slices to get the mean amplitude spectra $\bar{u}_\theta(k) = \frac{1}{N_t}\sum_j \widehat{u}_\theta(k,t_j)$ and $\bar{u}_{\mathrm{exact}}(k)$, defined the same way. We exclude the zero-frequency (DC) component from all spectral metrics because it reflects the mean solution level rather than its oscillatory content. The \emph{spectral error} is the normalized $L_2$ distance between the two mean spectra over all remaining modes,
	\begin{equation}
		\mathrm{SpecErr} = \frac{\left\lVert \bar{u}_\theta(2{:}K_{\max}) - \bar{u}_{\mathrm{exact}}(2{:}K_{\max}) \right\rVert_2}{\left\lVert \bar{u}_{\mathrm{exact}}(2{:}K_{\max}) \right\rVert_2}.
		\label{eq:spec-err}
	\end{equation}
	We split the remaining modes into a low-frequency band $\mathcal{K}_{\mathrm{LF}}$ (modes $2$--$5$) and a high-frequency band $\mathcal{K}_{\mathrm{HF}}$ (modes $6$--$K_{\max}$), and define a per-band normalized error $e_{\mathcal{K}} = \left\lVert \bar{u}_\theta(\mathcal{K}) - \bar{u}_{\mathrm{exact}}(\mathcal{K}) \right\rVert_2 / \left\lVert \bar{u}_{\mathrm{exact}}(\mathcal{K}) \right\rVert_2$. The \emph{LF recovery} and \emph{HF recovery} scores are then
	\begin{equation}
		\mathrm{Recovery}(\mathcal{K}) = \frac{1}{1 + e_{\mathcal{K}}}, \qquad \mathcal{K} \in \{\mathcal{K}_{\mathrm{LF}}, \mathcal{K}_{\mathrm{HF}}\},
		\label{eq:recovery}
	\end{equation}
	a score of $1$ means the two spectra match perfectly in that band, and the score drops toward $0$ as the normalized error grows. Unlike Eq.~\ref{eq:rel-l2}, these are diagnostic scores rather than accuracy metrics: they capture how well the amplitude spectrum is recovered in a given band, regardless of any phase or pointwise error elsewhere.
	
	A caveat applies when a benchmark's reference solution has little or no energy in one of these two bands. Reaction--Diffusion's solution is a single spatial mode ($\sin(\pi x)$, Section~\ref{sec:results}) with essentially all its Fourier energy at the lowest retained mode, and 1D Wave's solution is likewise a single spatial mode ($k=4$) sitting inside the LF band as we define it here. In both cases the reference amplitude in the opposite band is close to zero, so the per-band normalized error $e_{\mathcal{K}}$ divides by a near-zero denominator; the resulting recovery score is then driven mainly by whatever small, largely non-physical high-frequency content the trained network happens to produce (interpolation artifacts, activation-induced ripple, discretization leakage from the FFT treating a non-periodic Dirichlet solution as periodic), rather than by how well a genuine spectral feature is recovered. We still report these columns for completeness and consistency across benchmarks, but on Reaction--Diffusion (HF band) and 1D Wave (HF band, since its only real content sits in the LF band at $k=4$) the recovery score should be read as a noise-floor diagnostic rather than a substantive accuracy measure.
	
	\section{Results}
	\label{sec:results}
	
	Every number below comes from a single training run per model per benchmark (seed in Table~\ref{tab:hyperparameters}), so we read these as observations from an initial study rather than statistically confirmed effects -- Section~\ref{sec:discussion} says more about what that limitation should and should not be taken to mean. Metric definitions (relative $L_2$ error, spectral error, HF/LF recovery) are in Section~\ref{sec:metrics}.
	
	\subsection{1D Multimodal Wave Equation}
	We consider the multimodal wave equation
	\begin{equation}
		u_{tt} = c^2\, u_{xx},
		\qquad (x,t) \in [0,1] \times [0,1], \quad c = 1,
		\label{eq:multiwave}
	\end{equation}
	with a multimodal initial condition composed of modes $k \in \{1,3,5\}$
	with amplitudes $a_k \in \{1.0,\,0.7,\,0.4\}$,
	\begin{equation}
		u(x,0) = \sum_{k \in \{1,3,5\}} a_k \sin(k\pi x),
		\qquad u_t(x,0) = 0,
	\end{equation}
	and Dirichlet boundary conditions $u(0,t) = u(1,t) = 0$.
	This admits the closed-form solution
	\begin{equation}
		u(x,t) = \sum_{k \in \{1,3,5\}} a_k \sin(k\pi x)\cos(k\pi c\, t).
	\end{equation}
	\begin{table}[h]
		\centering
		\caption{Ablation Study of DBSG-PINN on the 1D Multimodal Wave Equation (single seed)}
		\footnotesize
		\setlength{\tabcolsep}{4pt}
		\begin{tabular}{lcccccc}
			\toprule
			Model & Relative L2 Error $\downarrow$ & $L_\infty$ Error $\downarrow$ & Mean PDE Residual $\downarrow$ & Spectral Error $\downarrow$ & HF Recovery $\uparrow$ & LF Recovery $\uparrow$ \\
			\midrule
			\textbf{DBSG-PINN (Full)} & \textbf{0.0762} & \textbf{0.1632} & 0.0218 & \textbf{0.0371} & \textbf{0.8603} & \textbf{0.9658} \\
			NoGate   & 0.1161 & 0.2235 & \textbf{0.0206} & 0.0840 & 0.8019 & 0.9239 \\
			LowOnly  & 0.1869 & 0.4649 & 0.0378 & 0.1387 & 0.7235 & 0.8798 \\
			HighOnly & 0.1632 & 0.2877 & 0.0257 & 0.1380 & 0.7797 & 0.8796 \\
			\bottomrule
		\end{tabular}
		\label{tab:multimodal-wave}
	\end{table}
	
	\subsection{1D Allen--Cahn Equation}
	The Allen--Cahn equation, subject to periodic boundary conditions, takes the form
	\begin{equation}
		u_t = \varepsilon\, u_{xx} + 5u - 5u^3,
		\qquad (x,t) \in [-1,1] \times [0,1],
		\label{eq:allencahn}
	\end{equation}
	with diffusion coefficient $\varepsilon = 10^{-4}$, initial condition
	\begin{equation}
		u(x,0) = x^2 \cos(\pi x),
	\end{equation}
	and periodic boundary conditions
	\begin{equation}
		u(-1,t) = u(1,t), \qquad u_x(-1,t) = u_x(1,t).
	\end{equation}
	This equation has no simple closed-form solution. Following the same benchmark as originally introduced for PINNs~\citep{raissi2019physics}, we evaluate against a high-resolution numerical reference solution (obtained independently of the network being trained) rather than an analytical one; all ``exact''/reference figures and metrics for Allen--Cahn refer to this numerical reference.
	\begin{table}[h]
		\centering
		\caption{Ablation Study of DBSG-PINN on the Allen--Cahn Equation (single seed)}
		\footnotesize
		\setlength{\tabcolsep}{3pt}
		\begin{tabular}{lcccccc}
			\toprule
			Model & Relative L2 $\downarrow$ & $L_\infty$ Error $\downarrow$ & Mean PDE Residual $\downarrow$ & Spectral Error $\downarrow$ & HF Recovery $\uparrow$ & LF Recovery $\uparrow$ \\
			\midrule
			\textbf{DBSG-PINN (Full)} & \textbf{0.1038} & \textbf{0.7416} & 0.0107 & \textbf{0.0700} & \textbf{0.7985} & \textbf{0.9459} \\
			NoGate   & 0.5892 & 1.8905 & 0.0183 & 0.3571 & 0.5898 & 0.7446 \\
			LowOnly  & 0.1108 & 0.7547 & \textbf{0.0084} & 0.0734 & 0.7933 & 0.9430 \\
			HighOnly & 0.6717 & 1.8997 & 0.0519 & 0.4116 & 0.5472 & 0.7174 \\
			\bottomrule
		\end{tabular}
		\label{tab:allen-cahn}
	\end{table}
	
	\subsection{1D Burgers Equation}
	We use the viscous Burgers equation with Dirichlet boundary conditions,
	\begin{equation}
		u_t + u\, u_x = \nu\, u_{xx},
		\qquad (x,t) \in [-1,1] \times [0,1],
		\label{eq:burgers}
	\end{equation}
	with viscosity $\nu = 0.01/\pi$, initial condition
	\begin{equation}
		u(x,0) = -\sin(\pi x),
	\end{equation}
	and boundary conditions
	\begin{equation}
		u(-1,t) = 0, \qquad u(1,t) = 0.
	\end{equation}
	This viscosity and initial condition admit no simple closed-form solution either; following the same benchmark as in~\citep{raissi2019physics}, we evaluate against a high-resolution numerical reference solution rather than an analytical one.
	
	\begin{table}[h]
		\centering
		\caption{Ablation Study of DBSG-PINN on the 1D Burgers Equation (single seed)}
		\footnotesize
		\setlength{\tabcolsep}{4pt}
		\begin{tabular}{lcccccc}
			\toprule
			Model & Relative L2 $\downarrow$ & $L_\infty$ Error $\downarrow$ & Mean PDE Residual $\downarrow$ & Spectral Error $\downarrow$ & HF Recovery $\uparrow$ & LF Recovery $\uparrow$ \\
			\midrule
			DBSG-PINN (Full) & 0.02398 & 0.16720 & \textbf{0.26810} & \textbf{0.01447} & \textbf{0.94387} & 0.99098 \\
			\textbf{NoGate}  & \textbf{0.02383} & \textbf{0.15984} & 0.26836 & 0.01495 & 0.94079 & \textbf{0.99107} \\
			LowOnly  & 0.02816 & 0.17997 & 0.27716 & 0.01593 & 0.93585 & 0.99089 \\
			HighOnly & 0.02455 & 0.16313 & 0.26866 & 0.01631 & 0.93767 & 0.99081 \\
			\bottomrule
		\end{tabular}
		\label{tab:burgers}
	\end{table}
	
	\subsection{1D Reaction--Diffusion Equation}
	This benchmark is governed by the linear reaction--diffusion equation
	\begin{equation}
		u_t = D\, u_{xx} + \lambda u,
		\qquad (x,t) \in [0,1] \times [0,1],
		\label{eq:reaction}
	\end{equation}
	with diffusion coefficient $D = 0.01$ and reaction rate $\lambda = 2$,
	initial condition $u(x,0) = \sin(\pi x)$, and Dirichlet boundary
	conditions $u(0,t) = u(1,t) = 0$, with exact solution
	\begin{equation}
		u(x,t) = \sin(\pi x)\,\exp\!\big((\lambda - D\pi^2)\,t\big).
	\end{equation}
	Unlike Multimodal Wave, this solution is a single spatial mode -- a $\sin(\pi x)$ profile that grows or decays in time depending on the sign of $\lambda - D\pi^2$ -- so we include this benchmark as a smooth, low-order case rather than a spectrally complex one; we return to what that implies for the ablation results in Section~\ref{sec:discussion}.
	
	\begin{table}[h]
		\centering
		\caption{Ablation Study of DBSG-PINN on the 1D Reaction--Diffusion Equation (single seed)}
		\footnotesize
		\setlength{\tabcolsep}{4pt}
		\begin{tabular}{lcccccc}
			\toprule
			Model & Relative L2 Error $\downarrow$ & $L_\infty$ Error $\downarrow$ & Mean PDE Residual $\downarrow$ & Spectral Error $\downarrow$ & HF Recovery $\uparrow$ & LF Recovery $\uparrow$ \\
			\midrule
			\textbf{DBSG-PINN (Full)} & $\bm{2.24\times10^{-4}}$ & $\bm{6.30\times10^{-3}}$ & $\bm{2.30\times10^{-3}}$ & $\bm{2.63\times10^{-4}}$ & 0.9956 & \textbf{0.9998} \\
			NoGate   & $3.25\times10^{-4}$ & $6.95\times10^{-3}$ & $2.51\times10^{-3}$ & $3.44\times10^{-4}$ & 0.9980 & 0.9997 \\
			LowOnly  & $4.77\times10^{-4}$ & $6.30\times10^{-3}$ & $2.53\times10^{-3}$ & $5.37\times10^{-4}$ & \textbf{0.9985} & 0.9995 \\
			HighOnly & $1.01\times10^{-3}$ & $8.85\times10^{-3}$ & $5.00\times10^{-3}$ & $1.18\times10^{-3}$ & 0.9937 & 0.9988 \\
			\bottomrule
		\end{tabular}
		\label{tab:reaction}
	\end{table}

	\subsection{1D Wave Equation}
	The final benchmark is the standard 1D wave equation
	\begin{equation}
		u_{tt} = c^2\, u_{xx},
		\qquad (x,t) \in [0,1] \times [0,1], \quad c = 1,
		\label{eq:wave}
	\end{equation}
	with a single-mode initial condition of wavenumber $k = 4$,
	\begin{equation}
		u(x,0) = \sin(4\pi x), \qquad u_t(x,0) = 0,
	\end{equation}
	and Dirichlet boundary conditions $u(0,t) = u(1,t) = 0$,
	whose exact solution is
	\begin{equation}
		u(x,t) = \sin(4\pi x)\cos(4\pi c\, t).
	\end{equation}
	
	\begin{table}[h]
		\centering
		\caption{Ablation Study of DBSG-PINN on the 1D Wave Equation (single seed)}
		\footnotesize
		\setlength{\tabcolsep}{4pt}
		\begin{tabular}{lcccccc}
			\toprule
			Model & Relative L2 $\downarrow$ & $L_\infty$ Error $\downarrow$ & Mean PDE Residual $\downarrow$ & Spectral Error $\downarrow$ & HF Recovery $\uparrow$ & LF Recovery $\uparrow$ \\
			\midrule
			DBSG-PINN (Full) & 0.05134 & 0.08320 & 0.02931 & 0.02699 & 0.63539 & 0.97581 \\
			NoGate   & \textbf{0.02397} & 0.05143 & 0.02711 & \textbf{0.00850} & 0.78424 & \textbf{0.99326} \\
			LowOnly  & 0.49170 & 0.67322 & \textbf{0.02240} & 0.35938 & 0.29961 & 0.73716 \\
			HighOnly & 0.02514 & \textbf{0.04897} & 0.03340 & 0.01565 & \textbf{0.98891} & 0.98475 \\
			\bottomrule
		\end{tabular}
		\label{tab:1d-wave}
	\end{table}
	\section{Discussion}
	\label{sec:discussion}
	
	Every number in this section comes from one trained model per benchmark per variant. We treat the pattern below as a hypothesis to test with more seeds and benchmarks, not as a settled result. To avoid repeating that caveat in every paragraph, we say it once here and come back to it, with specifics, in the Limitations paragraph.
	
	\paragraph{Decomposition helps most on genuinely multi-scale targets.} On Multimodal Wave -- the one benchmark whose solution is an explicit superposition of multiple spatial modes ($k \in \{1,3,5\}$, Section~\ref{sec:results}) -- the full model beats every ablation on nearly every metric (Table~\ref{tab:multimodal-wave}). LowOnly and HighOnly each specialize in one part of the spectrum, and each does measurably worse than the full model at recovering the content the other branch was meant to handle, which is roughly what we expected going in: separating the branches pays off when a solution genuinely mixes frequency regimes. Reaction--Diffusion also shows Full leading on every accuracy metric (Table~\ref{tab:reaction}) -- LowOnly edges it out only on HF recovery, which Section~\ref{sec:metrics} already flags as a noise-floor diagnostic rather than a substantive measure for this benchmark's near-single-mode spectrum -- but that benchmark's solution is a single spatial mode (Section~\ref{sec:results}), not a spectrally rich target, so we do not think the same explanation applies there; we return to this discrepancy below.
	
	\paragraph{On Allen--Cahn, LowOnly nearly matches the full model.} Full still comes out ahead (0.1038 vs.\ 0.1108 relative $L_2$ for LowOnly), though only just, whereas the gap to NoGate and HighOnly is large (Table~\ref{tab:allen-cahn}). The Allen--Cahn solution is smooth and low-frequency apart from sharp, localized transition layers, so a sinusoidal high-frequency branch mostly gets in the way here: HighOnly and the unweighted NoGate average both pay for that mismatch. It looks like the gate's real job on this benchmark is suppressing a branch that is not helping rather than blending two useful ones, though Full and LowOnly sit close enough together that a different seed could flip the ranking.
	
	\paragraph{On Burgers, gating showed no meaningful edge either way.} NoGate (fixed at $g=0.5$) and the full model end up close together across all six metrics for Burgers (Table~\ref{tab:burgers}): 0.0240 vs.\ 0.0238 relative $L_2$, say, with each variant ahead on roughly half the columns and by less than 5\% either way. Viscosity smooths Burgers out to something close to single-scale, so there is not much spectral content left for an adaptive gate to route between -- a wash is more or less what we would expect here.
	
	\paragraph{On 1D Wave, the full model did meaningfully worse than NoGate.} Unlike Burgers, this gap is not small: NoGate leads the full model on every metric we tracked (Table~\ref{tab:1d-wave}), by around 53\% on relative $L_2$ and 68\% on spectral error. The benchmark uses a single wavenumber ($k=4$), so we didn't expect the gate to have much to route between — instead, learning a gate actively hurt performance compared to a fixed average. We don't have a confident explanation for why the effect is this large on Wave but not Burgers; that's a priority for multi-seed follow-up.
	
	\paragraph{Interpreting the gate.} Stepping back, the gate's benefit is largest on the benchmark with the richest spectral content (Multimodal Wave), smaller but still positive on Allen--Cahn (smooth except for sharp transition layers), roughly neutral on Burgers, and negative on 1D Wave, whose target is a single spatial mode. This much fits what we would expect if the gate's aggregate benefit tracks spectral complexity across benchmarks. Reaction--Diffusion is an exception: its solution is also a single spatial mode, yet the full model still leads on every accuracy metric there (HF recovery aside, a noise-floor diagnostic for this benchmark), so whatever the gate is doing on that benchmark, it does not look like it is exploiting multi-mode spectral structure in the way our framing above assumes. We flag this as an open question rather than force it into the pattern, and note again that none of this is based on a direct visualization of $g(x,t)$ -- only on comparing aggregate performance across benchmarks of differing spectral content.
	
	\paragraph{Limitations.} Three limitations bound how far these results should be read. \emph{Seeds:} every result comes from a single training seed (Table~\ref{tab:hyperparameters}), so we have no variance estimate. This matters most for Allen--Cahn's Full-vs-LowOnly gap, which is small enough to plausibly flip under a different seed, and arguably even more for the 1D Wave Full-vs-NoGate gap — that gap is large in this run, but with a single seed we can't tell whether it's a real, reproducible effect or just an unlucky training run for the full model. \emph{Capacity matching:} branch depth and width are only roughly comparable, not identical, across the low- and high-frequency branches for most benchmarks (Section~\ref{sec:ablation-variants}); even where sizes do match, equal parameter count doesn't guarantee equal functional capacity, since $\tanh$ and $\sin(\omega x)$ networks of the same size aren't guaranteed equal expressivity. \emph{Scope:} all five benchmarks are one-dimensional, so we can't say whether the pattern holds in higher dimensions or under stronger nonlinearity.
	
	\section{Conclusion}
	\label{sec:conclusion}
	
	This work asked when frequency decomposition actually helps Physics-Informed Neural Networks, instead of assuming its value is universal. We introduced DBSG-PINN, a dual-branch architecture with an adaptive spectral gate built specifically for ablation: each of its three components — the low-frequency branch, the high-frequency branch, and the gating mechanism — can be removed without changing the optimization procedure, so we can compare their individual contributions directly.
	
	The pattern across the five benchmarks is not a single clean story. Decomposition helped most on Multimodal Wave, the one benchmark whose solution is a genuine superposition of spatial modes, cutting relative $L_2$ error by up to 59.2\%. The full gated model also beat every ablation on every accuracy metric for Reaction--Diffusion (HF recovery aside, a noise-floor diagnostic there), even though that benchmark's solution is a single spatial mode rather than a spectrally rich one -- we flag this as an open question rather than a confirmation of our starting hypothesis. Allen--Cahn was more mixed, with Full ahead but only barely so over LowOnly, suggesting the gate's job there was really screening out a poorly matched high-frequency branch rather than blending two useful ones. Burgers gave learned gating no meaningful edge over a simple fixed average, and 1D Wave went the other way entirely: NoGate beat the full model by a wide margin on every metric tracked, a gap large enough that it needs multi-seed confirmation before being read as more than a flag. Where the gate did help, its benefit tracked the benchmark's spectral complexity -- a preliminary, interpretable signal, not a proven mechanism.
	
	None of this settles when frequency decomposition helps, and it is not meant to. The single-seed, five-benchmark, one-dimensional setting is a real limitation, and the closer comparisons here, Allen--Cahn's Full-vs-LowOnly gap especially, should not be read past what is actually reported. What it does offer is a concrete, testable pattern: decomposition helps most on genuinely multi-scale solutions and least on smooth, single-scale ones. The natural next step is checking whether that pattern survives more seeds and a wider benchmark suite, before drawing any firmer conclusions for PINN practice.
	
	\begin{figure}[t]
		\centering
		\includegraphics[width=0.95\textwidth]{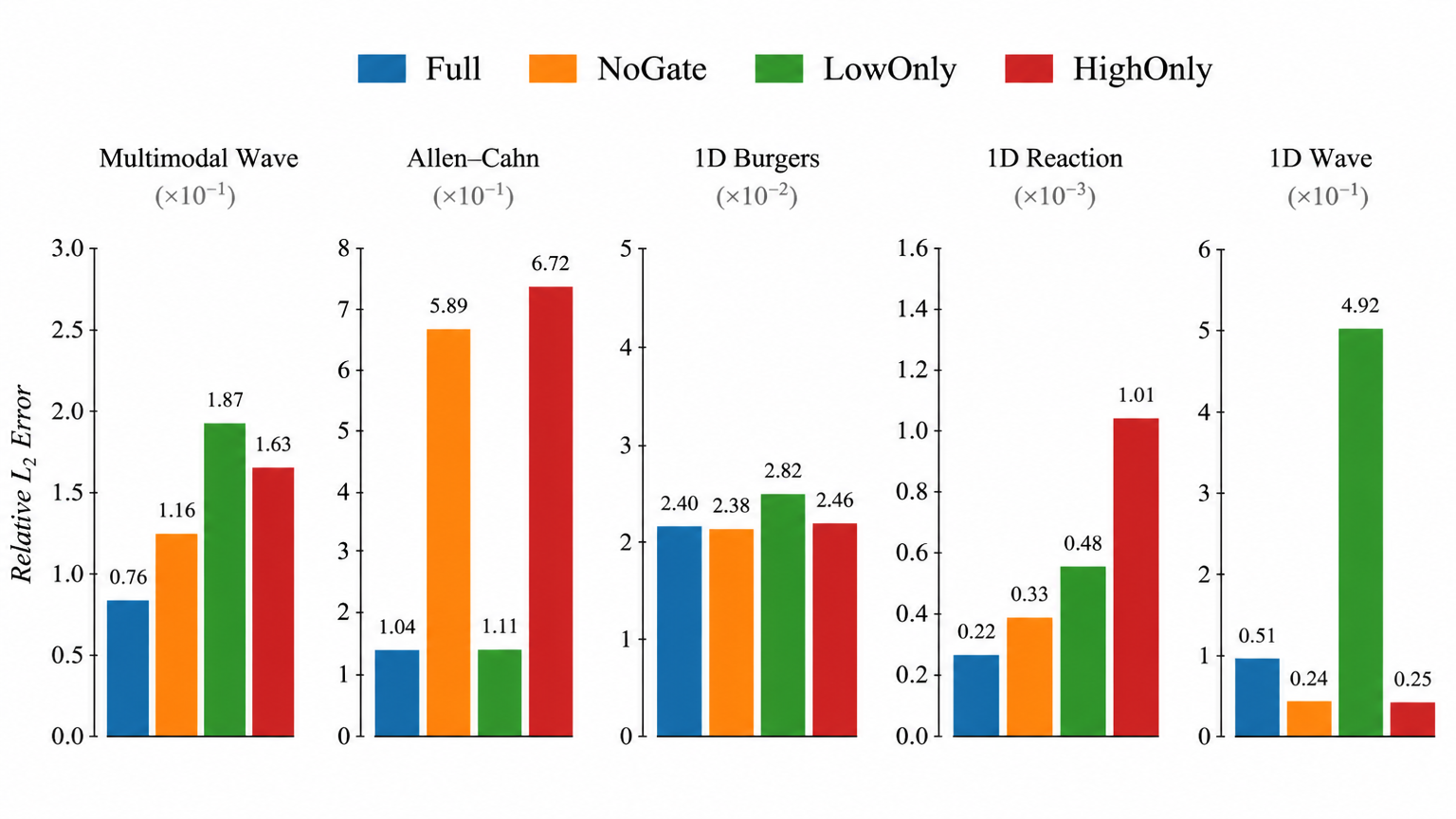}
		\caption{Relative $L_2$ error by ablation variant across the five benchmarks (single seed per bar). DBSG-PINN (Full) leads on Multimodal Wave, Allen--Cahn, and Reaction--Diffusion, while NoGate leads on Burgers and 1D Wave.}
		\label{fig:l2-ablation}
	\end{figure}
	\paragraph{Future work.} The most immediate next step is repeating this ablation study across multiple seeds to get variance estimates and confirm whether the reported orderings are stable; several of the comparisons in Section~\ref{sec:discussion} are close enough that this is a prerequisite for any stronger claim. Beyond that, natural extensions include: (i) extending the benchmark suite to 2D and time-dependent multi-scale PDEs (e.g., 2D Navier--Stokes, wave scattering); (ii) building a lightweight diagnostic, based on the spectral content of the initial/boundary data, that estimates in advance whether frequency decomposition is likely to help for a given PDE; and (iii) testing whether the gate network's link to spectral complexity holds up as an interpretability signal beyond the specific architecture studied here.
	
	\section*{Acknowledgements}
	Figures~\ref{fig:dbsg-pinn-architecture} and~\ref{fig:l2-ablation} were generated with the assistance of Paper Banana. Claude (Anthropic) was used for language refinement of the manuscript text. All AI-assisted content was reviewed and verified by the author, who takes full responsibility for the research, experiments, analysis, and conclusions presented in this work.
	
	\newpage
	\bibliographystyle{apalike}
	\bibliography{reference}
	\newpage
	
	\appendix
	\section{Appendix}
	\subsection{Hyperparameter Configuration}
	Table~\ref{tab:hyperparameters} reports the hyperparameter configuration we used for DBSG-PINN across all five benchmark PDEs, including each branch's architecture (activation function, hidden layers, width), the gate network configuration, and training settings (collocation grid spacing, Adam/L-BFGS iteration counts, loss weights, and random seed). All ablation variants (NoGate, LowOnly, HighOnly) reuse the same per-benchmark configuration from Table~\ref{tab:hyperparameters} for their active components. Training budget and per-branch capacity therefore stay matched across variants for a given benchmark, though this alone gives no variance estimate, since only one seed was run per configuration (Section~\ref{sec:discussion}).
	\begin{table}[h]
		\centering
		\caption{\small Hyperparameter configuration of DBSG-PINN across the five benchmark PDEs. All entries use a single random seed per benchmark; see Section~\ref{sec:discussion} for the resulting limitation on statistical confidence.}
		\label{tab:hyperparameters}
		\small
		\setlength{\tabcolsep}{4pt}
		\begin{tabular}{lccccc}
			\toprule
			Hyperparameter & Burgers & Reaction--Diff. & Allen--Cahn & Multimodal Wave & 1D Wave \\
			\midrule
			\multicolumn{6}{l}{\textit{Low-frequency branch}} \\
			Activation        & tanh & tanh & tanh & tanh & tanh \\
			Hidden layers     & 3 & 3 & 4 & 3 & 3 \\
			Width             & 24 & 32 & 64 & 32 & 64 \\
			\midrule
			\multicolumn{6}{l}{\textit{High-frequency branch}} \\
			Activation        & $\sin(\omega x)$ & $\sin(\omega x)$ & $\sin(\omega x)$ & $\sin(\omega x)$ & $\sin(\omega x)$ \\
			Frequency $\omega$ & 2 & 2 & 1 & 2 & 4 \\
			Hidden layers     & 4 & 4 & 2 & 3 & 3 \\
			Width             & 32 & 32 & 128 & 64 & 128 \\
			\midrule
			\multicolumn{6}{l}{\textit{Gate network}} \\
			Activation        & tanh\,/\,sigmoid & tanh\,/\,sigmoid & tanh\,/\,sigmoid & tanh\,/\,sigmoid & tanh\,/\,sigmoid \\
			Hidden layers     & 1 & 2 & 1 & 2 & 2 \\
			Width             & 16 & 16 & 32 & 16 & 32 \\
			Sigmoid gain      & 6 & 6 & 1 & 1 & 1 \\
			\midrule
			\multicolumn{6}{l}{\textit{Training}} \\
			Collocation grid spacing & 0.02 & 0.02 & 0.025 & 0.02 & 0.01 \\
			Adam learning rate & $5\times10^{-4}$ & $5\times10^{-4}$ & $5\times10^{-4}$ & $5\times10^{-4}$ & $5\times10^{-4}$ \\
			Adam iterations   & 15{,}000 & 10{,}000 & 18{,}000 & 15{,}000 & 25{,}000 \\
			L-BFGS iterations & 3{,}000 & 3{,}000 & 6{,}000 & 4{,}000 & 8{,}000 \\
			IC loss weight    & 1 & 1 & 50 & 1 & 1 \\
			Random seed       & 42 & 42 & 42 & 42 & 42 \\
			\bottomrule
		\end{tabular}
	\end{table}
	\paragraph{Implementation.} All models are implemented in Julia using NeuralPDE.jl~\citep{zubov2021neuralpde} for the physics-informed training loop and Lux.jl version 1.2.3 ~\citep{pal2023lux} for network parameterization, with grid-based collocation (\texttt{GridTraining}) at the per-benchmark spacings listed above. Training uses Adam~\citep{kingma2014adam} at the learning rate given above, followed by L-BFGS~\citep{liu1989limited} for the iteration counts listed, both with default (Julia \texttt{Optim.jl}/\texttt{OptimizationOptimJL}) line-search settings. All evaluation metrics in Section~\ref{sec:results} (relative $L_2$, $L_\infty$, mean PDE residual, spectral error, HF/LF recovery) are computed on a uniform $100\times100$ evaluation grid, as stated in Section~\ref{sec:metrics}; this grid is independent of, and in most cases finer than, the training collocation spacing above.
	\subsection{1D Multimodal Wave Equation: Qualitative Comparison Across Variants}
\begin{figure}[t]
		\centering
		\begin{subfigure}[b]{0.33\textwidth}
			\centering
			\includegraphics[width=\textwidth]{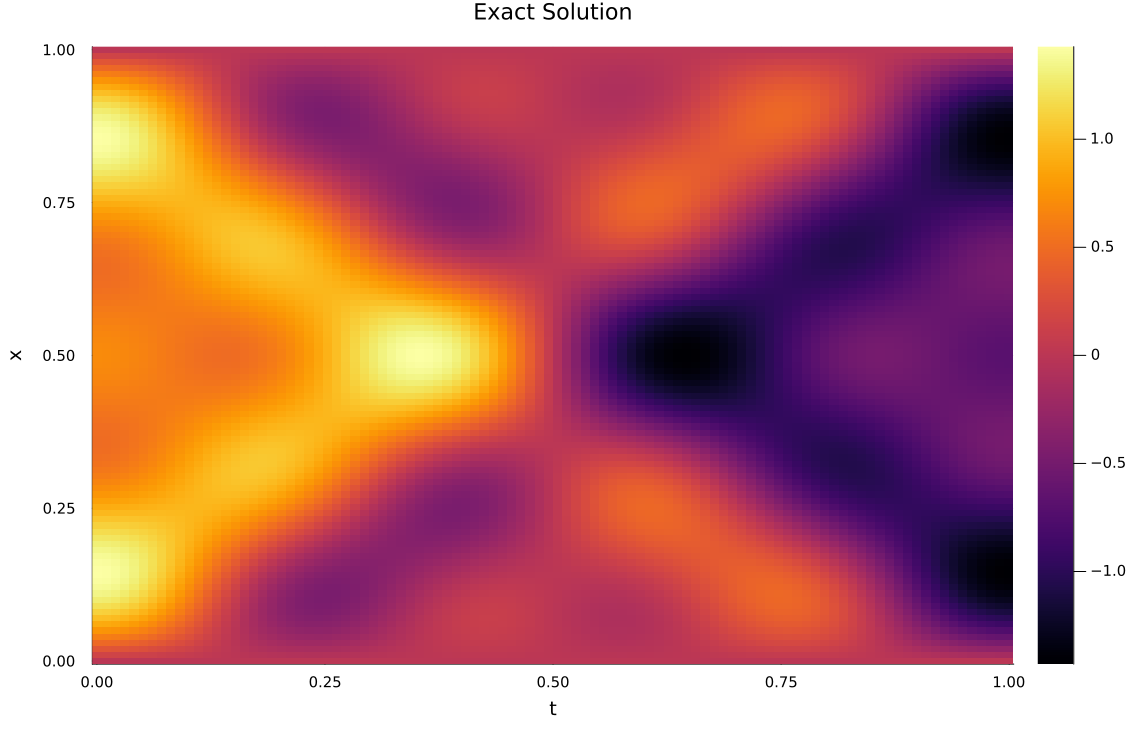}
		\end{subfigure}
		\hfill
		\begin{subfigure}[b]{0.33\textwidth}
			\centering
			\includegraphics[width=\textwidth]{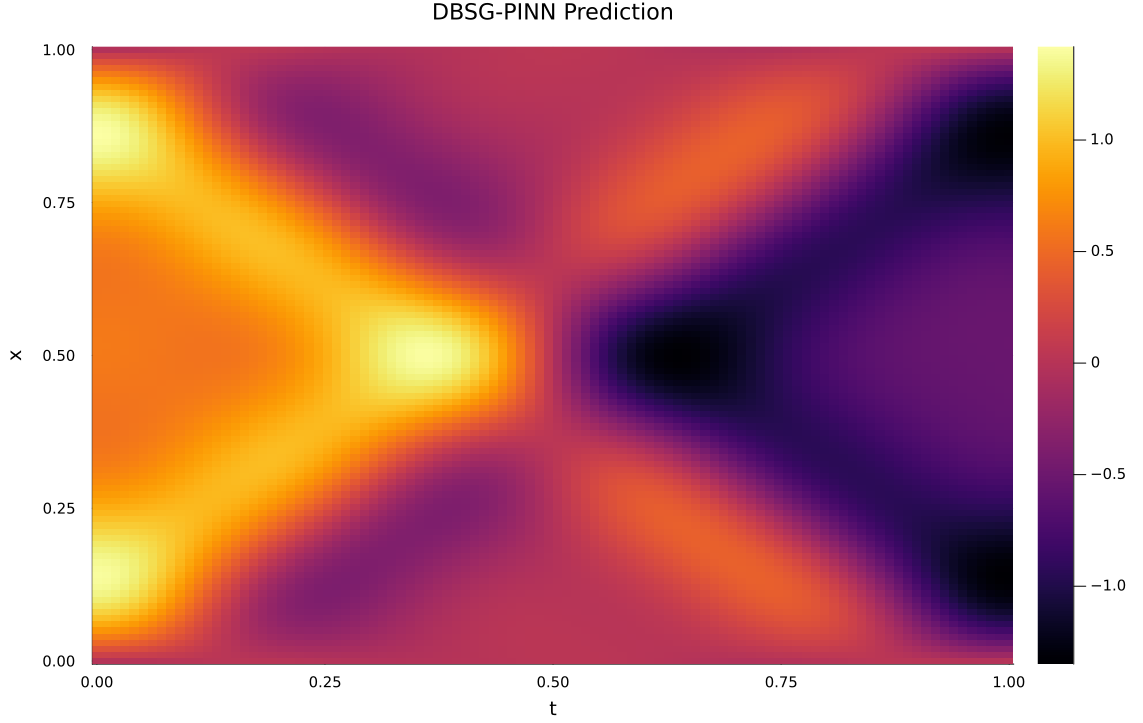}
		\end{subfigure}
		\hfill
		\begin{subfigure}[b]{0.33\textwidth}
			\centering
			\includegraphics[width=\textwidth]{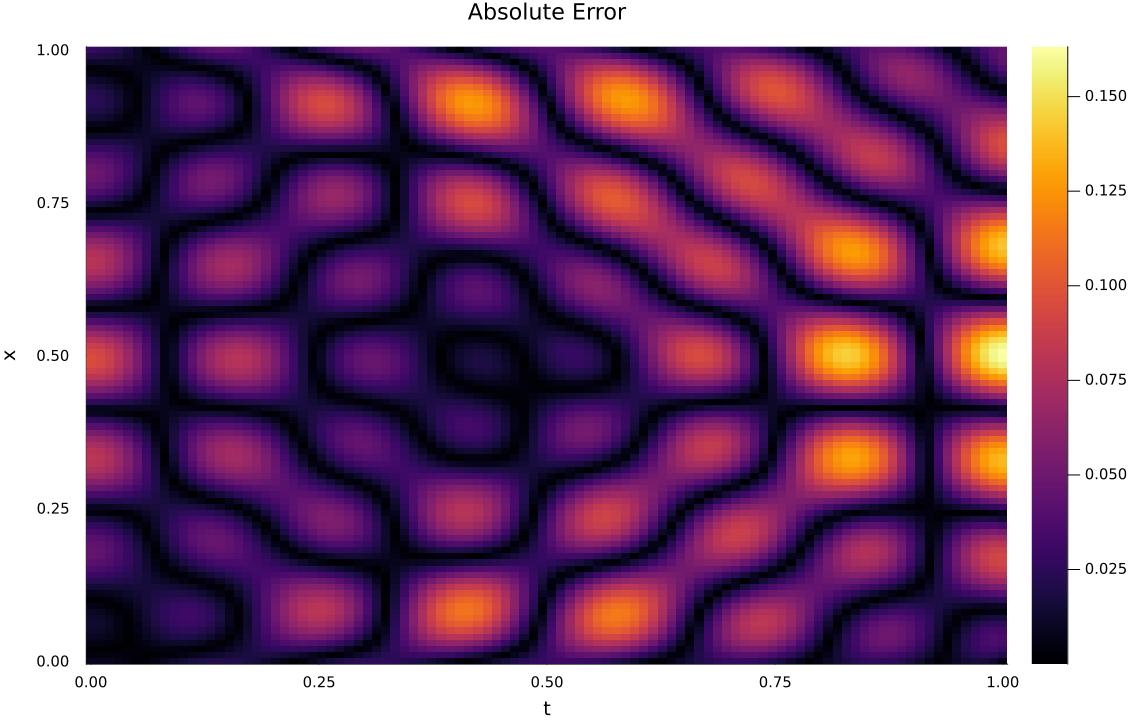}
		\end{subfigure}
		\caption{1D Multimodal Wave, Full DBSG-PINN: exact solution, prediction, and absolute error.}
		\label{fig:1D_Multimodal_Wave_Full_DBSG}
	\end{figure}

\begin{figure}[t]
		\centering
		\begin{subfigure}[b]{0.33\textwidth}
			\centering
			\includegraphics[width=\textwidth]{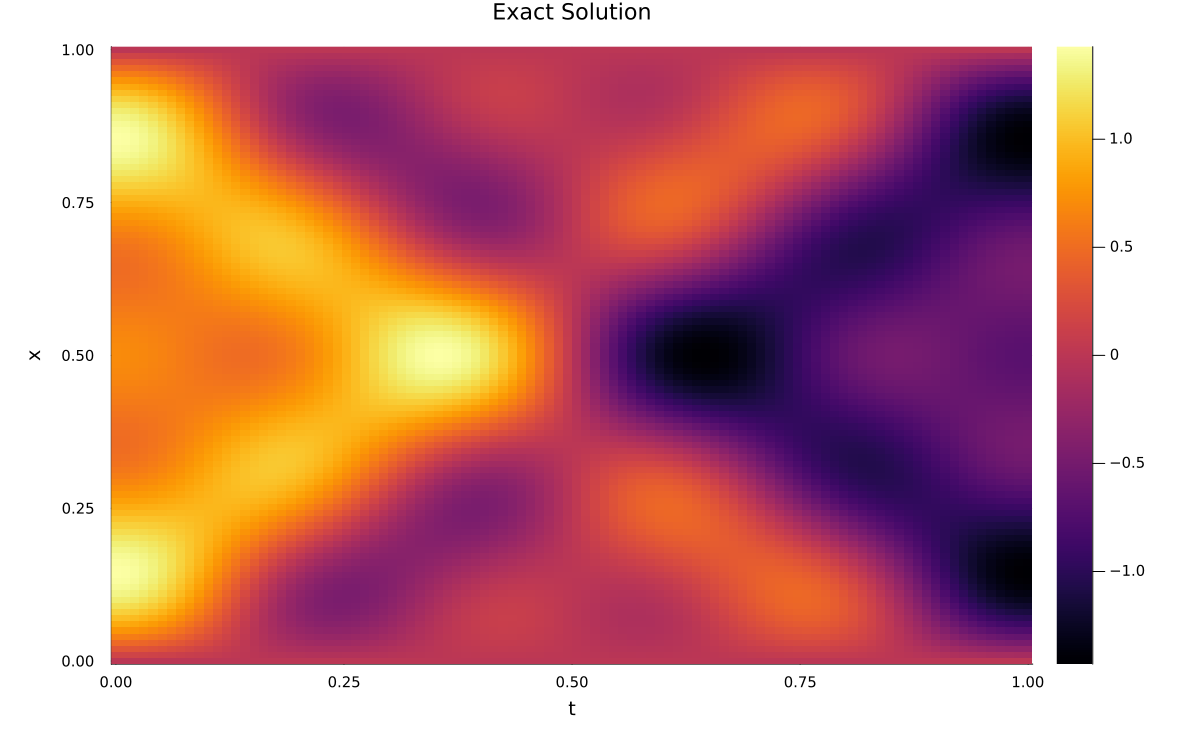}
		\end{subfigure}
		\hfill
		\begin{subfigure}[b]{0.33\textwidth}
			\centering
			\includegraphics[width=\textwidth]{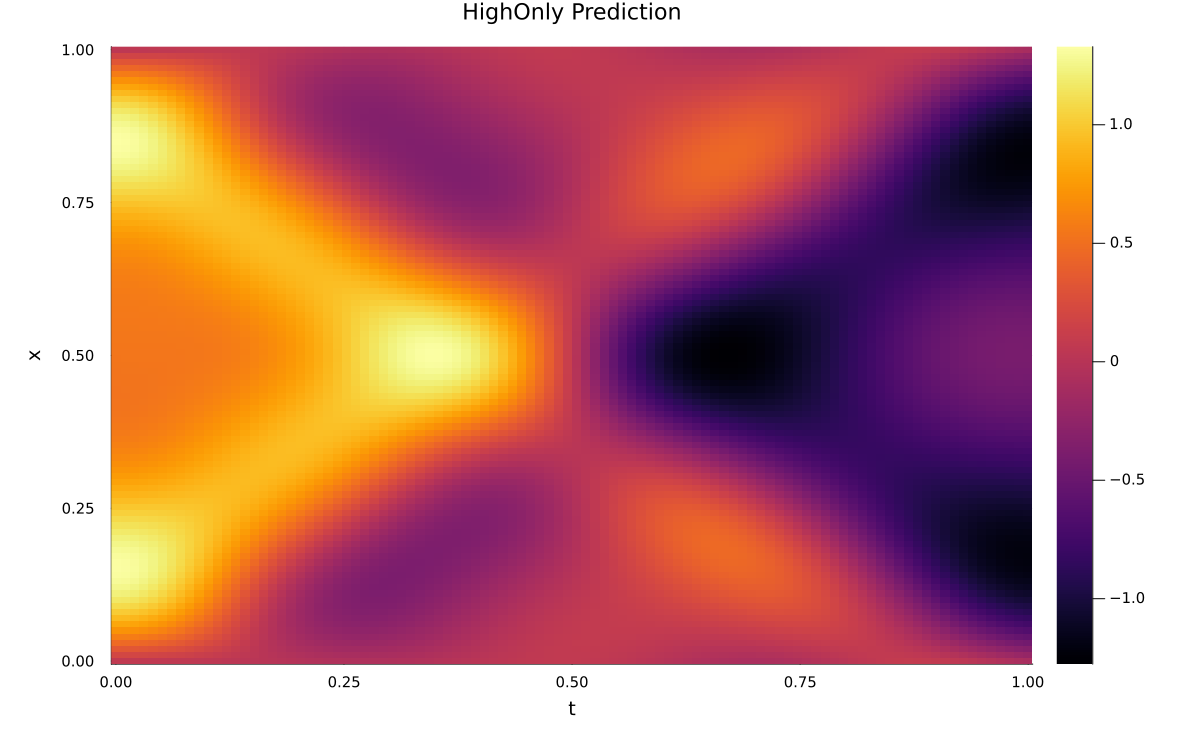}
		\end{subfigure}
		\hfill
		\begin{subfigure}[b]{0.33\textwidth}
			\centering
			\includegraphics[width=\textwidth]{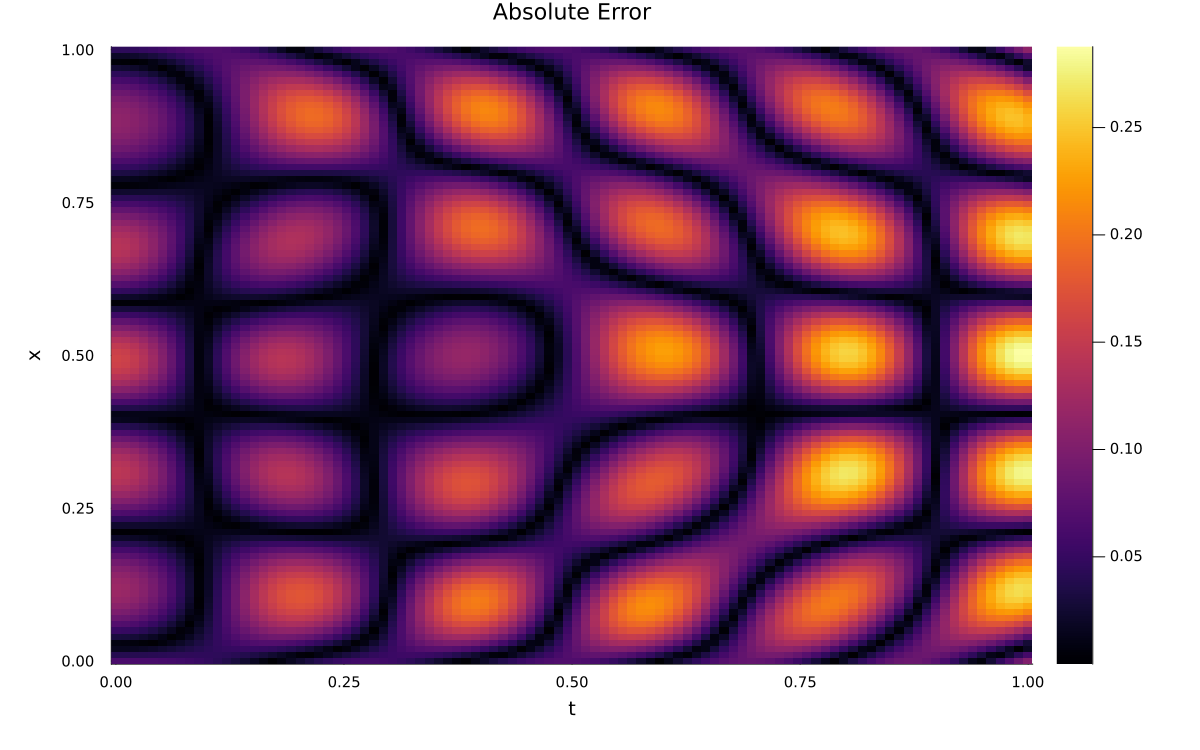}
		\end{subfigure}
		\caption{1D Multimodal Wave, HighOnly variant.}
		\label{fig:1D_MM_WAVE_HIGHONLY}
	\end{figure}

\begin{figure}[t]
		\centering
		\begin{subfigure}[b]{0.33\textwidth}
			\centering
			\includegraphics[width=\textwidth]{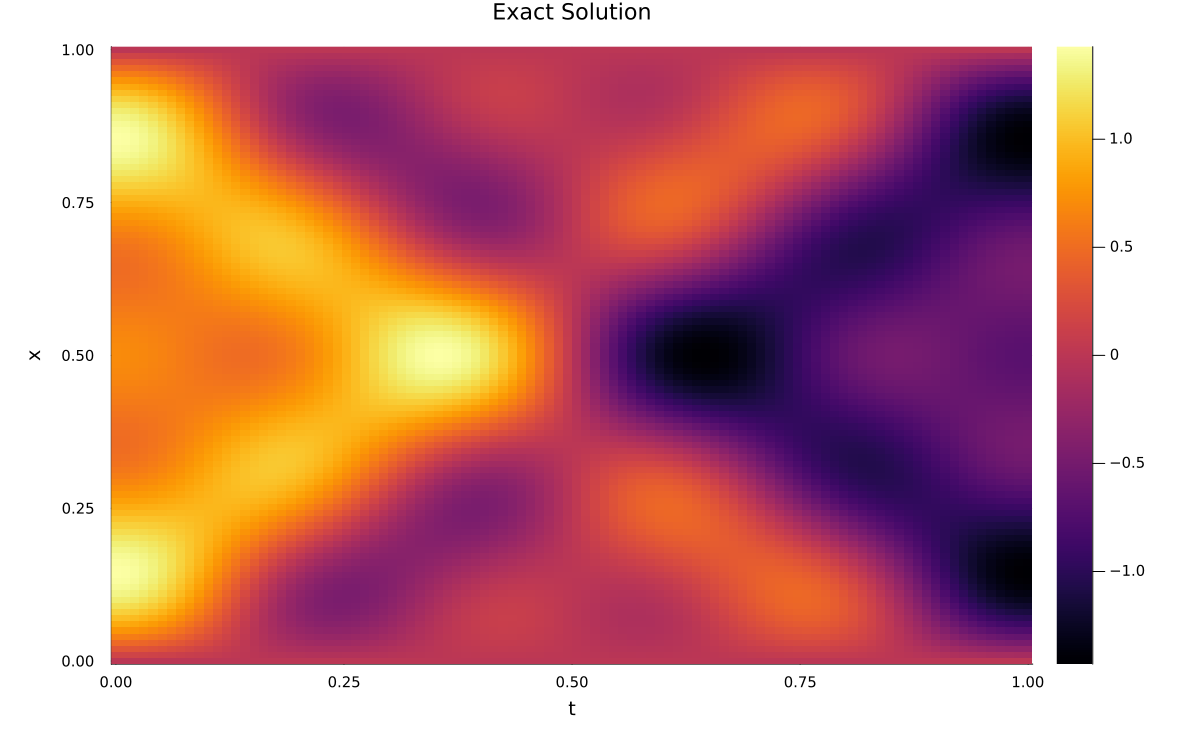}
		\end{subfigure}
		\hfill
		\begin{subfigure}[b]{0.33\textwidth}
			\centering
			\includegraphics[width=\textwidth]{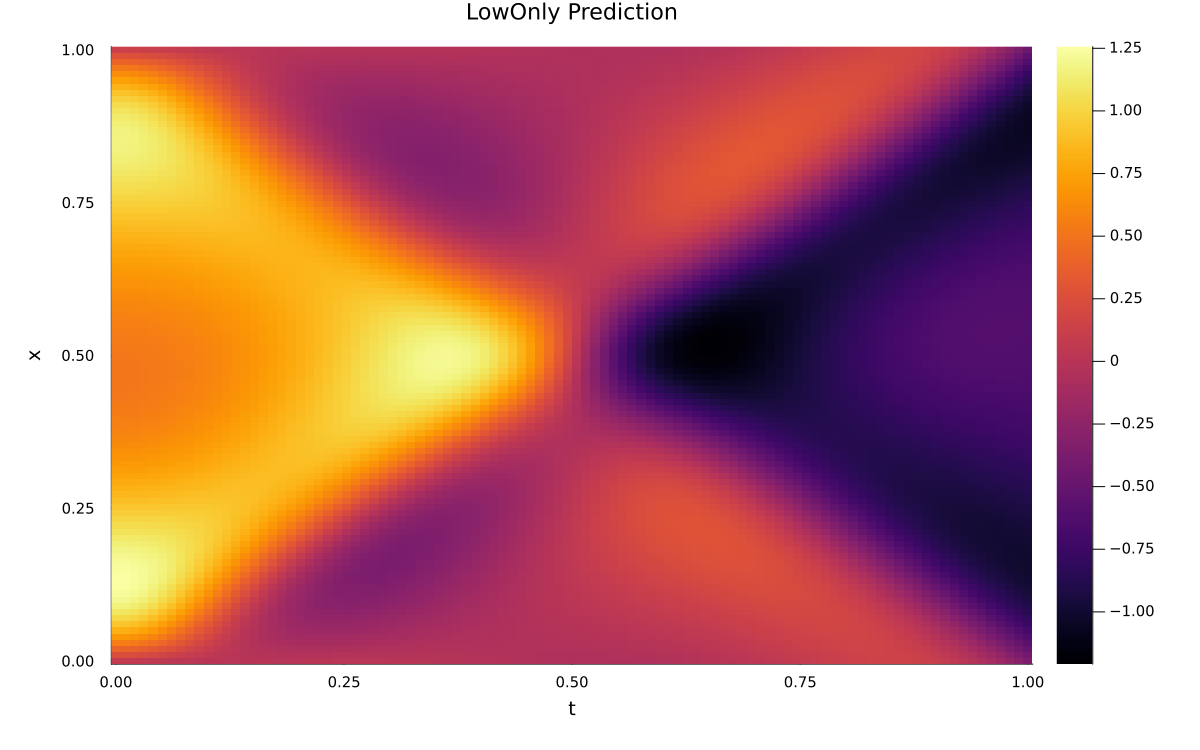}
		\end{subfigure}
		\hfill
		\begin{subfigure}[b]{0.33\textwidth}
			\centering
			\includegraphics[width=\textwidth]{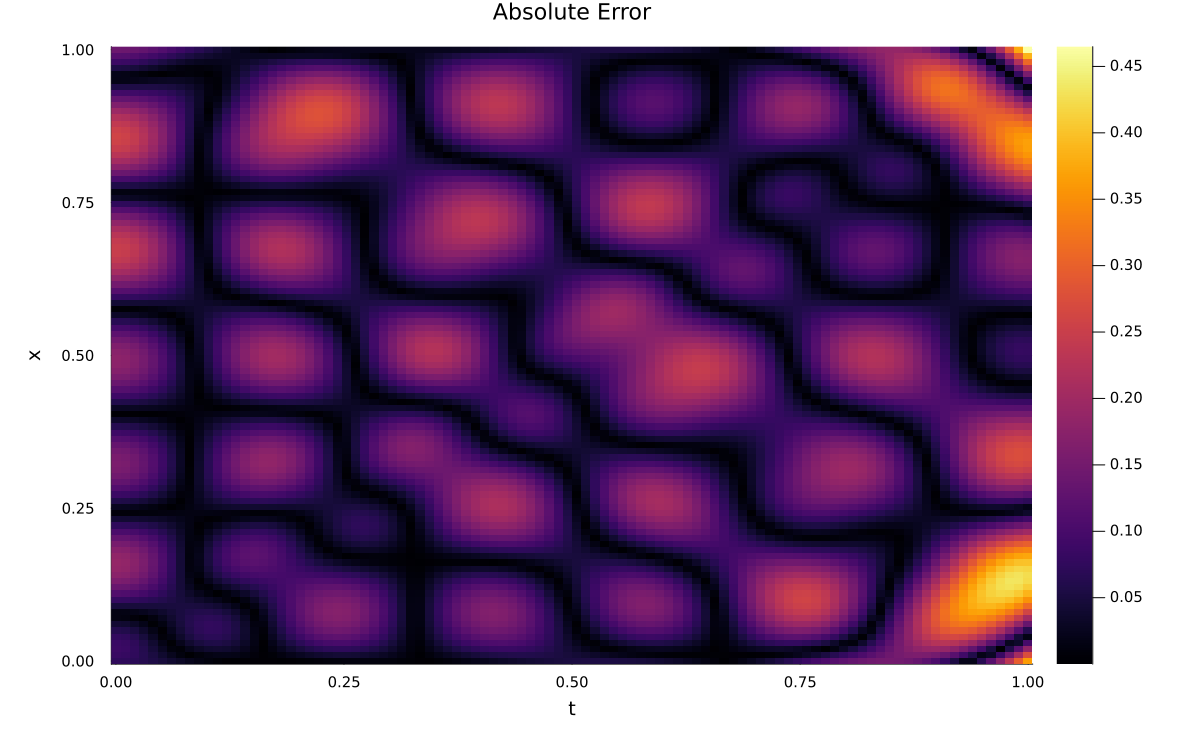}
		\end{subfigure}
		\caption{1D Multimodal Wave, LowOnly variant.}
		\label{fig:1D_MM_WAVE_LOW_ONLY}
	\end{figure}

\begin{figure}[t]
		\centering
		\begin{subfigure}[b]{0.33\textwidth}
			\centering
			\includegraphics[width=\textwidth]{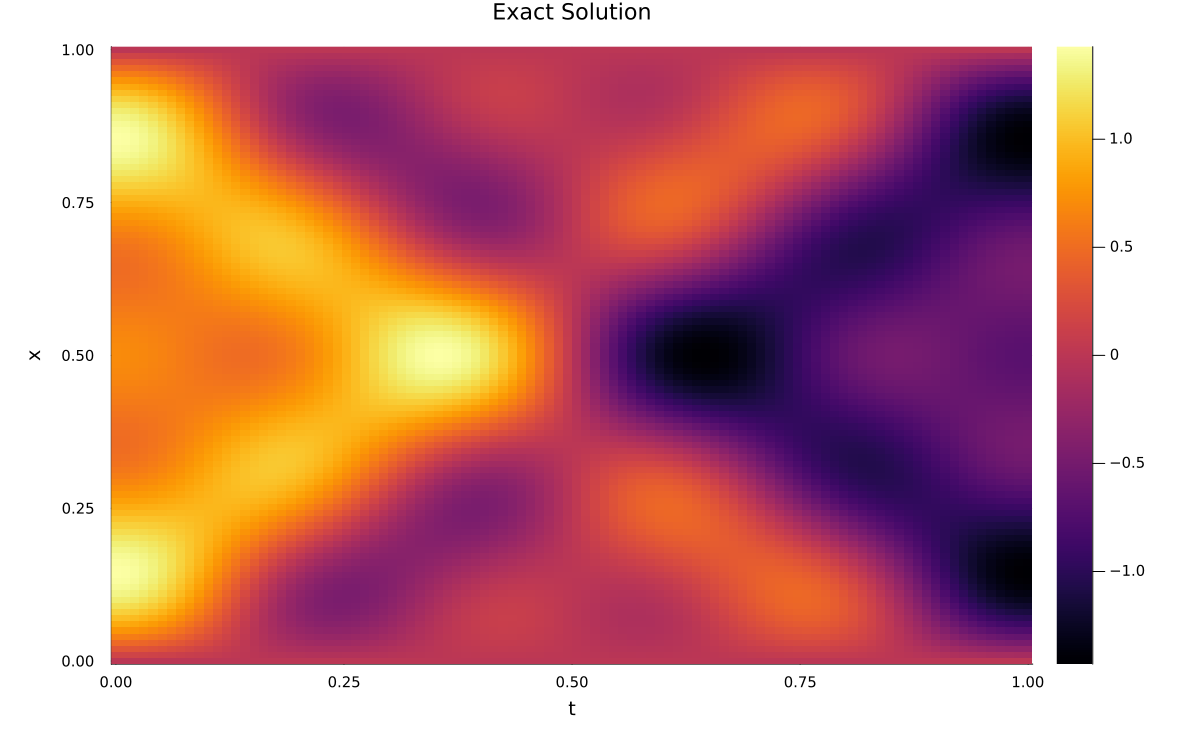}
		\end{subfigure}
		\hfill
		\begin{subfigure}[b]{0.33\textwidth}
			\centering
			\includegraphics[width=\textwidth]{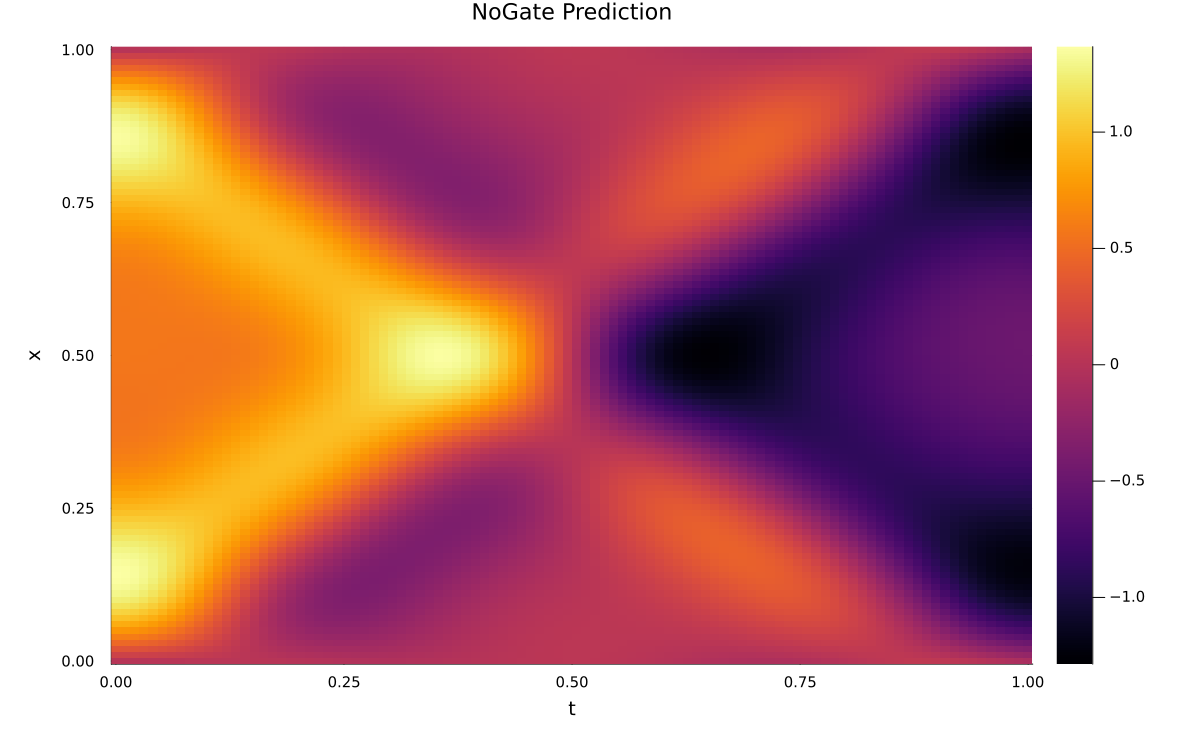}
		\end{subfigure}
		\hfill
		\begin{subfigure}[b]{0.33\textwidth}
			\centering
			\includegraphics[width=\textwidth]{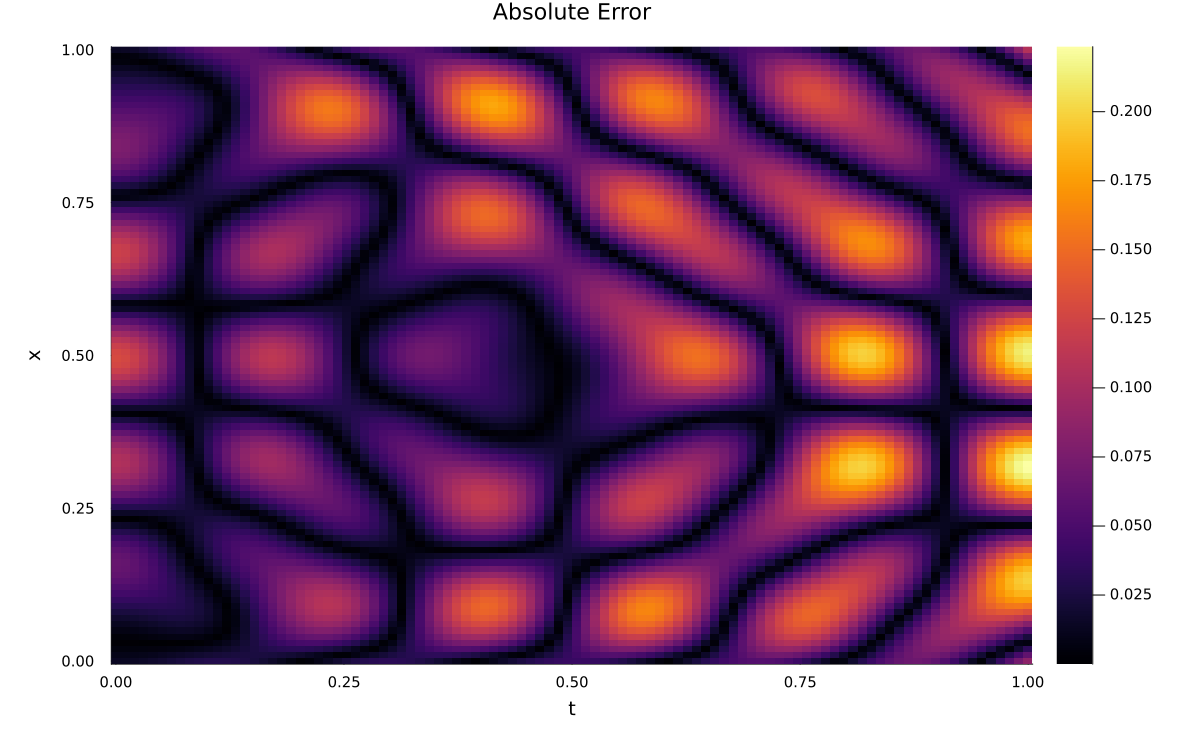}
		\end{subfigure}
		\caption{1D Multimodal Wave, NoGate variant.}
		\label{fig:1D_MM_WAVE_NO_gate}
	\end{figure}

\begin{figure}[t]
		\centering
		\begin{subfigure}[b]{0.48\textwidth}
			\centering
			\includegraphics[width=\textwidth]{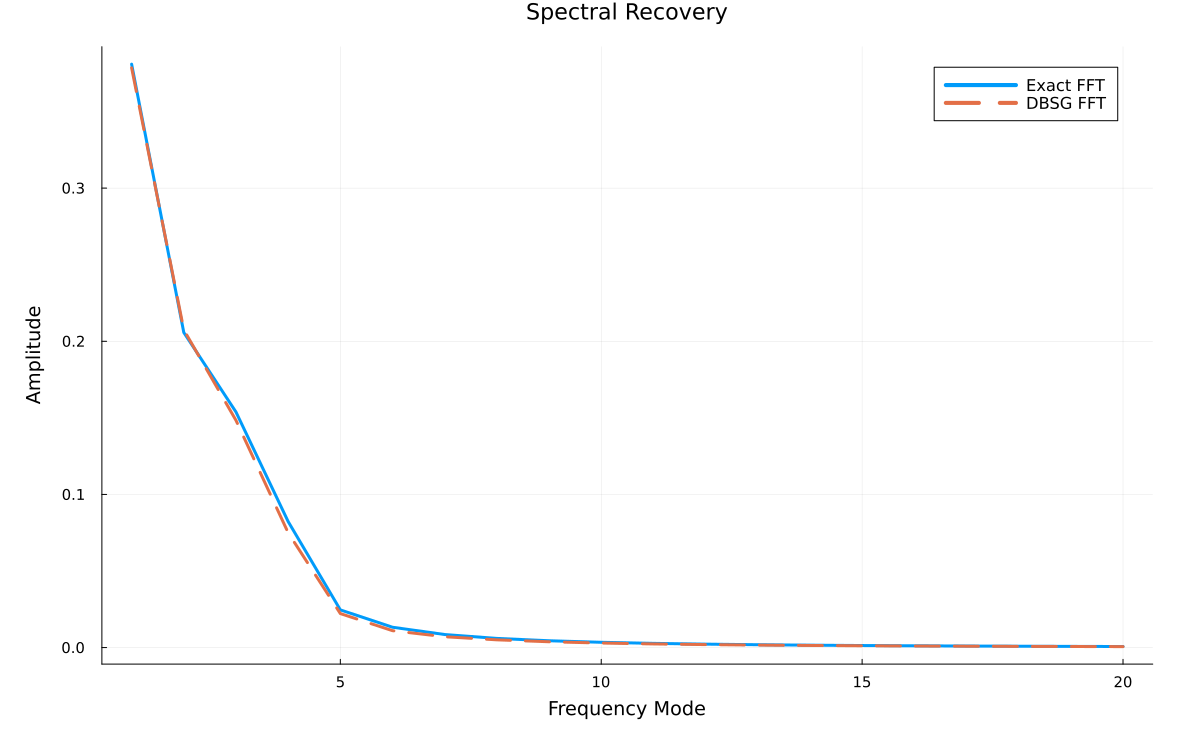}
			\caption{Full (DBSG-PINN)}
		\end{subfigure}
		\hfill
		\begin{subfigure}[b]{0.48\textwidth}
			\centering
			\includegraphics[width=\textwidth]{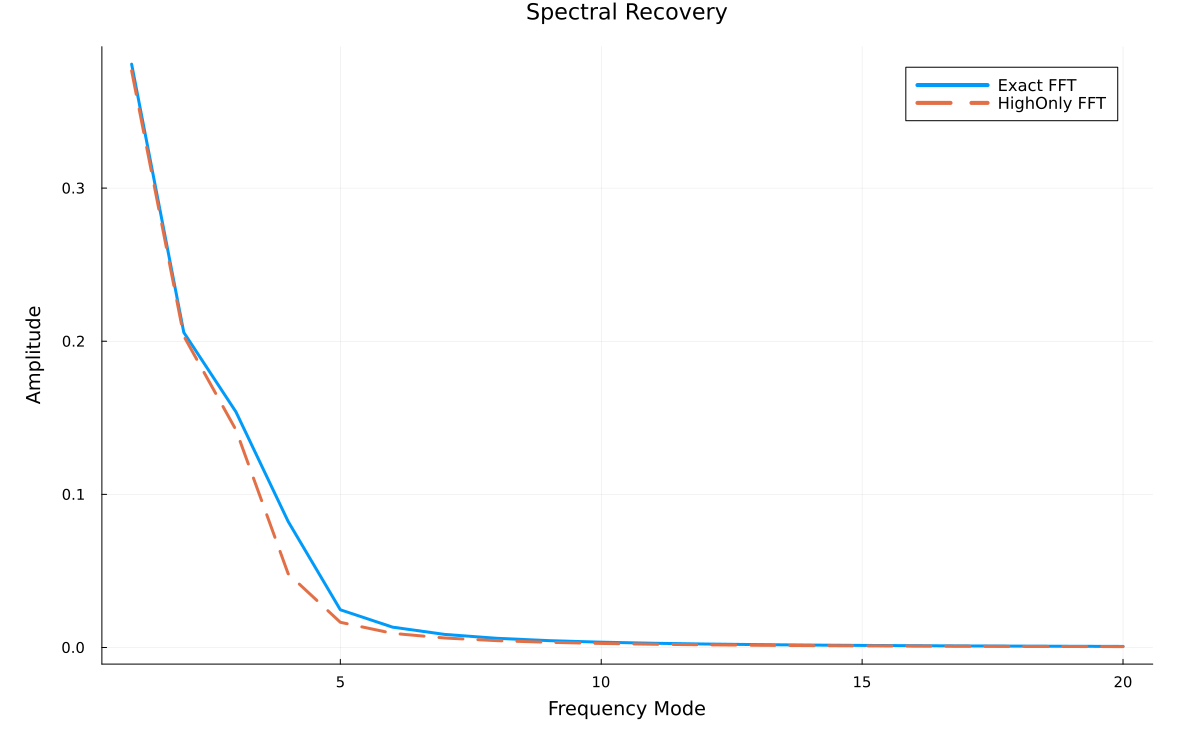}
			\caption{HighOnly}
		\end{subfigure}
		\\[1em]
		\begin{subfigure}[b]{0.48\textwidth}
			\centering
			\includegraphics[width=\textwidth]{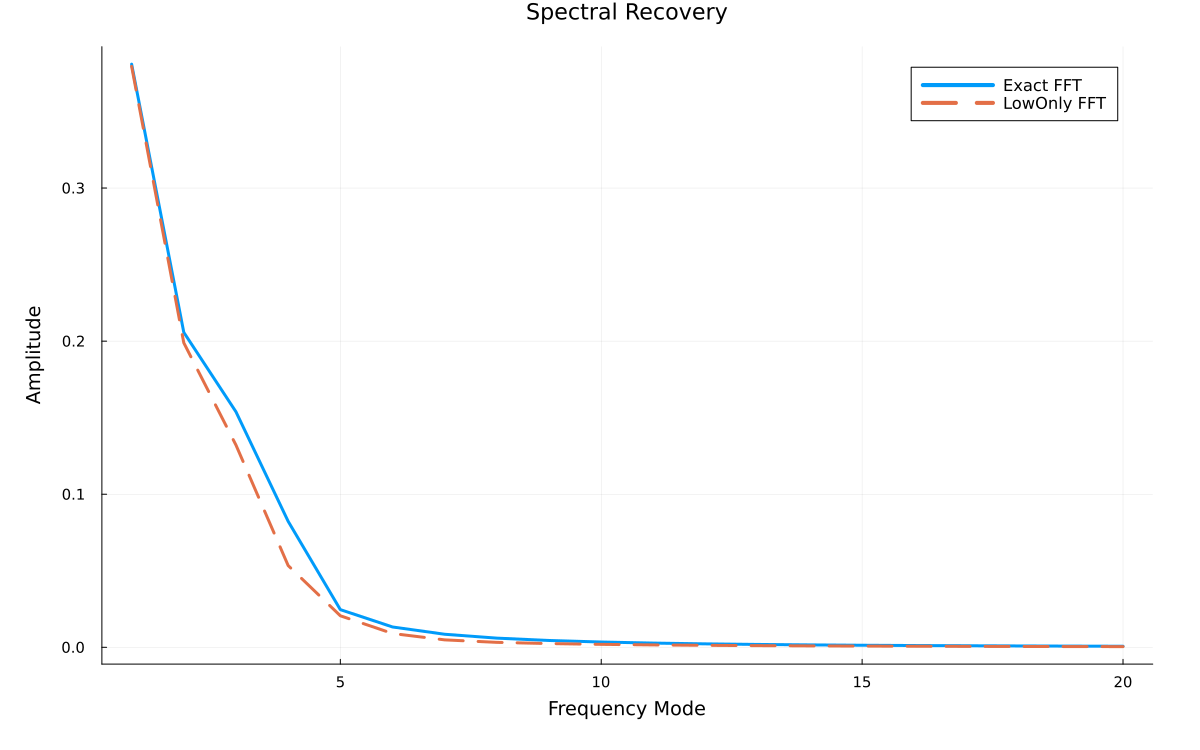}
			\caption{LowOnly}
		\end{subfigure}
		\hfill
		\begin{subfigure}[b]{0.48\textwidth}
			\centering
			\includegraphics[width=\textwidth]{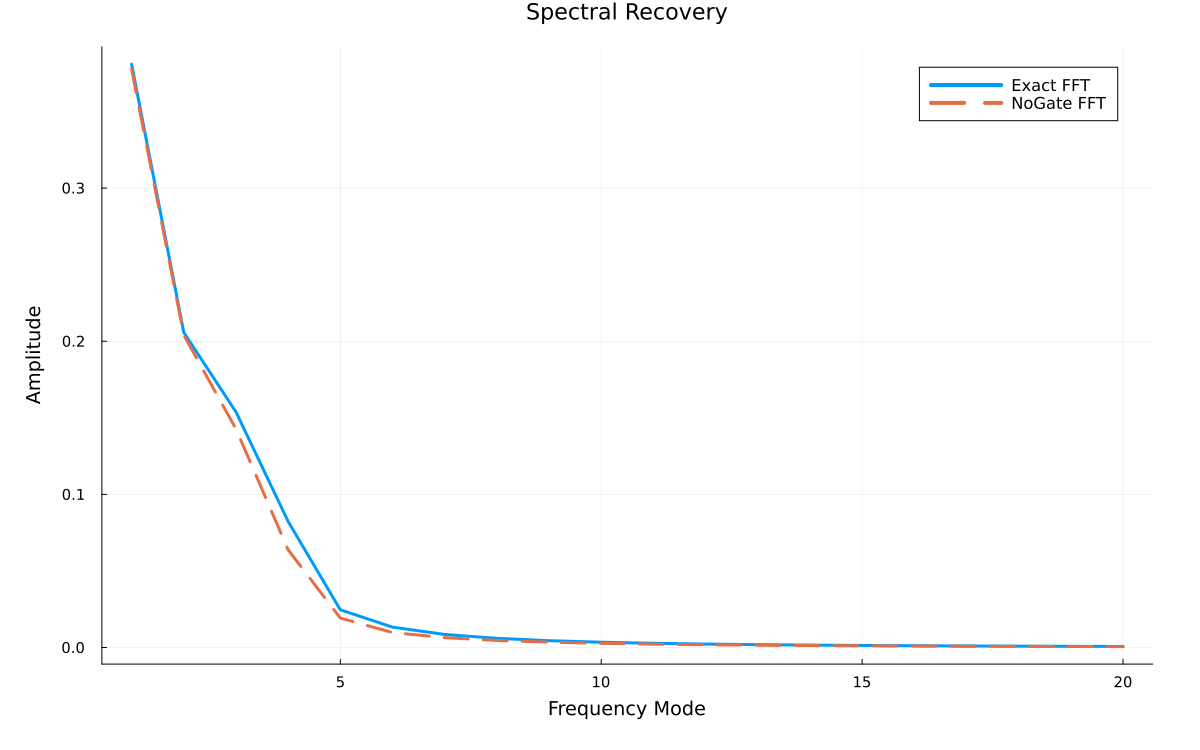}
			\caption{NoGate}
		\end{subfigure}
		\caption{1D Multimodal Wave spectral recovery across ablation variants (FFT amplitude vs.\ frequency mode, single seed).}
		\label{fig:1D_MM_WAVE_Spectral_Recovery}
	\end{figure}

	Figures~\ref{fig:1D_Multimodal_Wave_Full_DBSG}--\ref{fig:1D_MM_WAVE_Spectral_Recovery} show, for each ablation variant, the exact solution, the model's prediction, the pointwise absolute error, and (where shown) the FFT-based spectral recovery on the 1D Multimodal Wave benchmark. All panels share a consistent color scale within a figure to support direct visual comparison; each figure corresponds to a single trained model (Section~\ref{sec:discussion}).
	
	\subsection{1D Burgers Equation: Qualitative Comparison Across Variants}
\begin{figure}[t]
		\centering
		\begin{subfigure}[b]{0.33\textwidth}
			\centering
			\includegraphics[width=\textwidth]{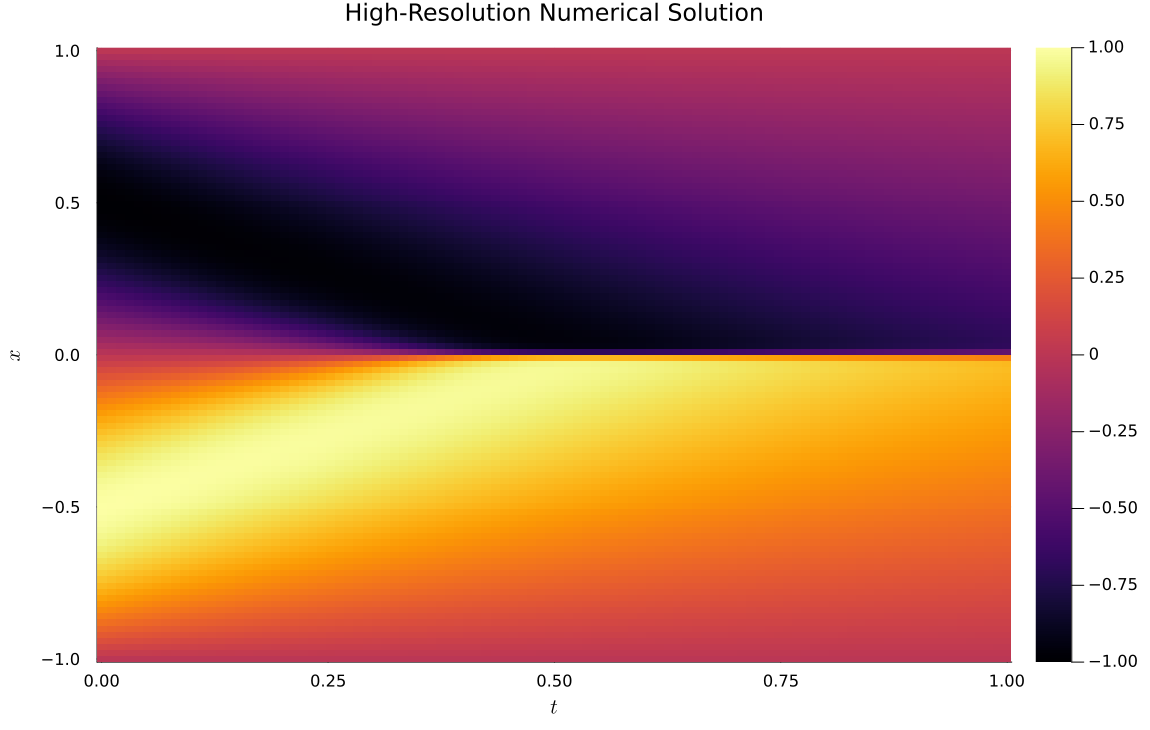}
		\end{subfigure}
		\hfill
		\begin{subfigure}[b]{0.33\textwidth}
			\centering
			\includegraphics[width=\textwidth]{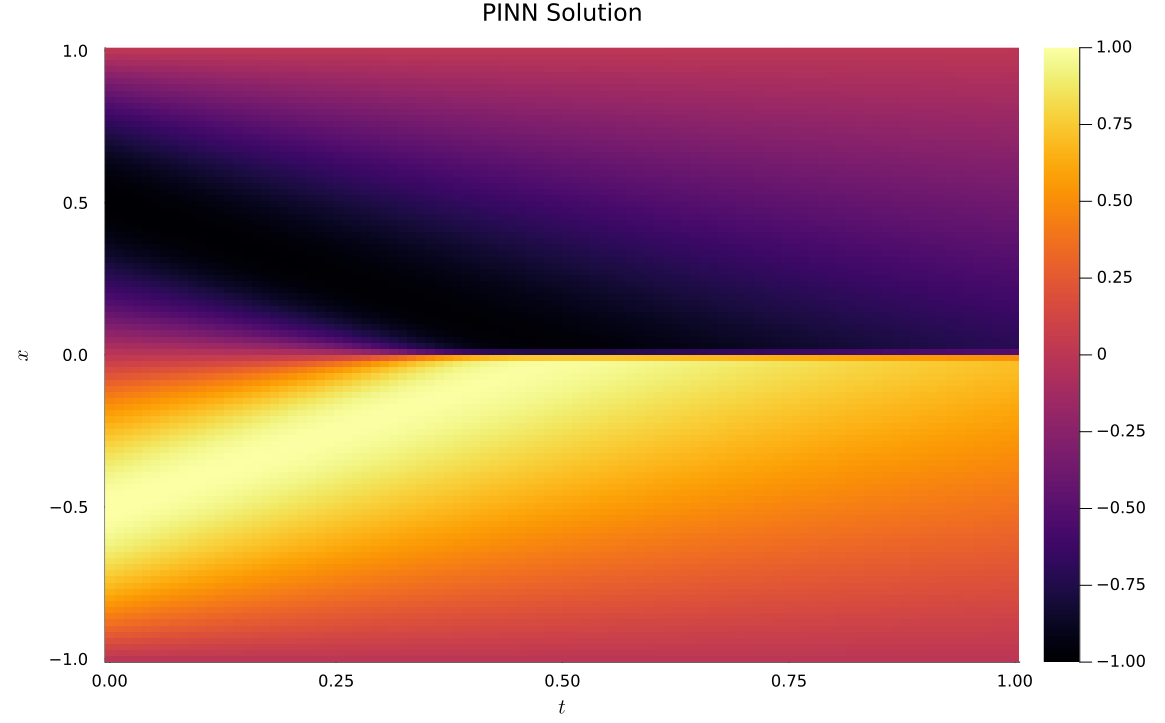}
		\end{subfigure}
		\hfill
		\begin{subfigure}[b]{0.33\textwidth}
			\centering
			\includegraphics[width=\textwidth]{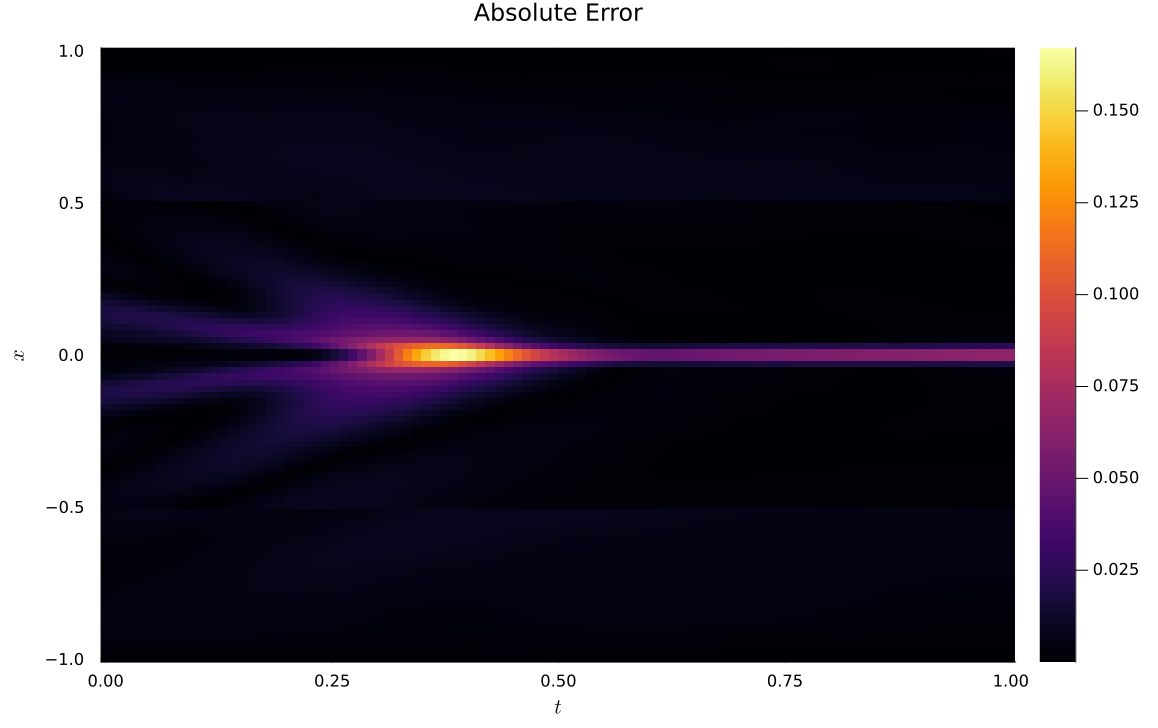}
		\end{subfigure}
		\caption{1D Burgers Equation, Full DBSG-PINN: high-resolution reference solution, prediction, and absolute error.}
		\label{fig:1D_Burgers_Equation_Full_DBSG}
	\end{figure}

\begin{figure}[t]
		\centering
		\begin{subfigure}[b]{0.33\textwidth}
			\centering
			\includegraphics[width=\textwidth]{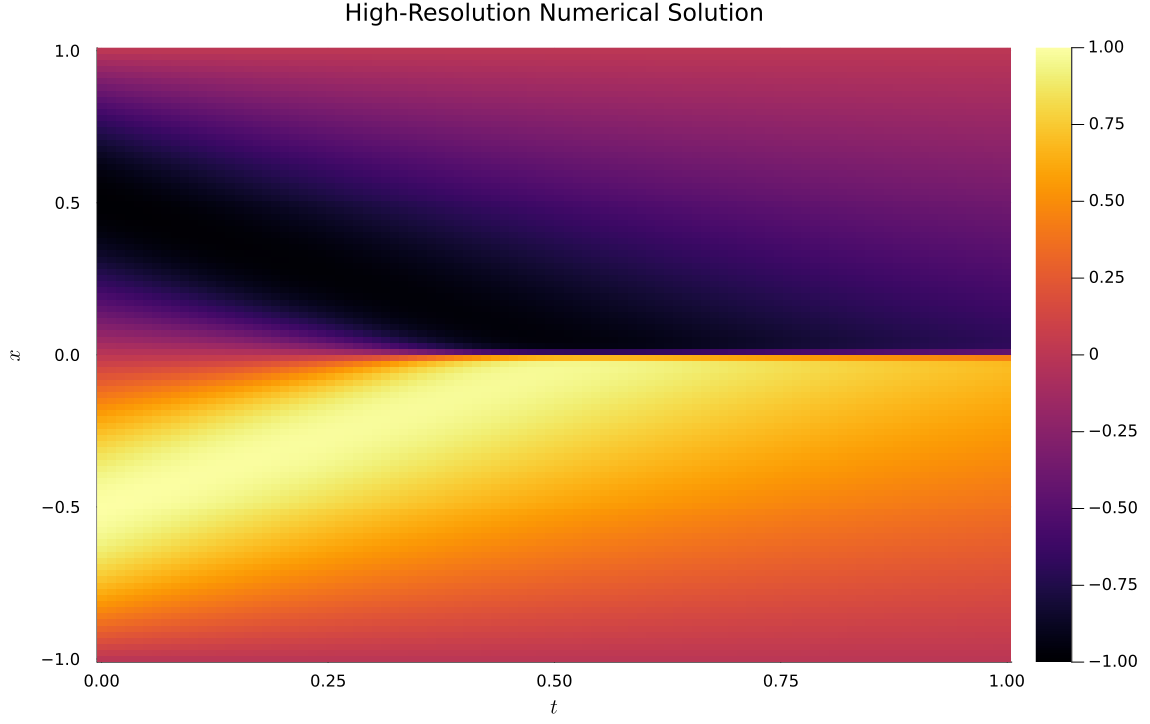}
		\end{subfigure}
		\hfill
		\begin{subfigure}[b]{0.33\textwidth}
			\centering
			\includegraphics[width=\textwidth]{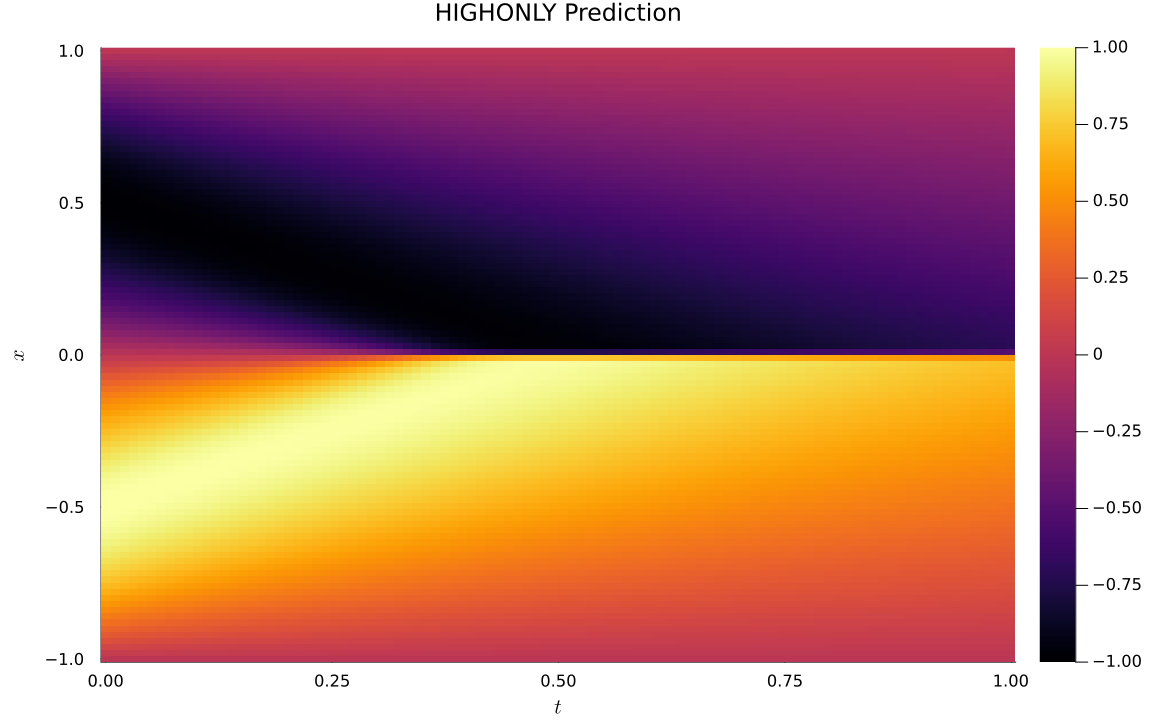}
		\end{subfigure}
		\hfill
		\begin{subfigure}[b]{0.33\textwidth}
			\centering
			\includegraphics[width=\textwidth]{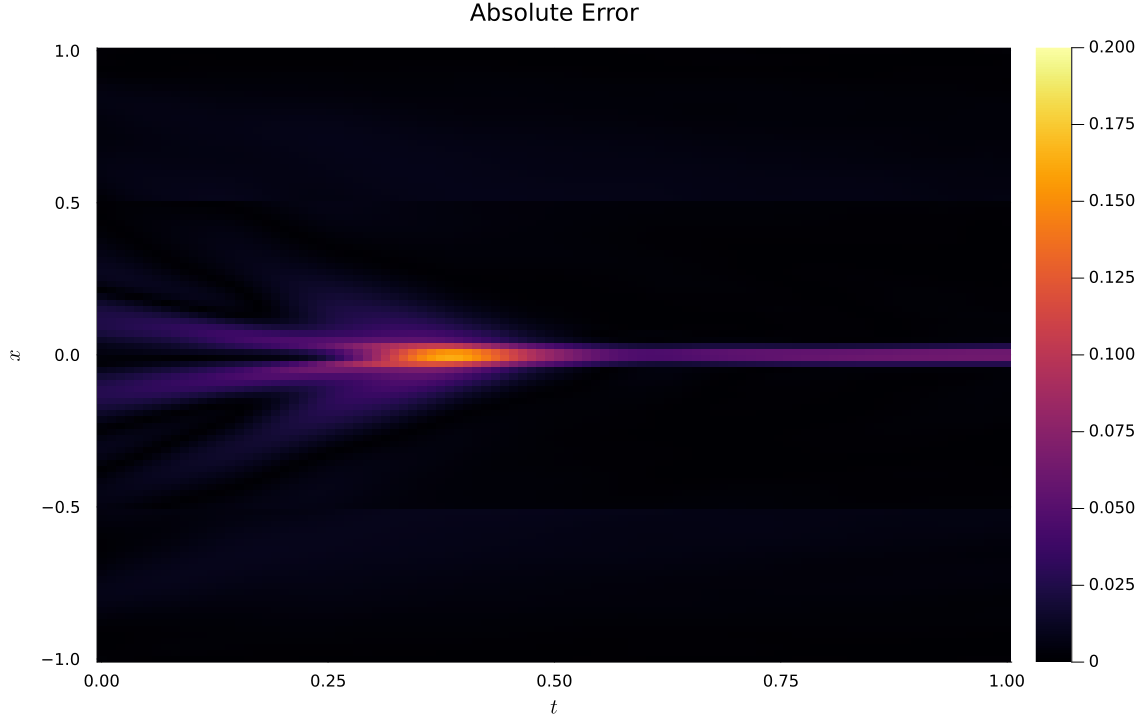}
		\end{subfigure}
		\caption{1D Burgers Equation, HighOnly variant.}
		\label{fig:1D_Burgers_Equation_HIGHONLY}
	\end{figure}

\begin{figure}[t]
		\centering
		\begin{subfigure}[b]{0.33\textwidth}
			\centering
			\includegraphics[width=\textwidth]{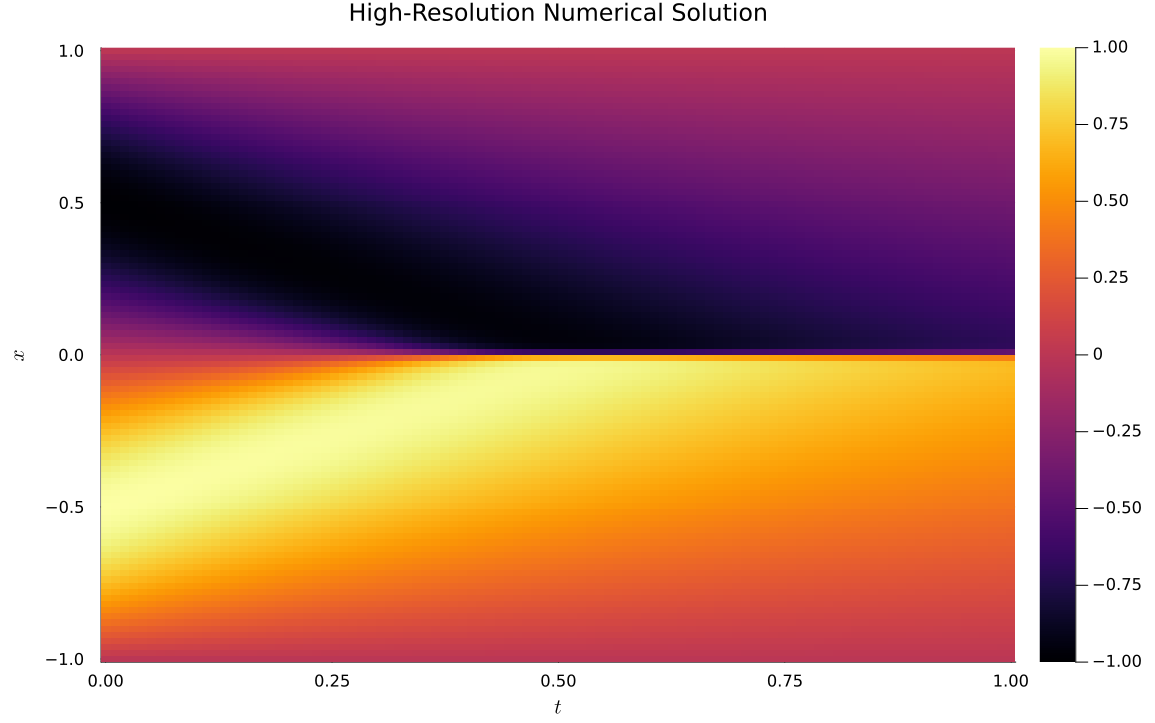}
		\end{subfigure}
		\hfill
		\begin{subfigure}[b]{0.33\textwidth}
			\centering
			\includegraphics[width=\textwidth]{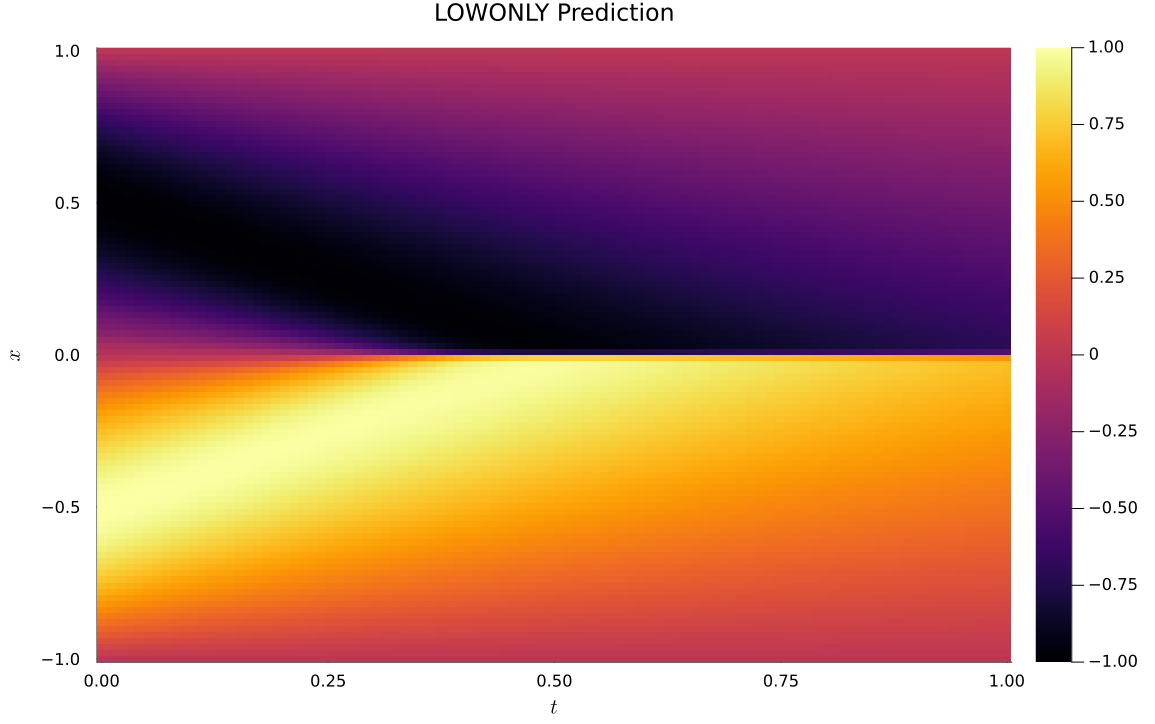}
		\end{subfigure}
		\hfill
		\begin{subfigure}[b]{0.33\textwidth}
			\centering
			\includegraphics[width=\textwidth]{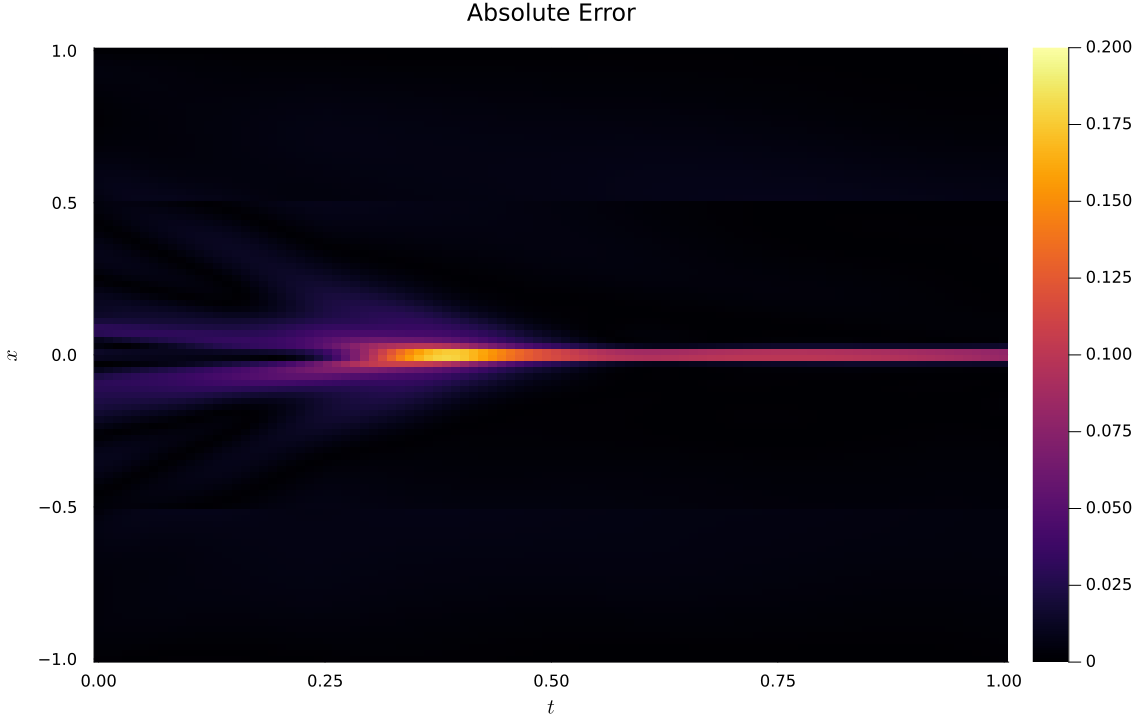}
		\end{subfigure}
		\caption{1D Burgers Equation, LowOnly variant.}
		\label{fig:1D_Burgers_Equation_LOW_ONLY}
	\end{figure}

\begin{figure}[t]
		\centering
		\begin{subfigure}[b]{0.33\textwidth}
			\centering
			\includegraphics[width=\textwidth]{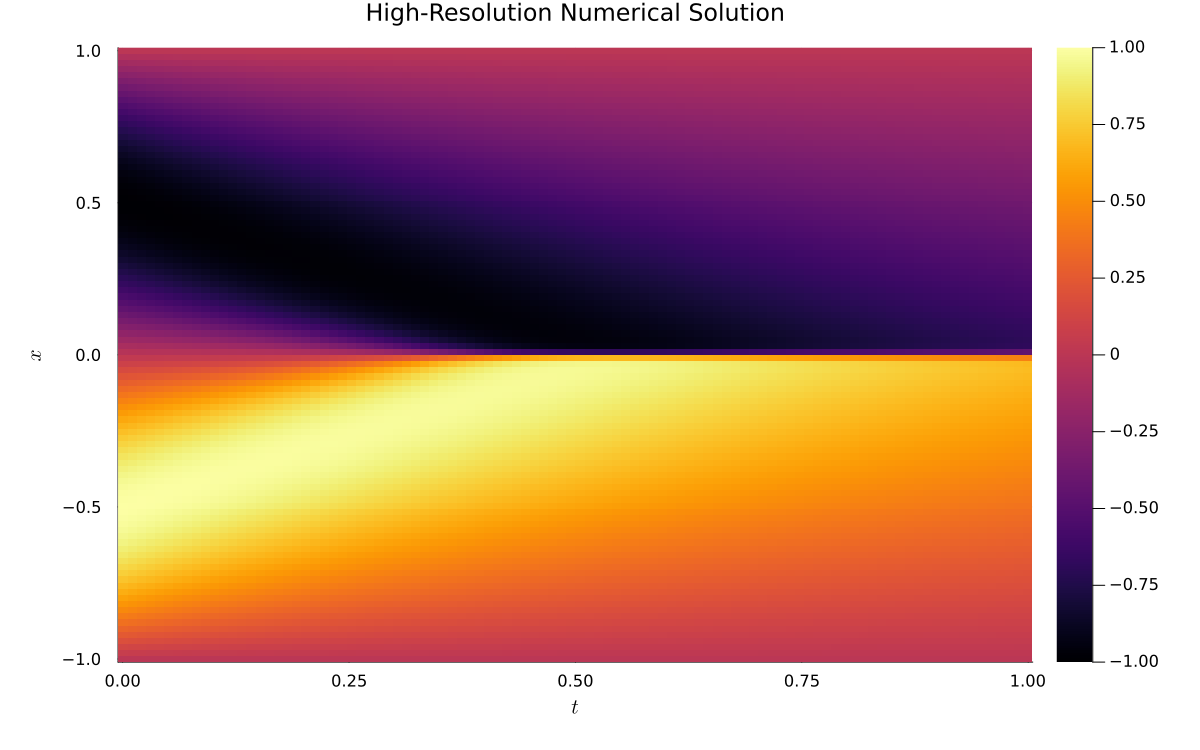}
		\end{subfigure}
		\hfill
		\begin{subfigure}[b]{0.33\textwidth}
			\centering
			\includegraphics[width=\textwidth]{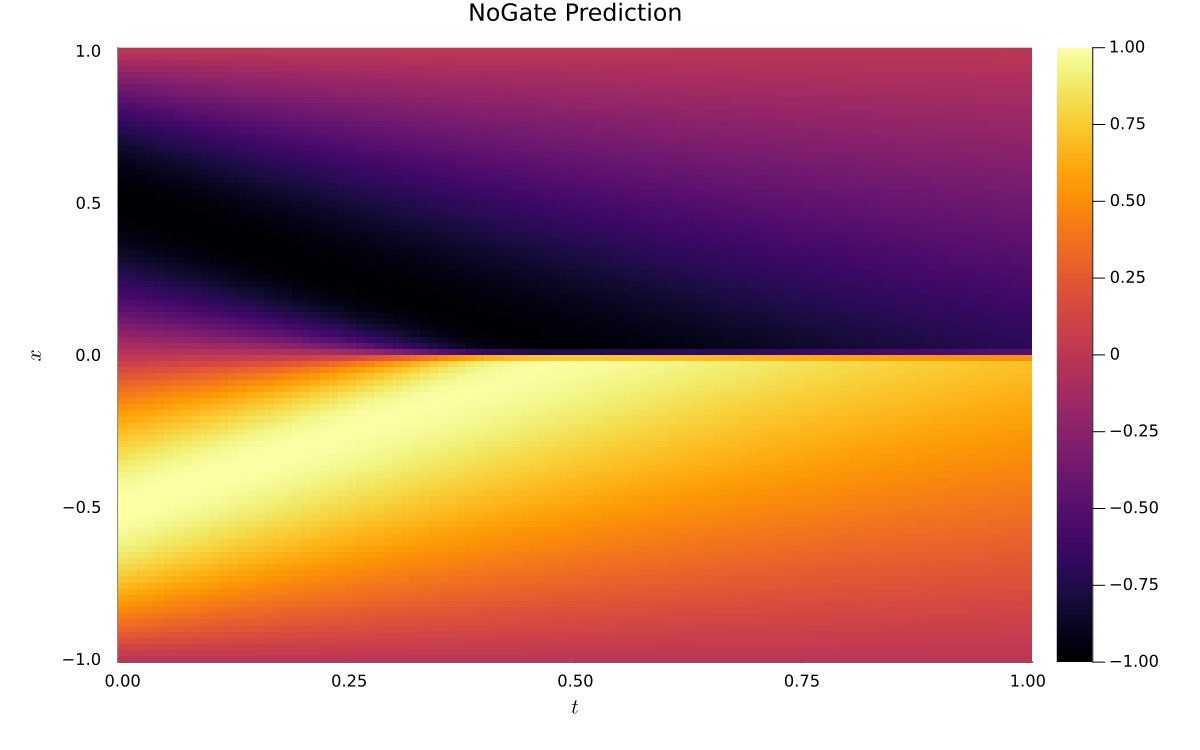}
		\end{subfigure}
		\hfill
		\begin{subfigure}[b]{0.33\textwidth}
			\centering
			\includegraphics[width=\textwidth]{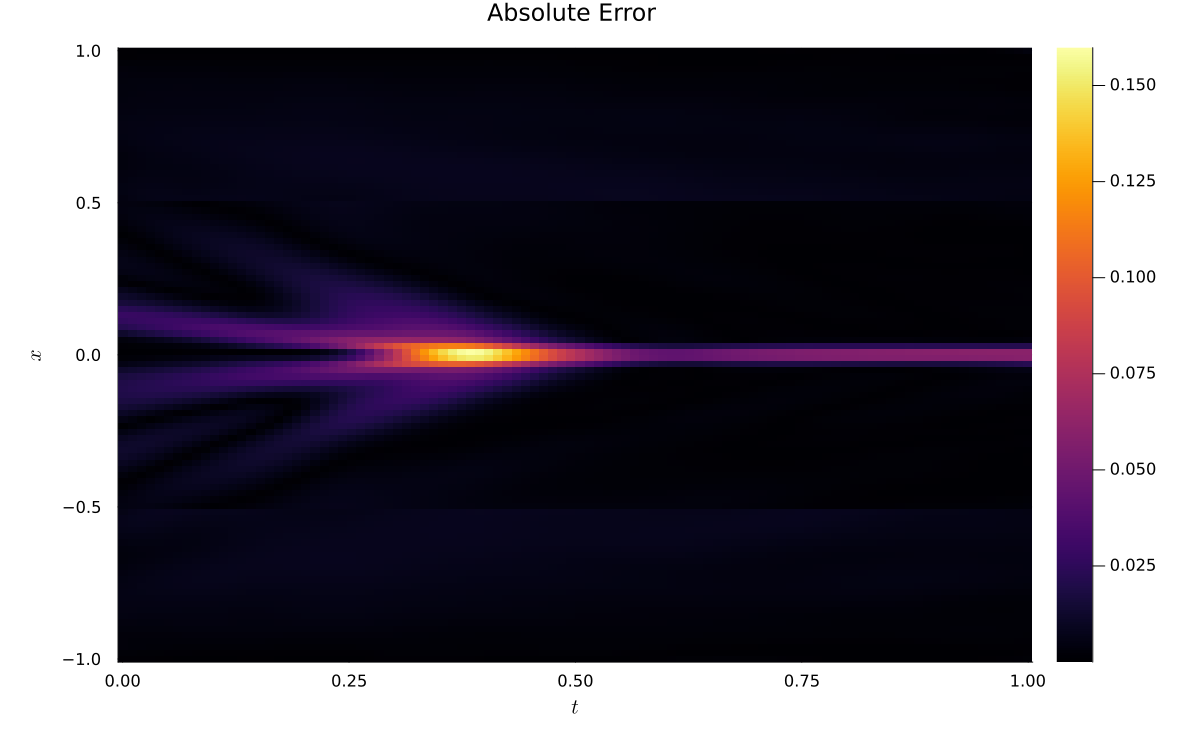}
		\end{subfigure}
		\caption{1D Burgers Equation, NoGate variant.}
		\label{fig:1D_Burgers_Equation_NO_gate}
	\end{figure}

\begin{figure}[t]
		\centering
		\begin{subfigure}[b]{0.48\textwidth}
			\centering
			\includegraphics[width=\textwidth]{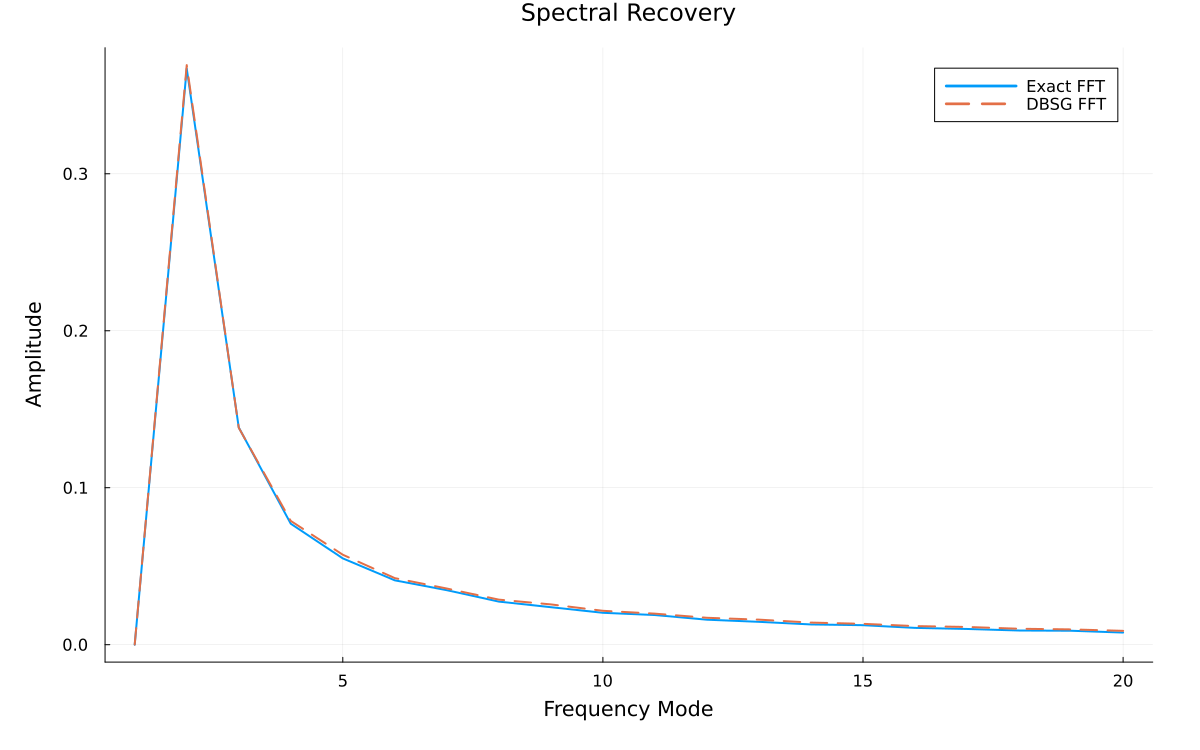}
			\caption{Full (DBSG-PINN)}
		\end{subfigure}
		\hfill
		\begin{subfigure}[b]{0.48\textwidth}
			\centering
			\includegraphics[width=\textwidth]{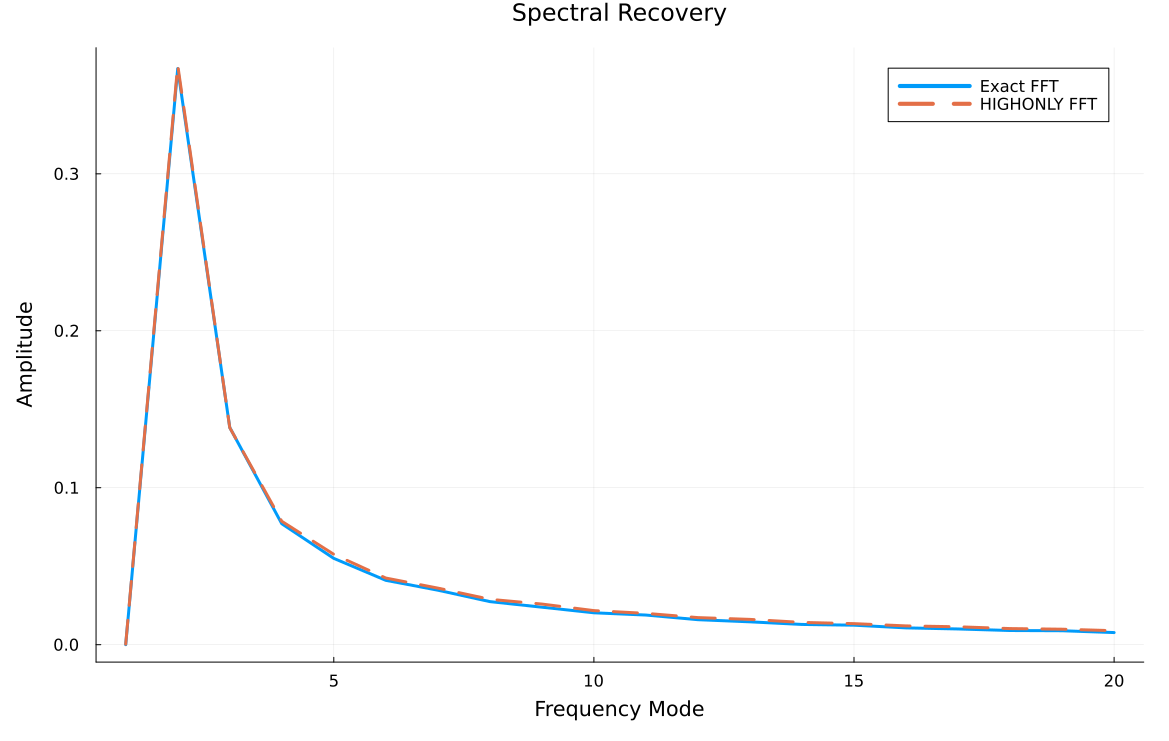}
			\caption{HighOnly}
		\end{subfigure}
		\\[1em]
		\begin{subfigure}[b]{0.48\textwidth}
			\centering
			\includegraphics[width=\textwidth]{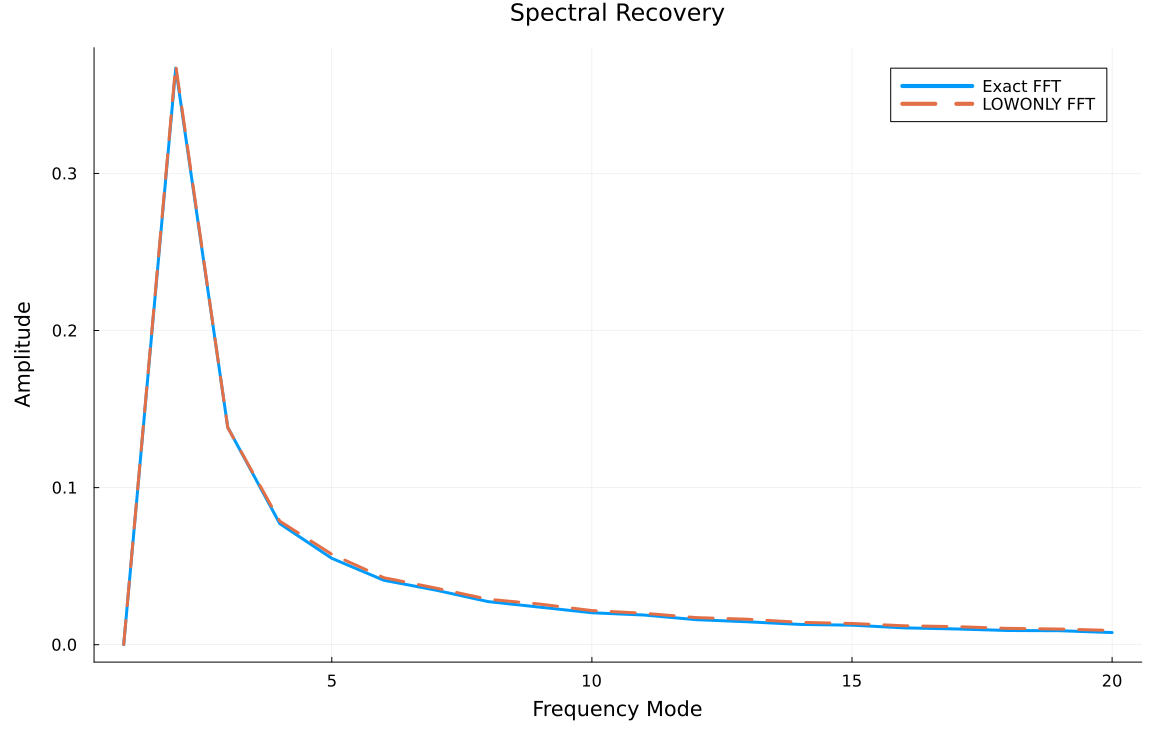}
			\caption{LowOnly}
		\end{subfigure}
		\hfill
		\begin{subfigure}[b]{0.48\textwidth}
			\centering
			\includegraphics[width=\textwidth]{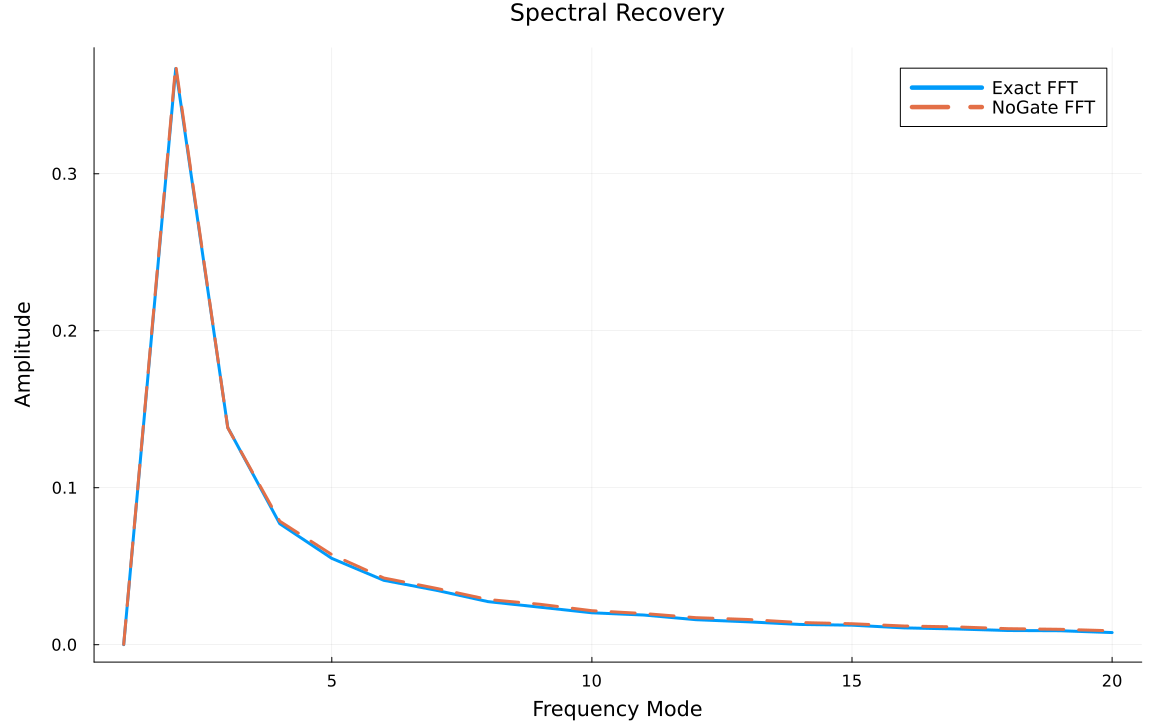}
			\caption{NoGate}
		\end{subfigure}
		\caption{1D Burgers Equation spectral recovery across ablation variants (FFT amplitude vs.\ frequency mode, single seed).}
		\label{fig:1D_Burgers_Equation_Spectral_Recovery}
	\end{figure}

	Figures~\ref{fig:1D_Burgers_Equation_Full_DBSG}--\ref{fig:1D_Burgers_Equation_Spectral_Recovery} show the same reference-solution / prediction / absolute-error / spectral-recovery comparison for the 1D Burgers benchmark (reference here meaning the high-resolution numerical solution described in Section~\ref{sec:results}, not a closed-form exact solution).
	
	\subsection{1D Allen--Cahn Equation: Qualitative Comparison Across Variants}
\begin{figure}[t]
		\centering
		\begin{subfigure}[b]{0.33\textwidth}
			\centering
			\includegraphics[width=\textwidth]{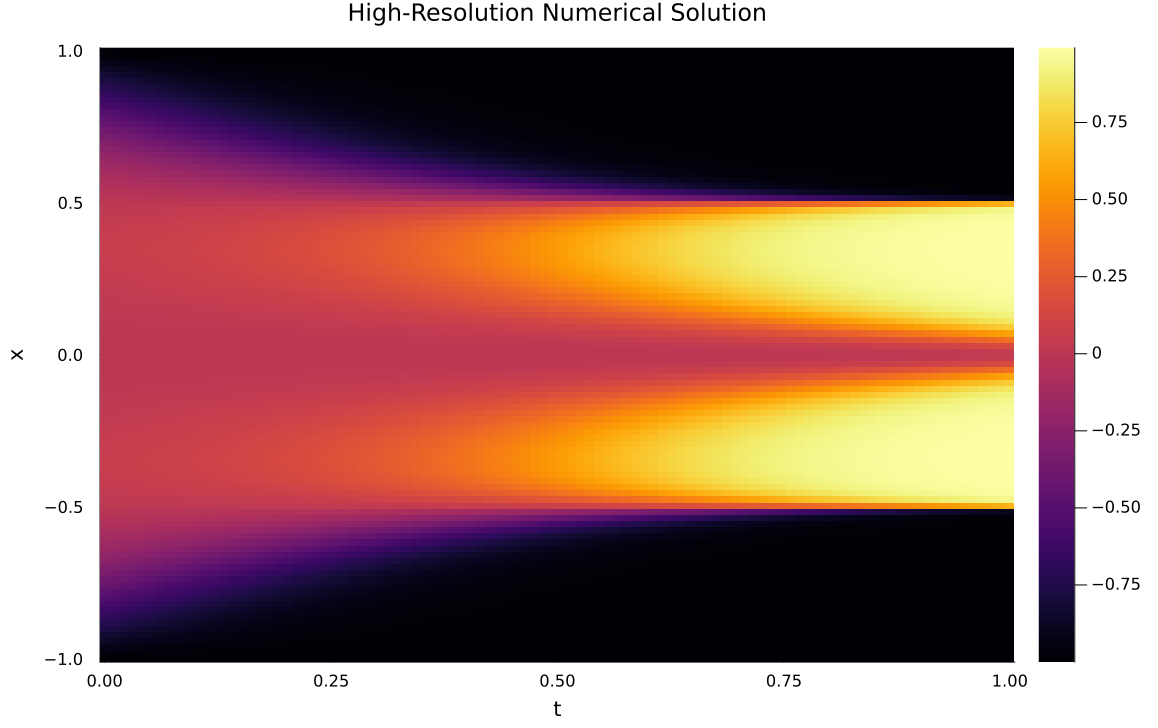}
		\end{subfigure}
		\hfill
		\begin{subfigure}[b]{0.33\textwidth}
			\centering
			\includegraphics[width=\textwidth]{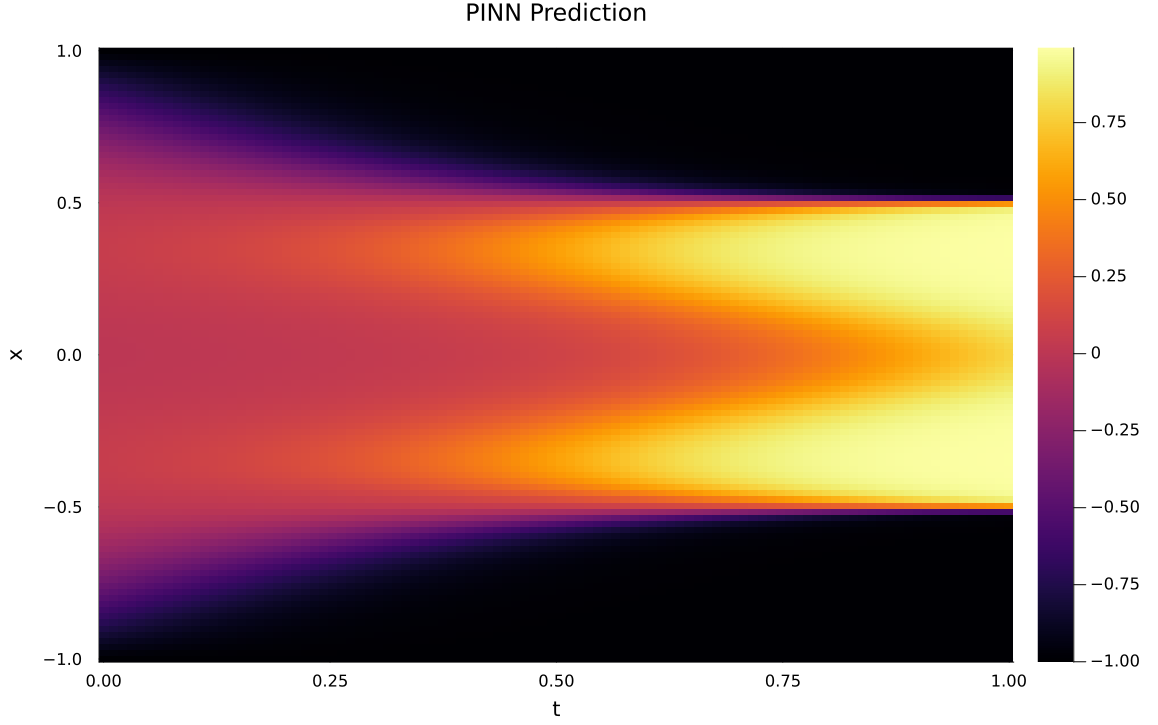}
		\end{subfigure}
		\hfill
		\begin{subfigure}[b]{0.33\textwidth}
			\centering
			\includegraphics[width=\textwidth]{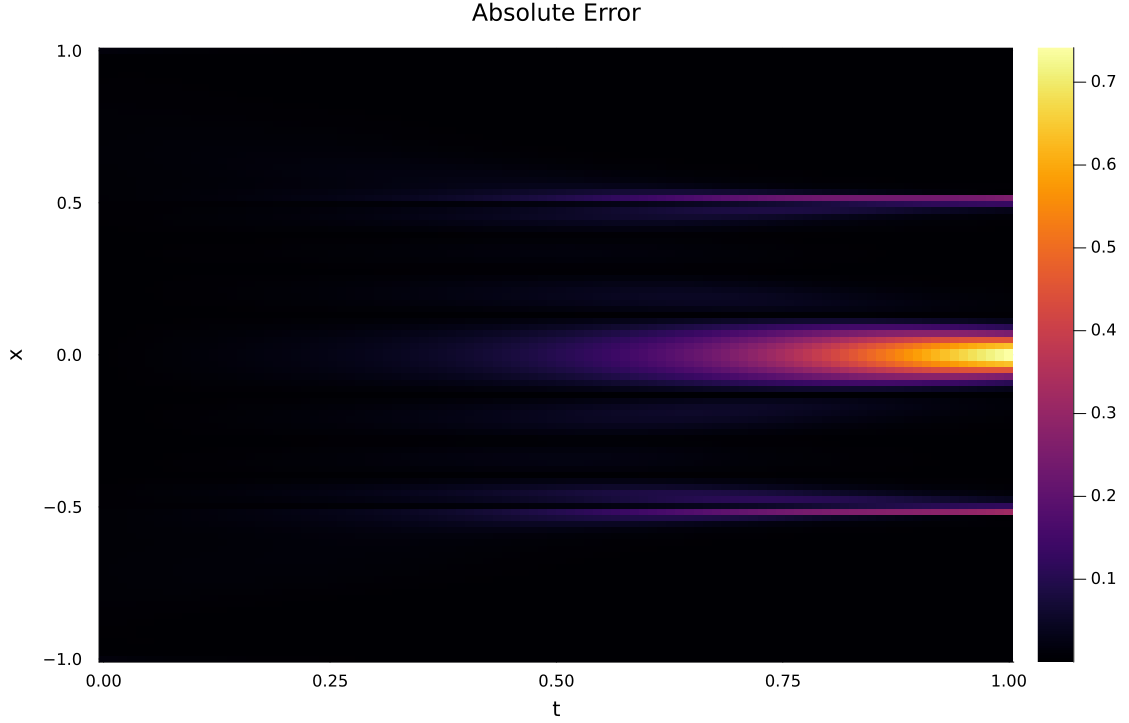}
		\end{subfigure}
		\caption{1D Allen--Cahn Equation, Full DBSG-PINN: high-resolution reference solution, prediction, and absolute error.}
		\label{fig:1D_Allen_Cahn_Equation_Full_DBSG}
	\end{figure}

\begin{figure}[t]
		\centering
		\begin{subfigure}[b]{0.33\textwidth}
			\centering
			\includegraphics[width=\textwidth]{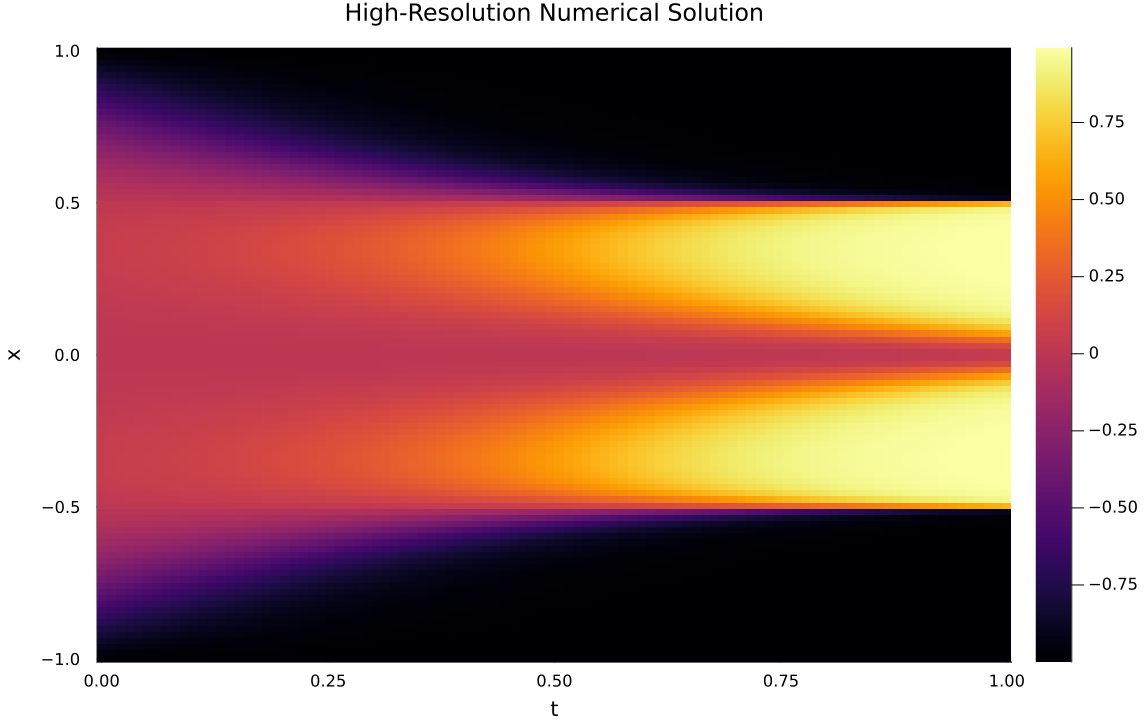}
		\end{subfigure}
		\hfill
		\begin{subfigure}[b]{0.33\textwidth}
			\centering
			\includegraphics[width=\textwidth]{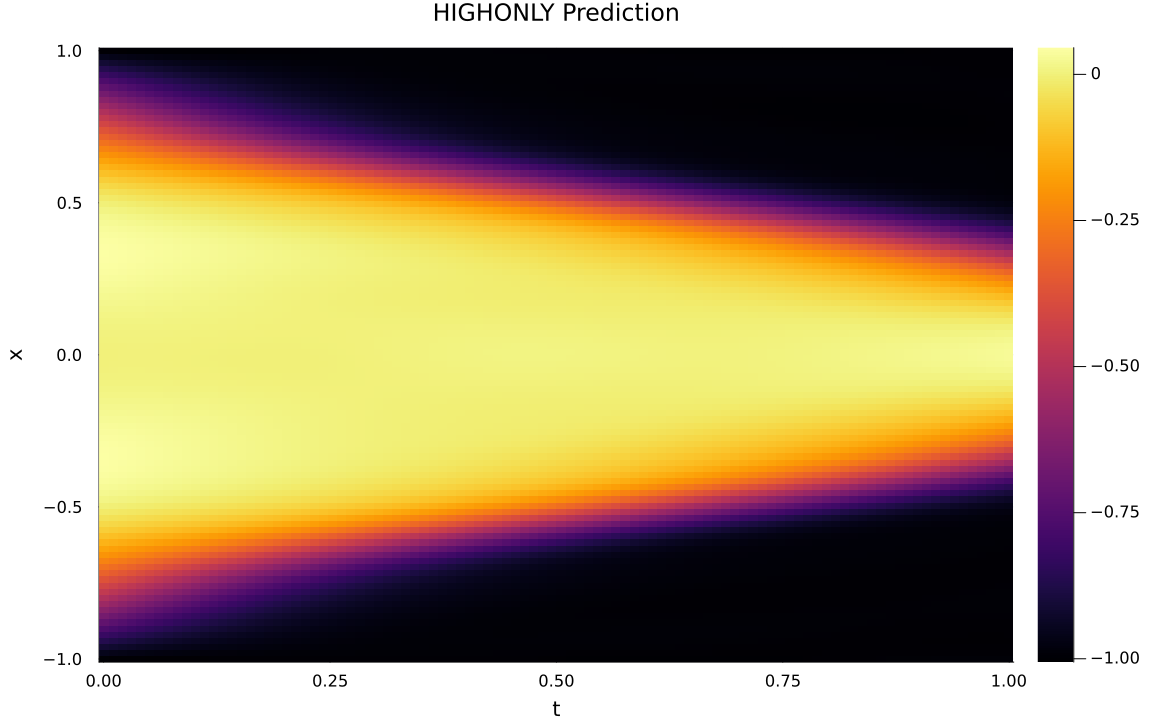}
		\end{subfigure}
		\hfill
		\begin{subfigure}[b]{0.33\textwidth}
			\centering
			\includegraphics[width=\textwidth]{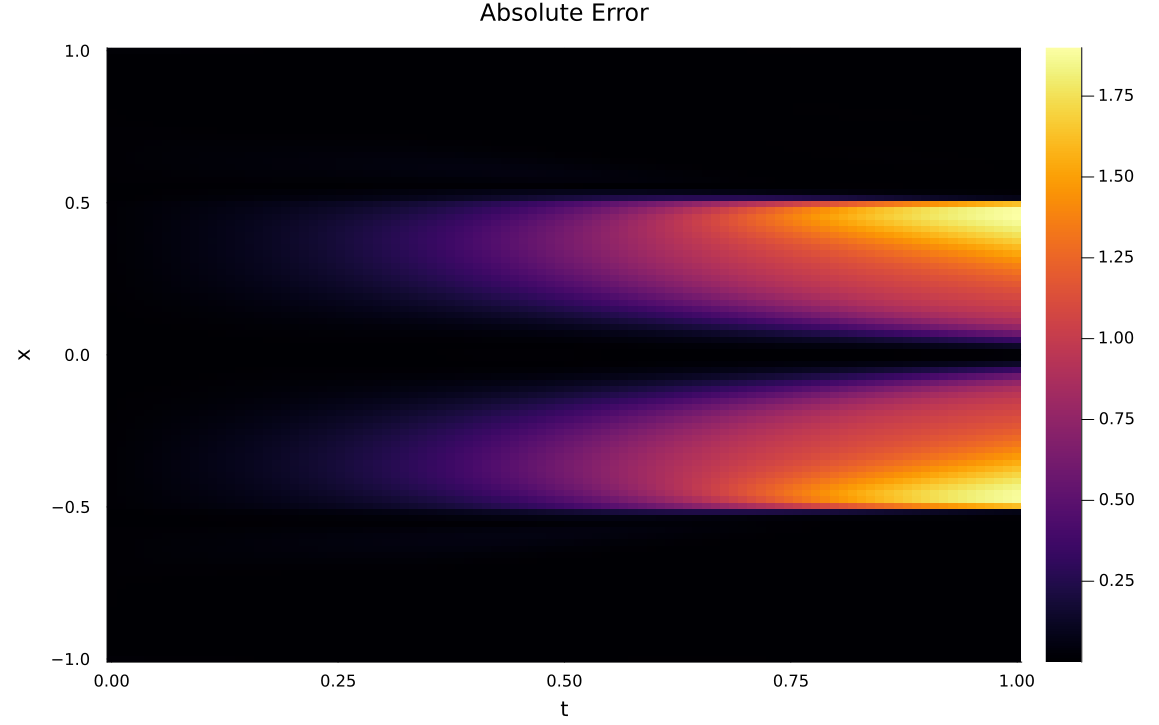}
		\end{subfigure}
		\caption{1D Allen--Cahn Equation, HighOnly variant.}
		\label{fig:1D_Allen_Cahn_Equation_HIGH_ONLY}
	\end{figure}

\begin{figure}[t]
		\centering
		\begin{subfigure}[b]{0.33\textwidth}
			\centering
			\includegraphics[width=\textwidth]{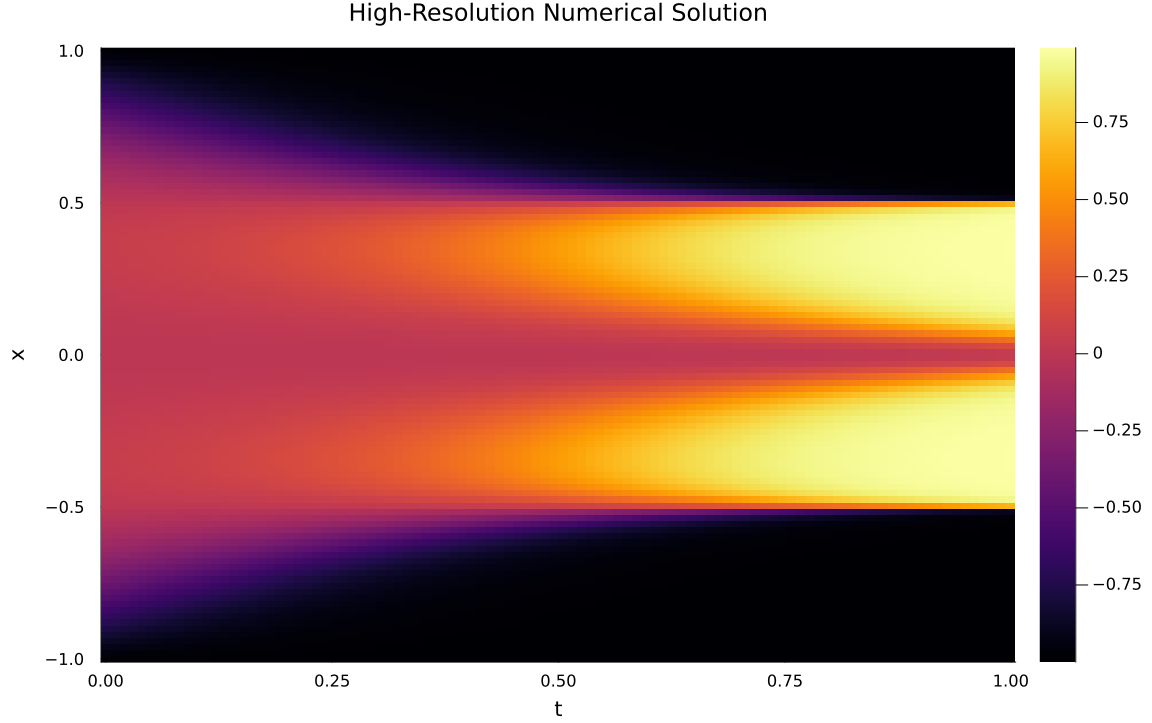}
		\end{subfigure}
		\hfill
		\begin{subfigure}[b]{0.33\textwidth}
			\centering
			\includegraphics[width=\textwidth]{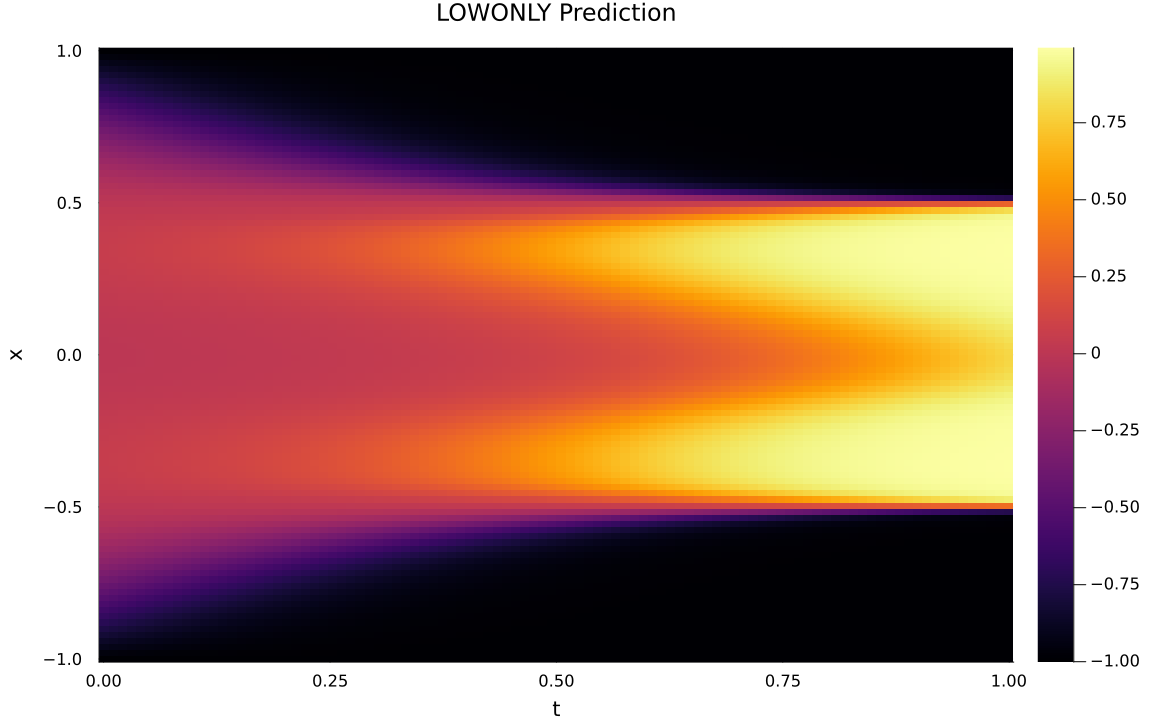}
		\end{subfigure}
		\hfill
		\begin{subfigure}[b]{0.33\textwidth}
			\centering
			\includegraphics[width=\textwidth]{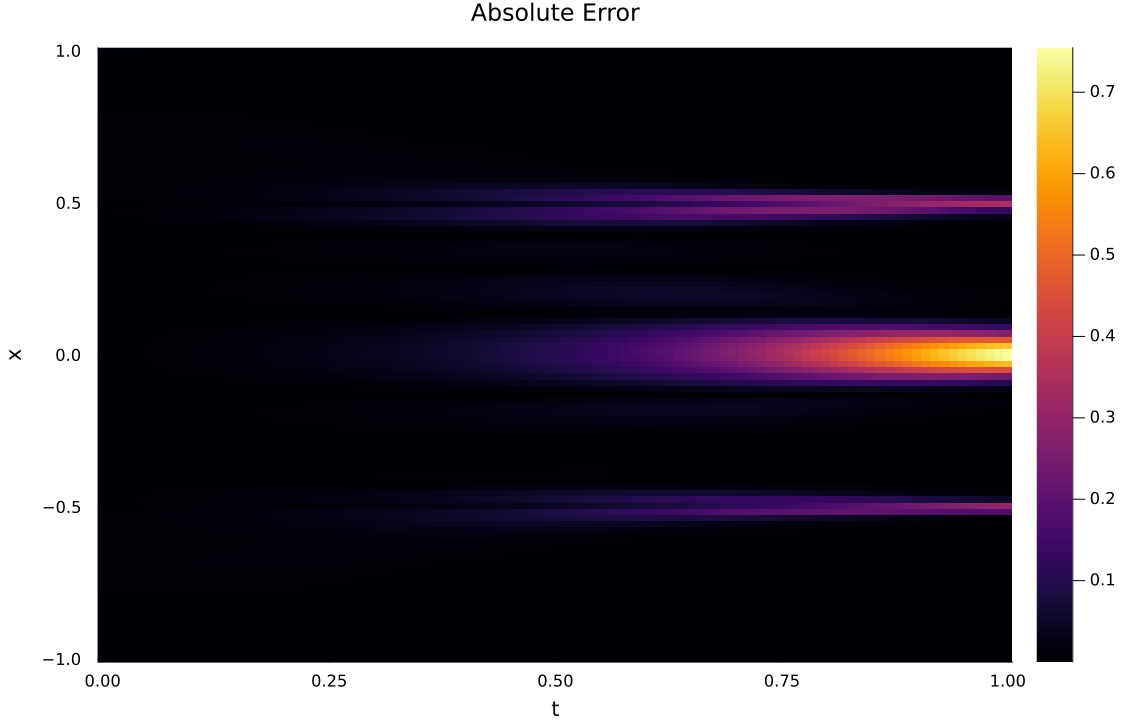}
		\end{subfigure}
		\caption{1D Allen--Cahn Equation, LowOnly variant.}
		\label{fig:1D_Allen_Cahn_Equation_LOW_ONLY}
	\end{figure}

\begin{figure}[t]
		\centering
		\begin{subfigure}[b]{0.33\textwidth}
			\centering
			\includegraphics[width=\textwidth]{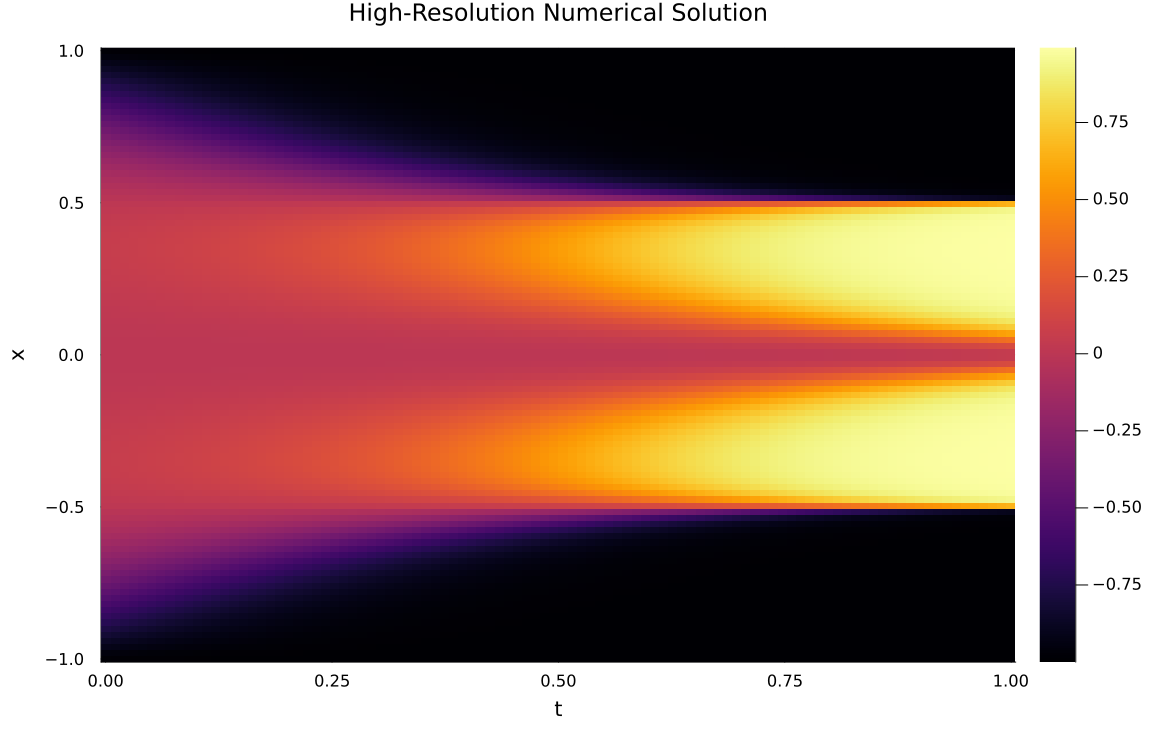}
		\end{subfigure}
		\hfill
		\begin{subfigure}[b]{0.33\textwidth}
			\centering
			\includegraphics[width=\textwidth]{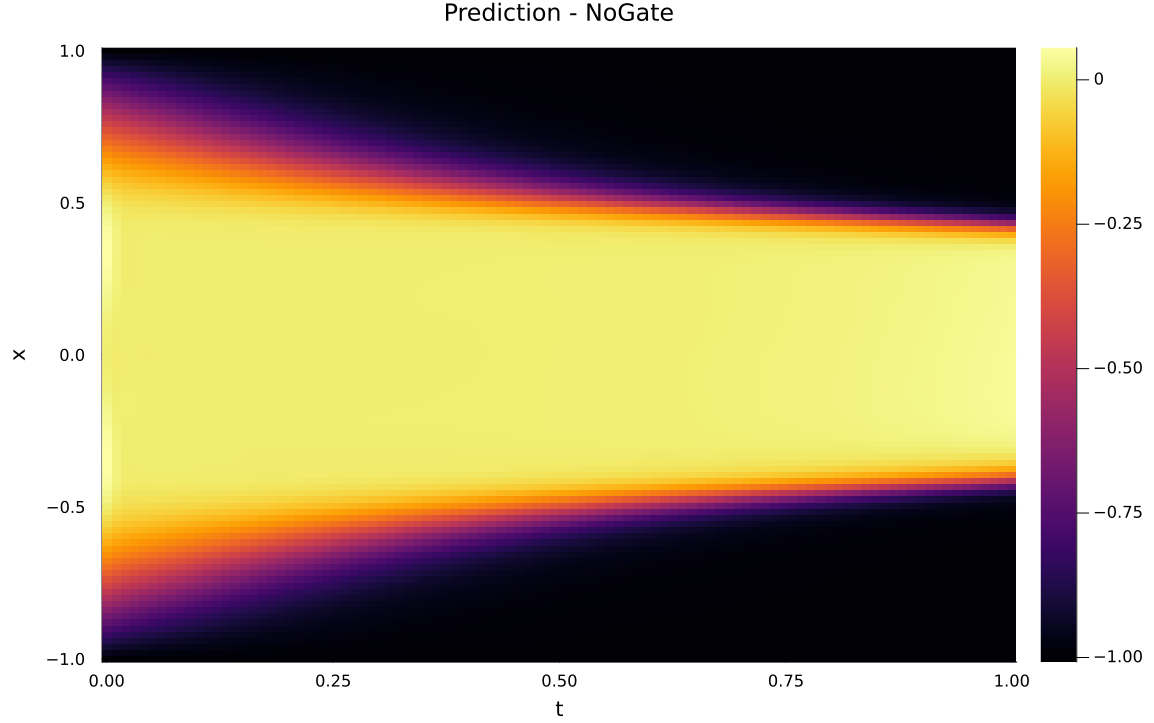}
		\end{subfigure}
		\hfill
		\begin{subfigure}[b]{0.33\textwidth}
			\centering
			\includegraphics[width=\textwidth]{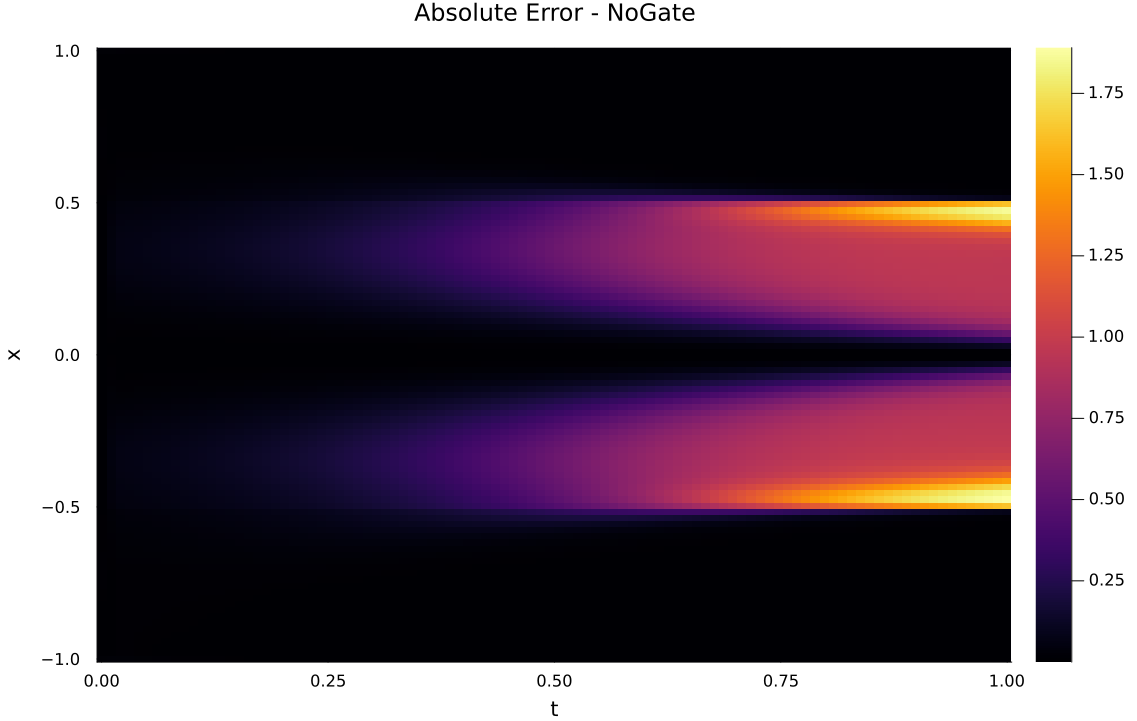}
		\end{subfigure}
		\caption{1D Allen--Cahn Equation, NoGate variant.}
		\label{fig:1D_Allen_Cahn_NO_gate}
	\end{figure}

\begin{figure}[t]
		\centering
		\begin{subfigure}[b]{0.48\textwidth}
			\centering
			\includegraphics[width=\textwidth]{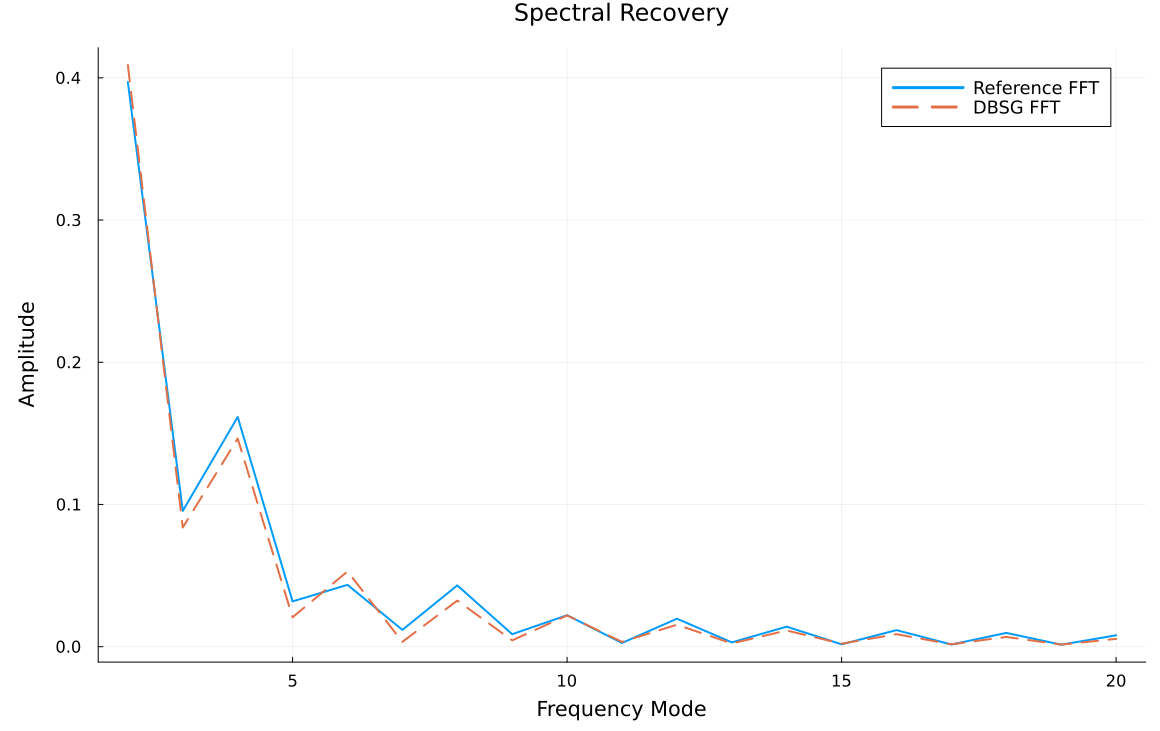}
			\caption{Full (DBSG-PINN)}
		\end{subfigure}
		\hfill
		\begin{subfigure}[b]{0.48\textwidth}
			\centering
			\includegraphics[width=\textwidth]{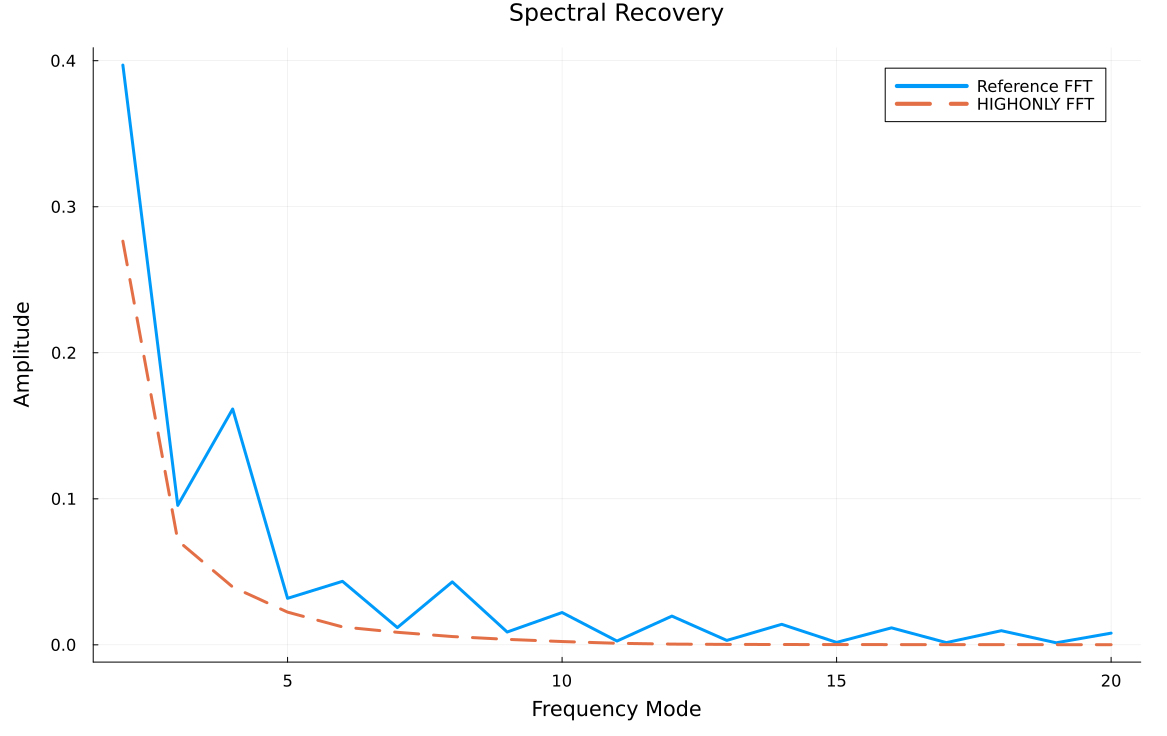}
			\caption{HighOnly}
		\end{subfigure}
		\\[1em]
		\begin{subfigure}[b]{0.48\textwidth}
			\centering
			\includegraphics[width=\textwidth]{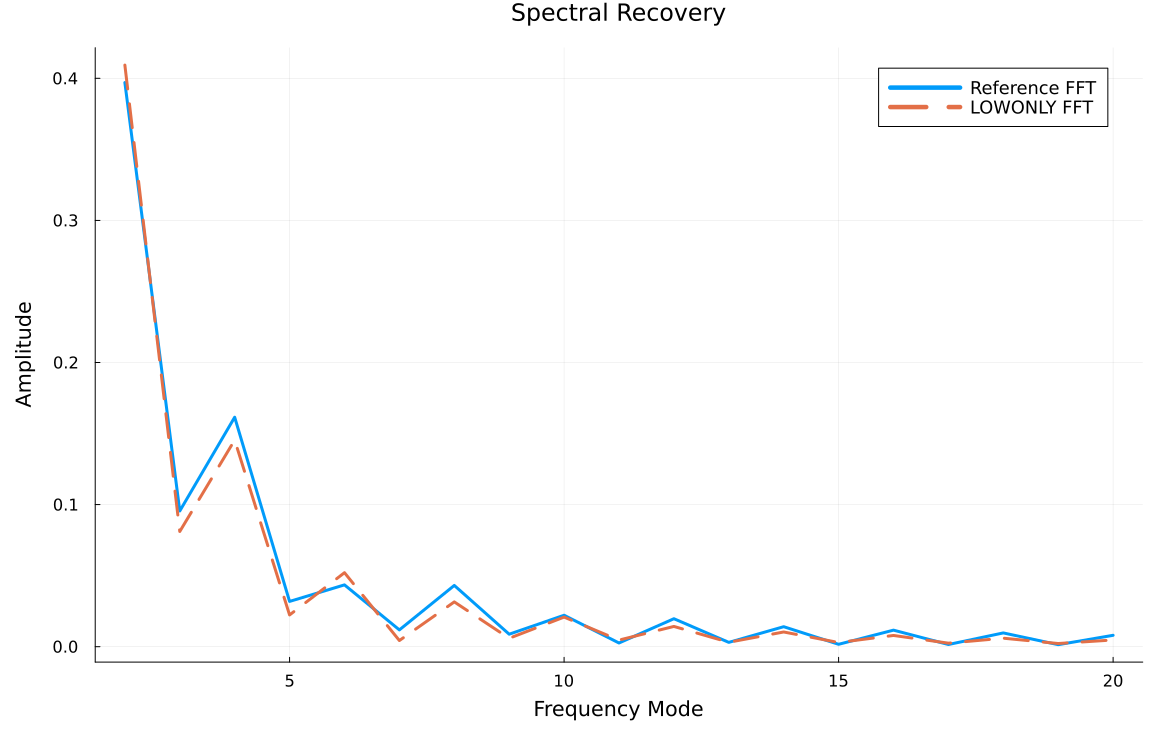}
			\caption{LowOnly}
		\end{subfigure}
		\hfill
		\begin{subfigure}[b]{0.48\textwidth}
			\centering
			\includegraphics[width=\textwidth]{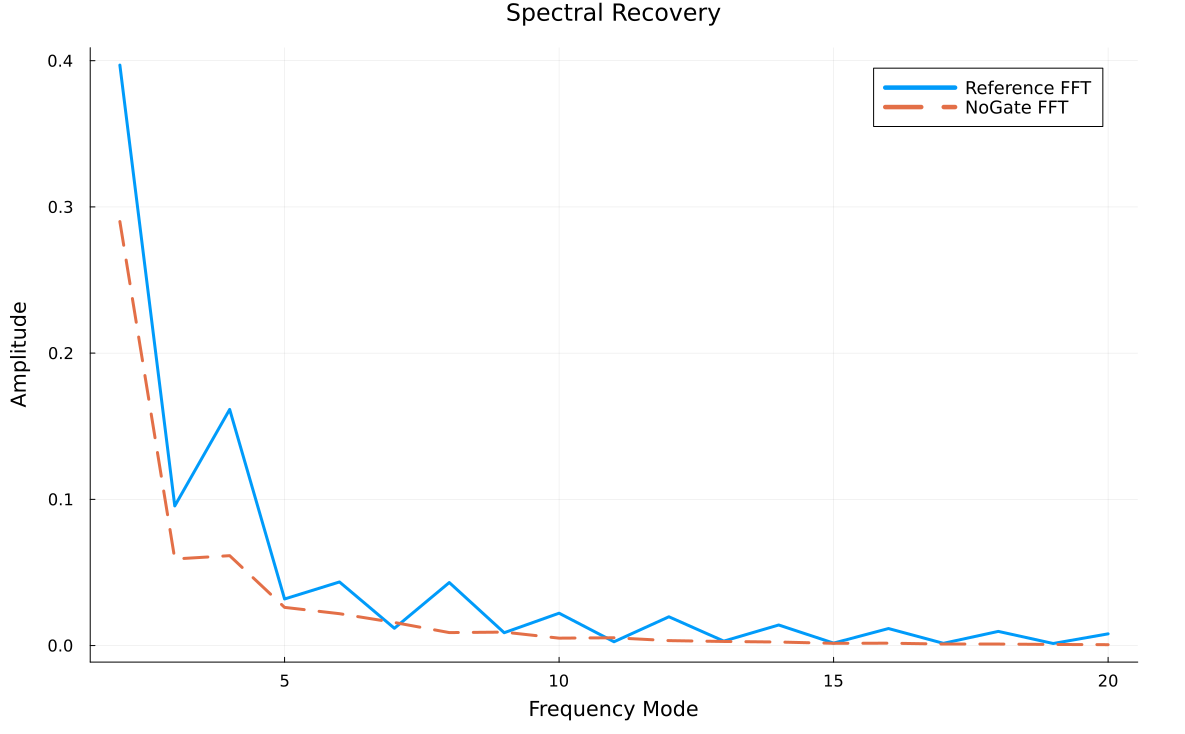}
			\caption{NoGate}
		\end{subfigure}
		\caption{1D Allen--Cahn Equation spectral recovery across ablation variants (FFT amplitude vs.\ frequency mode, single seed).}
		\label{fig:1D_Allen_Cahn_Equation_Spectral_Recovery}
	\end{figure}

	Figures~\ref{fig:1D_Allen_Cahn_Equation_Full_DBSG}--\ref{fig:1D_Allen_Cahn_Equation_Spectral_Recovery} show the same comparison for the 1D Allen--Cahn benchmark.
	
	\subsection{1D Reaction--Diffusion Equation: Qualitative Comparison Across Variants}
\begin{figure}[t]
		\centering
		\begin{subfigure}[b]{0.33\textwidth}
			\centering
			\includegraphics[width=\textwidth]{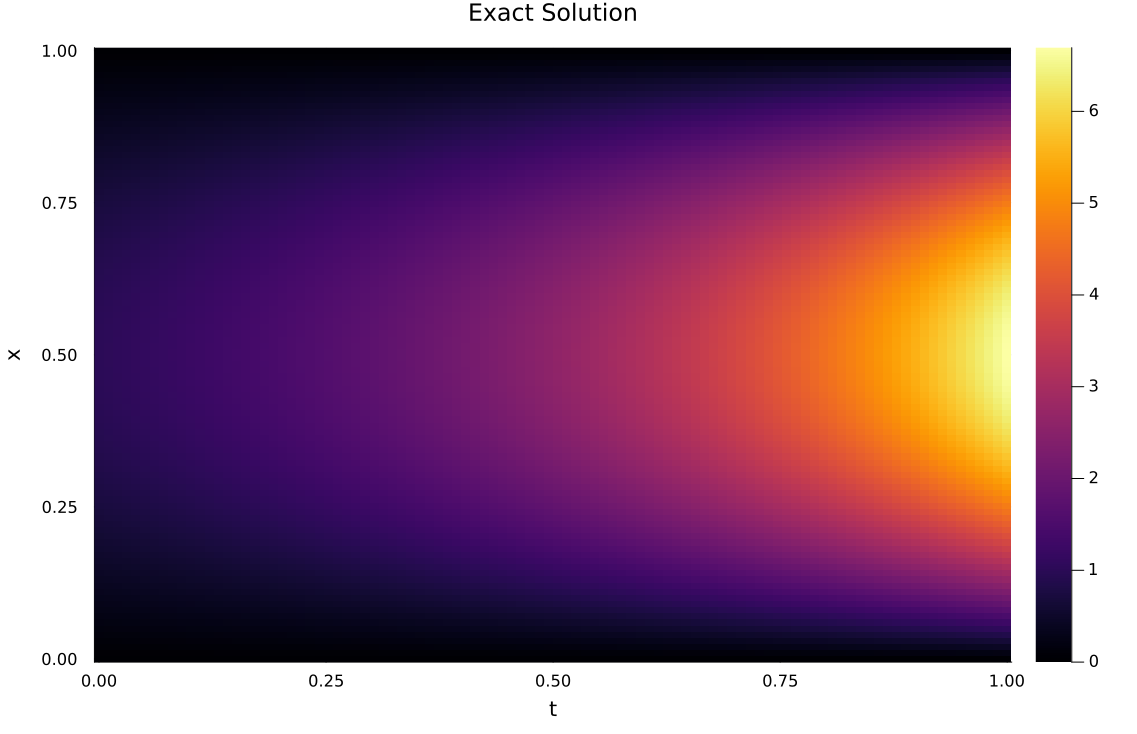}
		\end{subfigure}
		\hfill
		\begin{subfigure}[b]{0.33\textwidth}
			\centering
			\includegraphics[width=\textwidth]{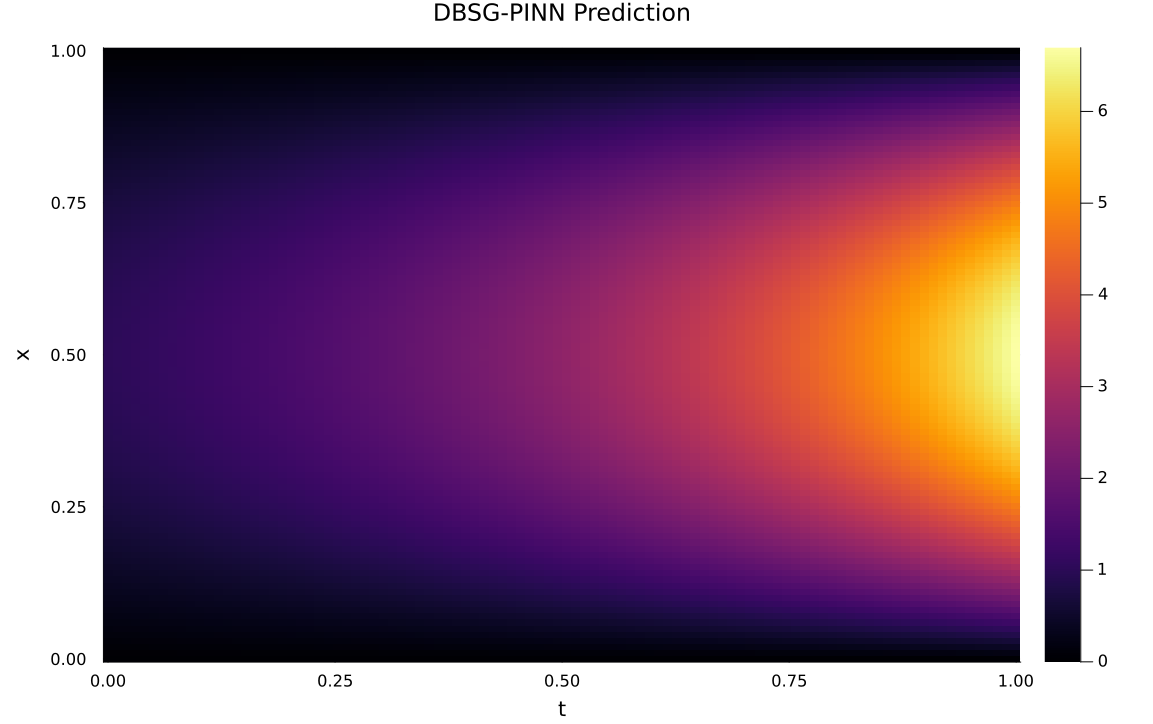}
		\end{subfigure}
		\hfill
		\begin{subfigure}[b]{0.33\textwidth}
			\centering
			\includegraphics[width=\textwidth]{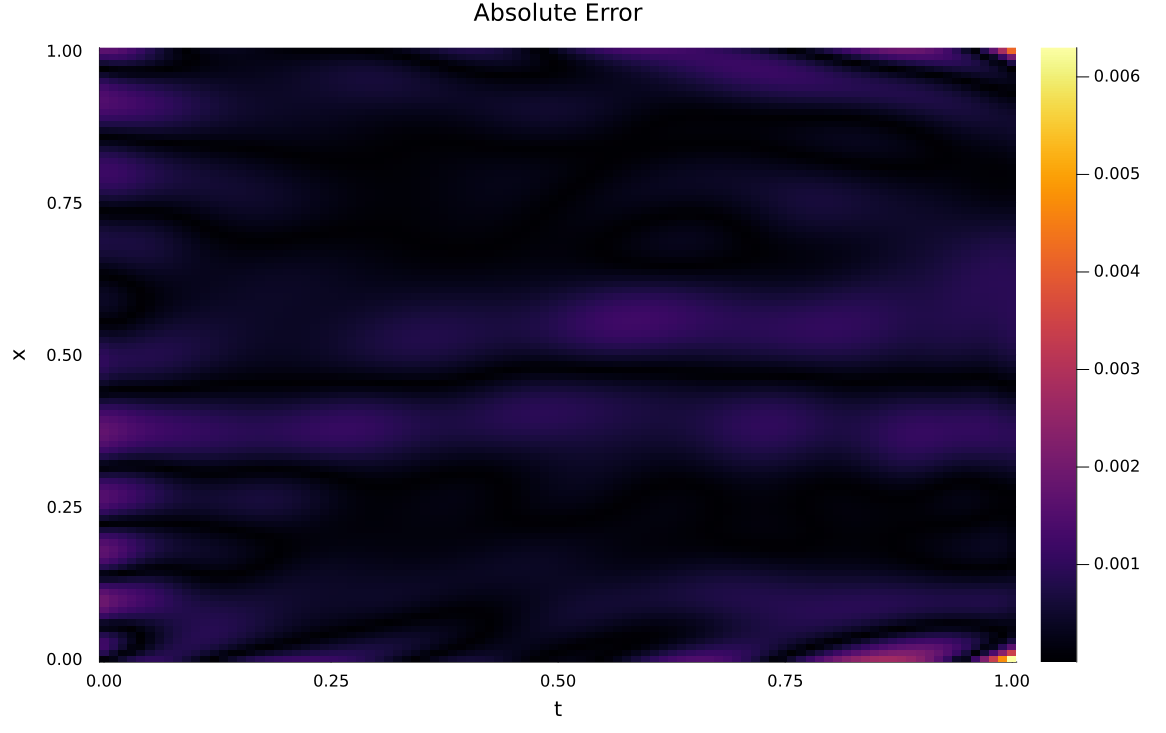}
		\end{subfigure}
		\caption{1D Reaction--Diffusion Equation, Full DBSG-PINN: exact solution, prediction, and absolute error.}
		\label{fig:1D_Reaction_Equation_Full_DBSG}
	\end{figure}

\begin{figure}[t]
		\centering
		\begin{subfigure}[b]{0.33\textwidth}
			\centering
			\includegraphics[width=\textwidth]{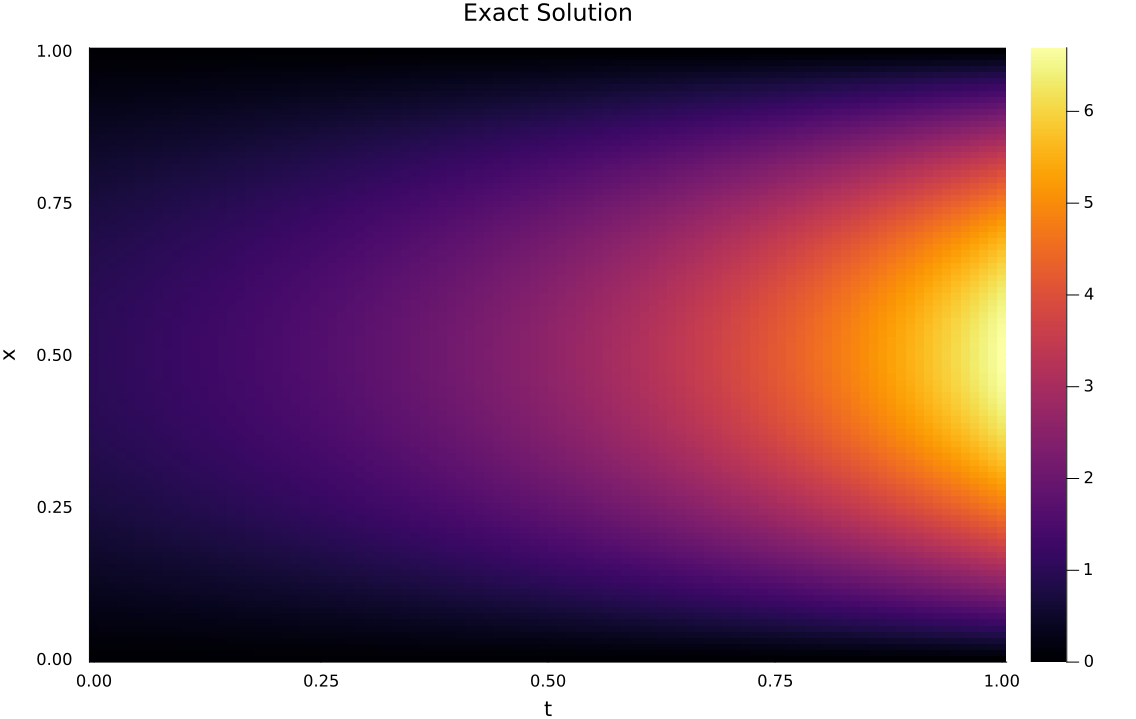}
		\end{subfigure}
		\hfill
		\begin{subfigure}[b]{0.33\textwidth}
			\centering
			\includegraphics[width=\textwidth]{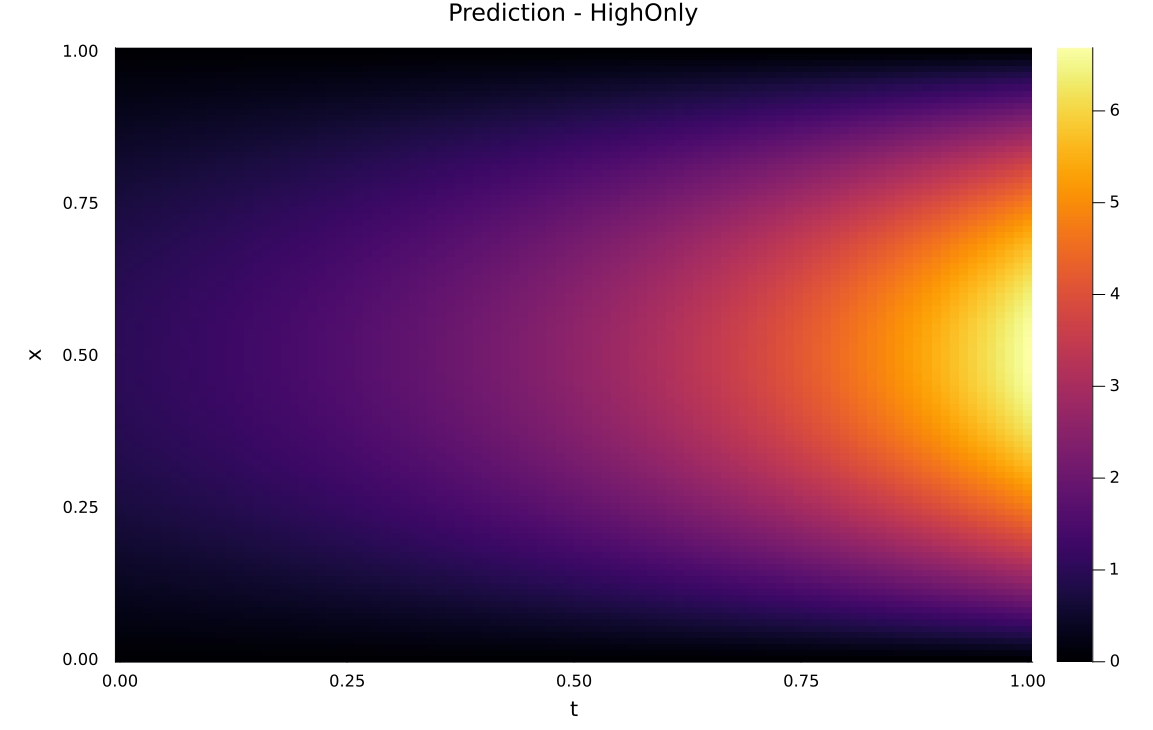}
		\end{subfigure}
		\hfill
		\begin{subfigure}[b]{0.33\textwidth}
			\centering
			\includegraphics[width=\textwidth]{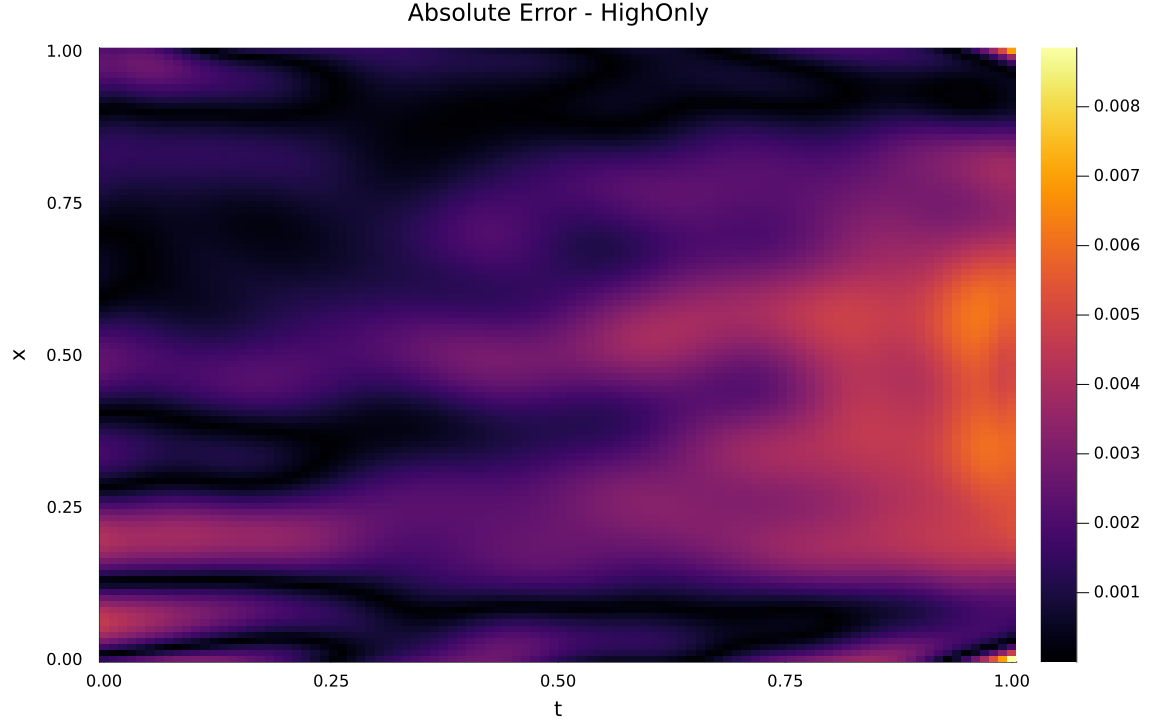}
		\end{subfigure}
		\caption{1D Reaction--Diffusion Equation, HighOnly variant.}
		\label{fig:1D_Reaction_Equation_HIGH_ONLY}
	\end{figure}

\begin{figure}[t]
		\centering
		\begin{subfigure}[b]{0.33\textwidth}
			\centering
			\includegraphics[width=\textwidth]{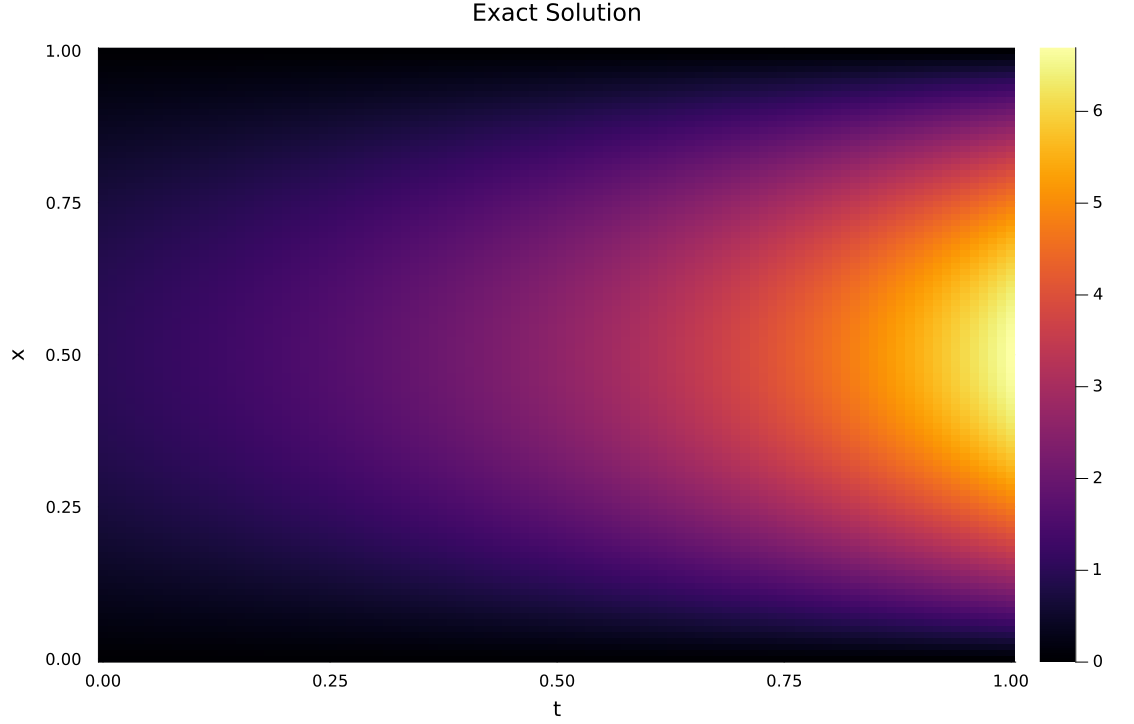}
		\end{subfigure}
		\hfill
		\begin{subfigure}[b]{0.33\textwidth}
			\centering
			\includegraphics[width=\textwidth]{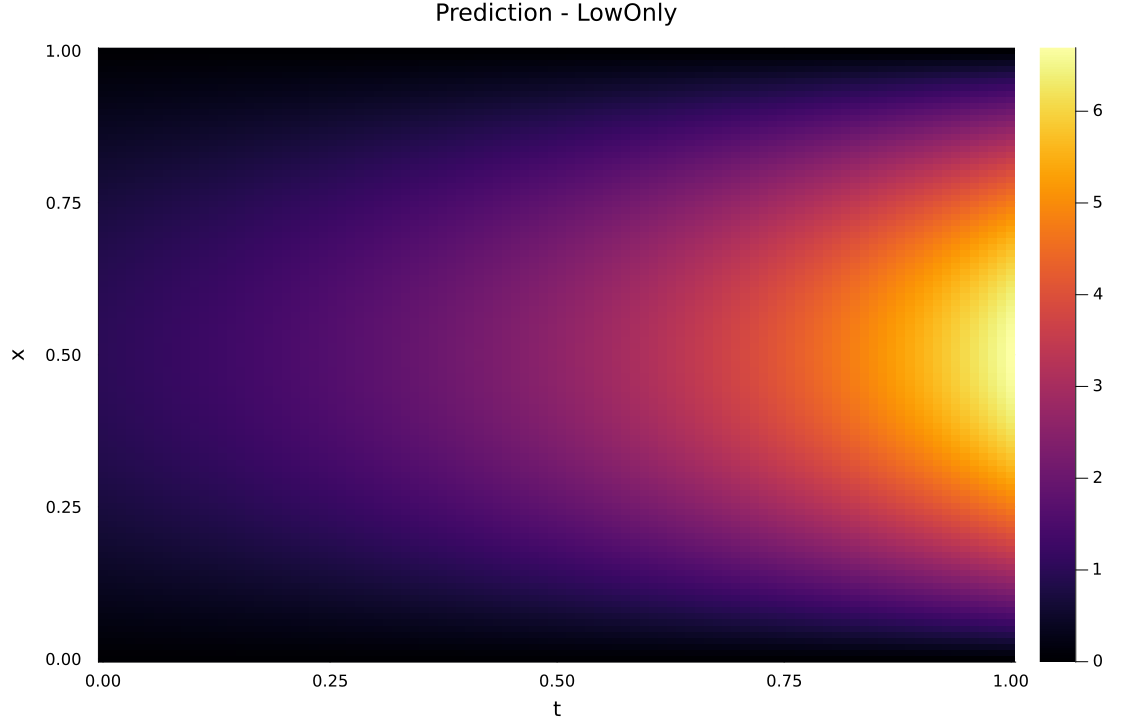}
		\end{subfigure}
		\hfill
		\begin{subfigure}[b]{0.33\textwidth}
			\centering
			\includegraphics[width=\textwidth]{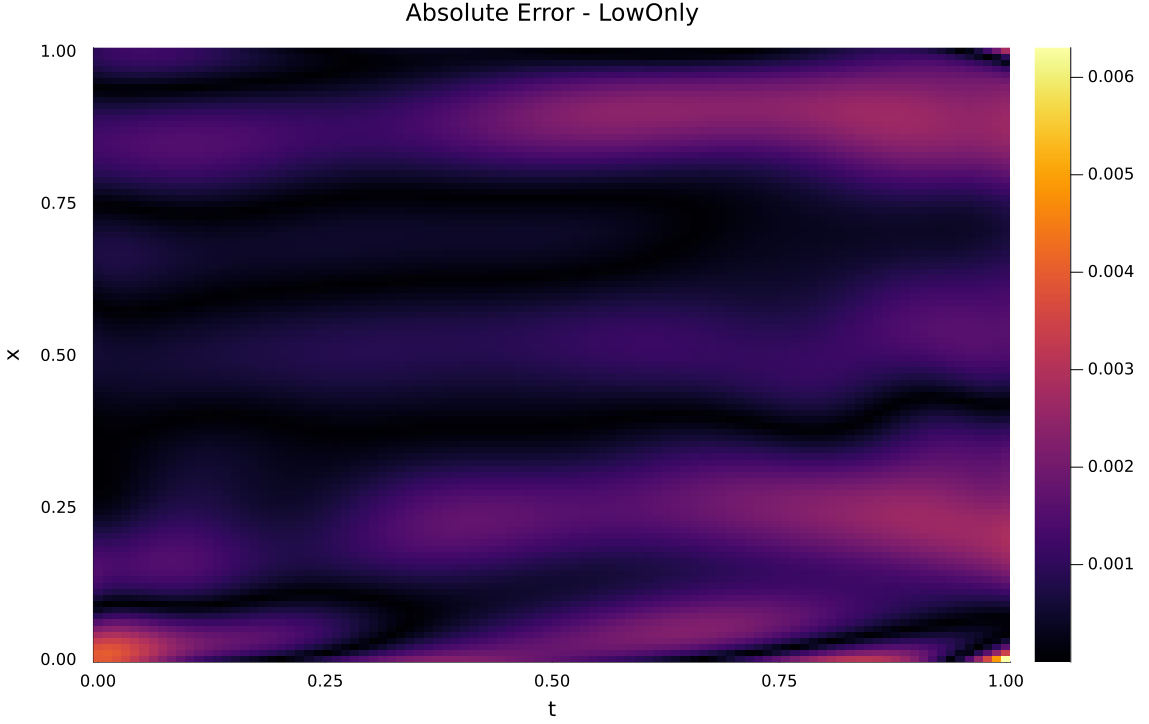}
		\end{subfigure}
		\caption{1D Reaction--Diffusion Equation, LowOnly variant.}
		\label{fig:1D_Reaction_Equation_LOW_ONLY}
	\end{figure}

\begin{figure}[t]
		\centering
		\begin{subfigure}[b]{0.33\textwidth}
			\centering
			\includegraphics[width=\textwidth]{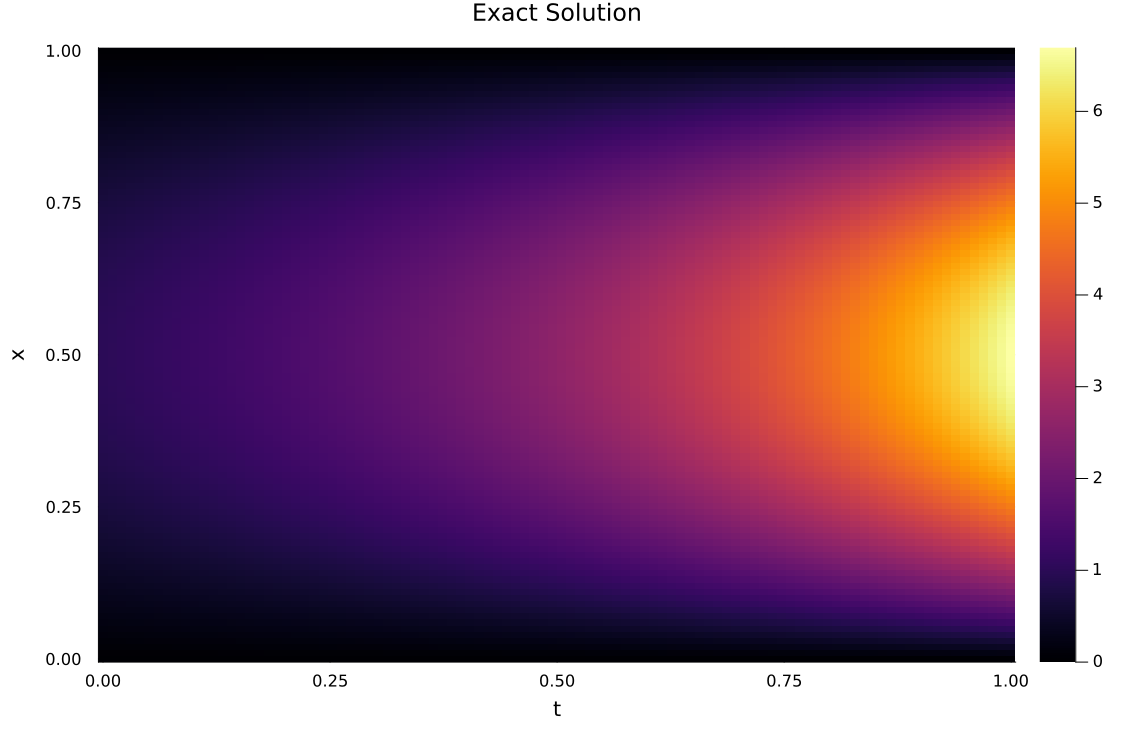}
		\end{subfigure}
		\hfill
		\begin{subfigure}[b]{0.33\textwidth}
			\centering
			\includegraphics[width=\textwidth]{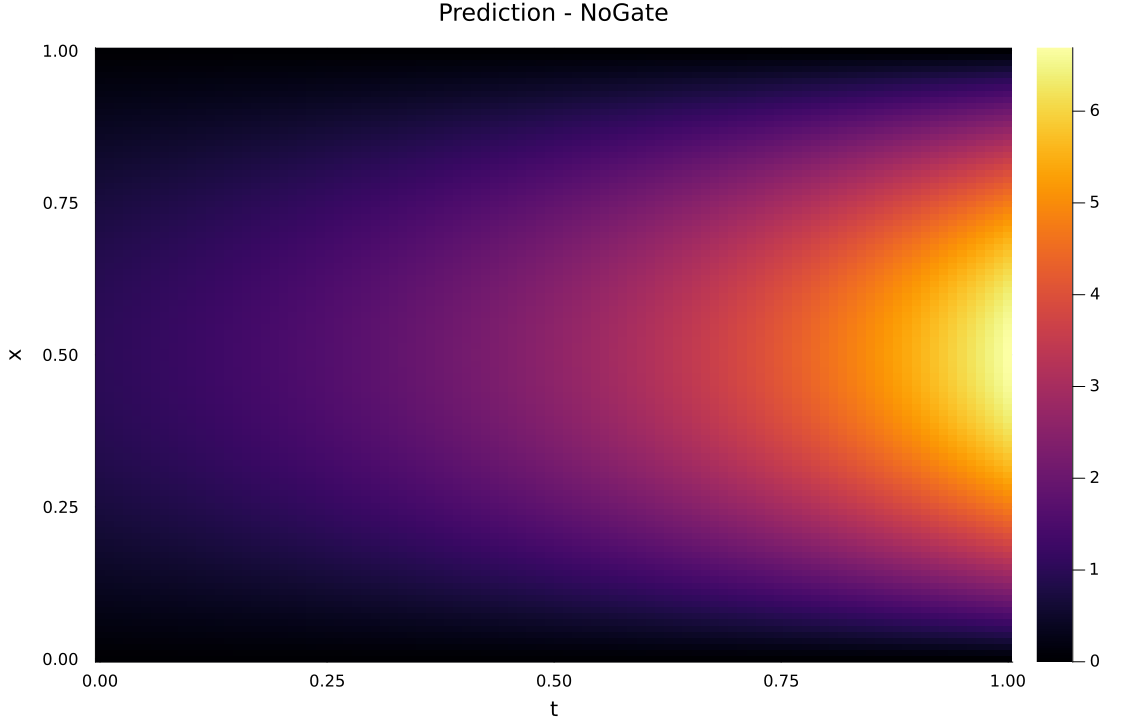}
		\end{subfigure}
		\hfill
		\begin{subfigure}[b]{0.33\textwidth}
			\centering
			\includegraphics[width=\textwidth]{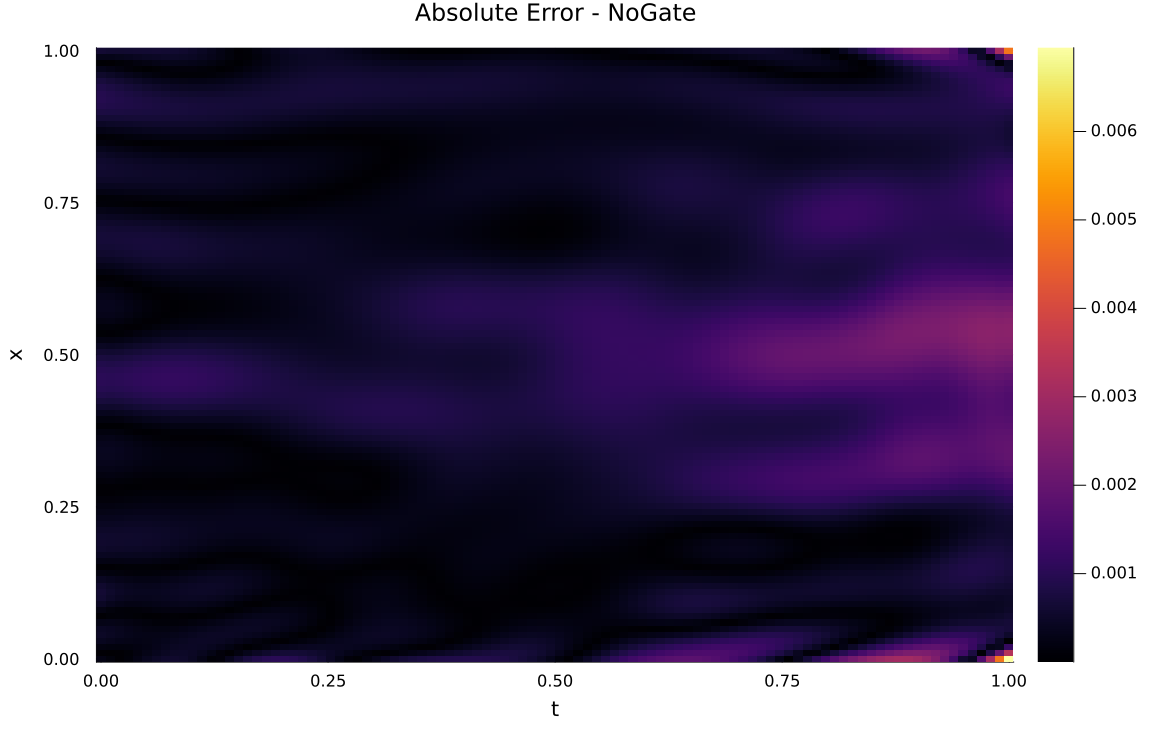}
		\end{subfigure}
		\caption{1D Reaction--Diffusion Equation, NoGate variant.}
		\label{fig:1D_Reaction_Equation_NO_gate}
	\end{figure}

\begin{figure}[t]
		\centering
		\begin{subfigure}[b]{0.48\textwidth}
			\centering
			\includegraphics[width=\textwidth]{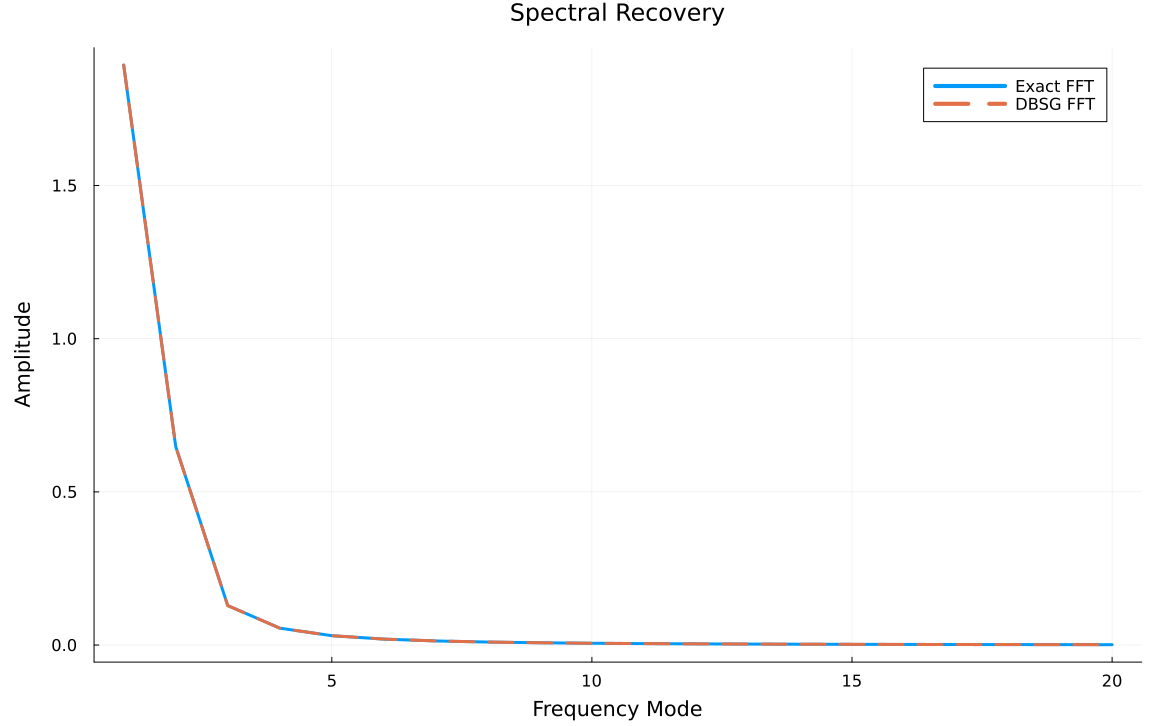}
			\caption{Full (DBSG-PINN)}
		\end{subfigure}
		\hfill
		\begin{subfigure}[b]{0.48\textwidth}
			\centering
			\includegraphics[width=\textwidth]{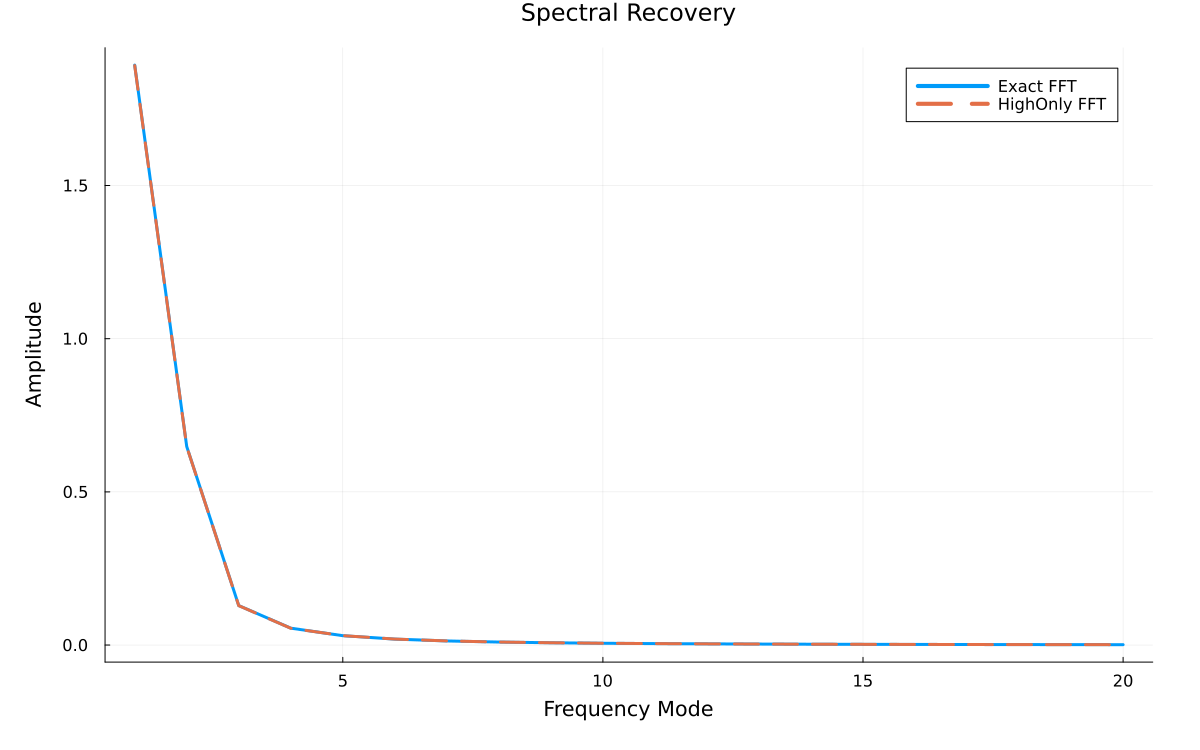}
			\caption{HighOnly}
		\end{subfigure}
		\\[1em]
		\begin{subfigure}[b]{0.48\textwidth}
			\centering
			\includegraphics[width=\textwidth]{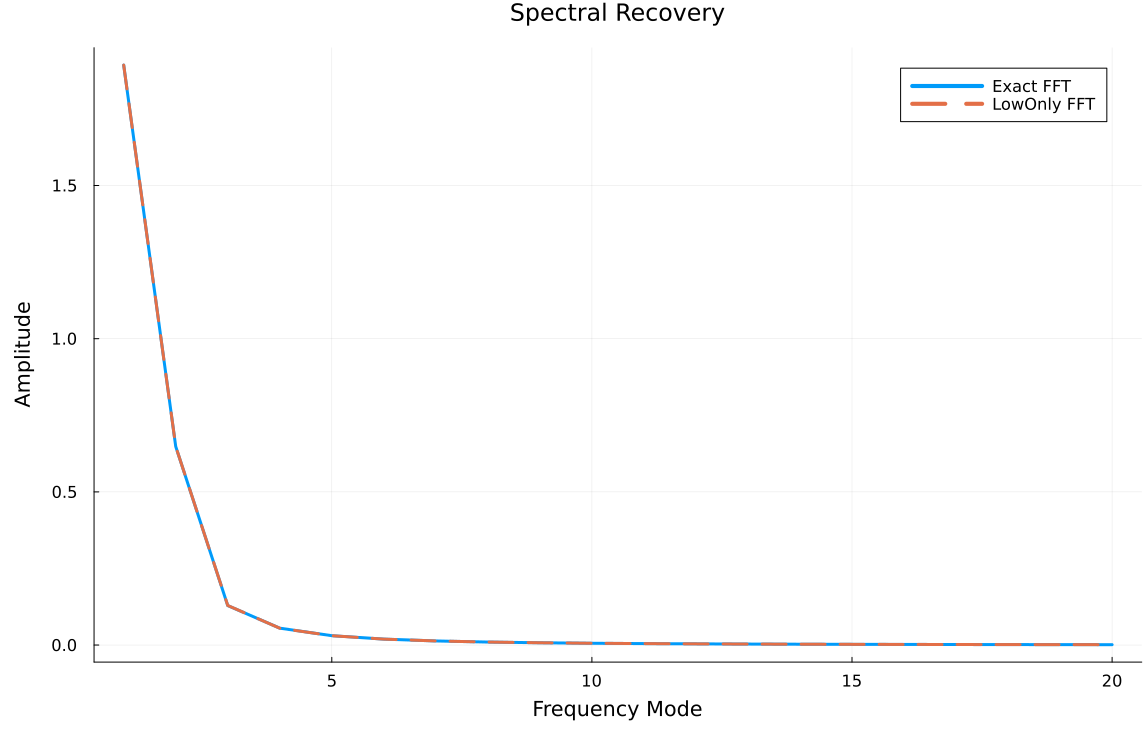}
			\caption{LowOnly}
		\end{subfigure}
		\hfill
		\begin{subfigure}[b]{0.48\textwidth}
			\centering
			\includegraphics[width=\textwidth]{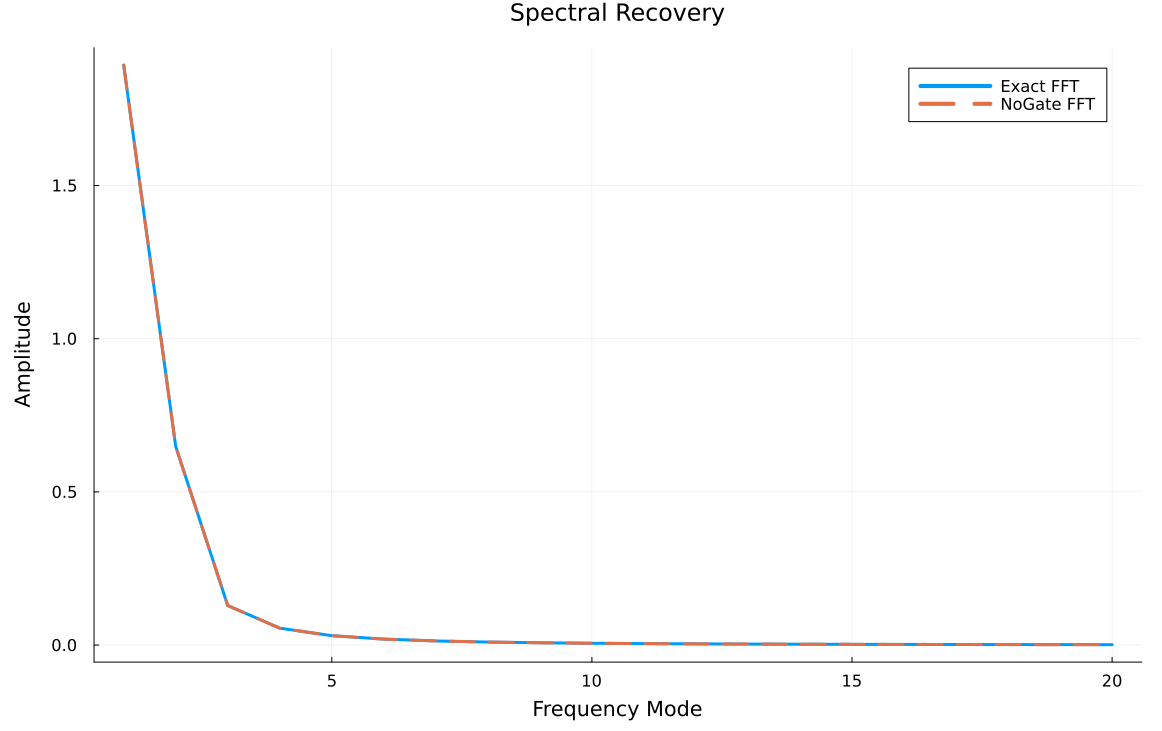}
			\caption{NoGate}
		\end{subfigure}
		\caption{1D Reaction--Diffusion Equation spectral recovery across ablation variants (FFT amplitude vs.\ frequency mode, single seed).}
		\label{fig:1D_Reaction_Equation_Spectral_Recovery}
	\end{figure}

	Figures~\ref{fig:1D_Reaction_Equation_Full_DBSG}--\ref{fig:1D_Reaction_Equation_Spectral_Recovery} show the same comparison for the 1D Reaction--Diffusion benchmark.
	
	\subsection{1D Wave Equation: Qualitative Comparison Across Variants}
\begin{figure}[t]
		\centering
		\begin{subfigure}[b]{0.33\textwidth}
			\centering
			\includegraphics[width=\textwidth]{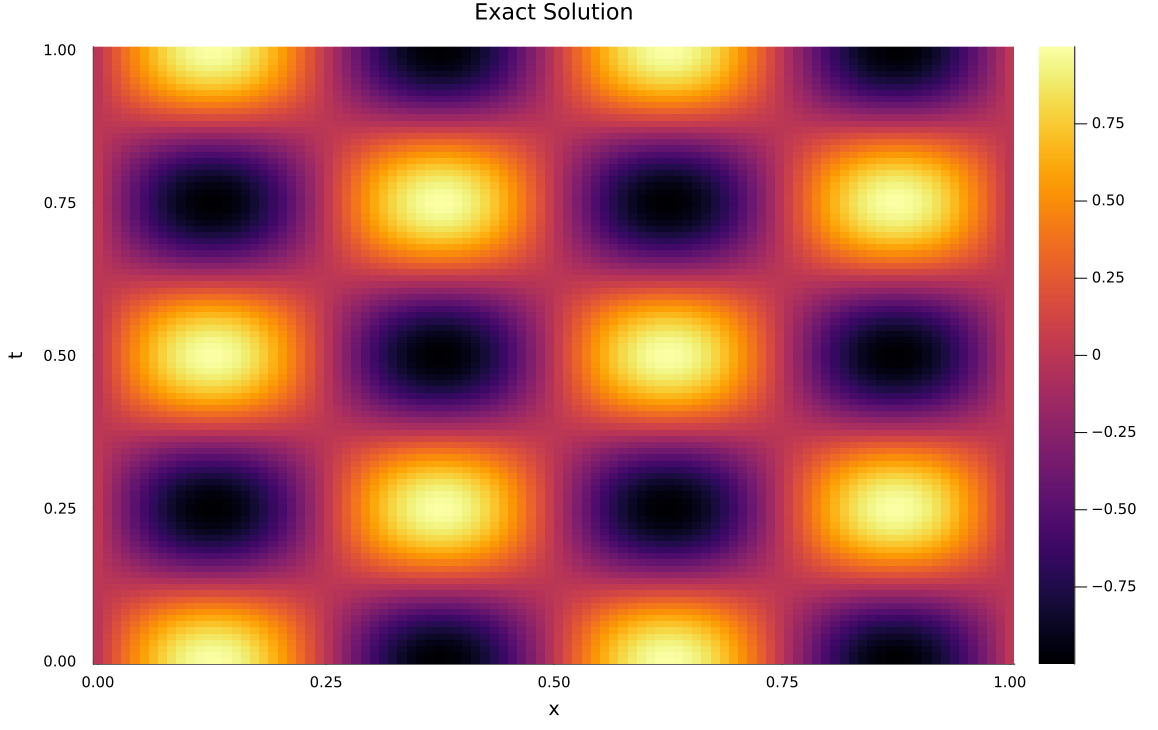}
		\end{subfigure}
		\hfill
		\begin{subfigure}[b]{0.33\textwidth}
			\centering
			\includegraphics[width=\textwidth]{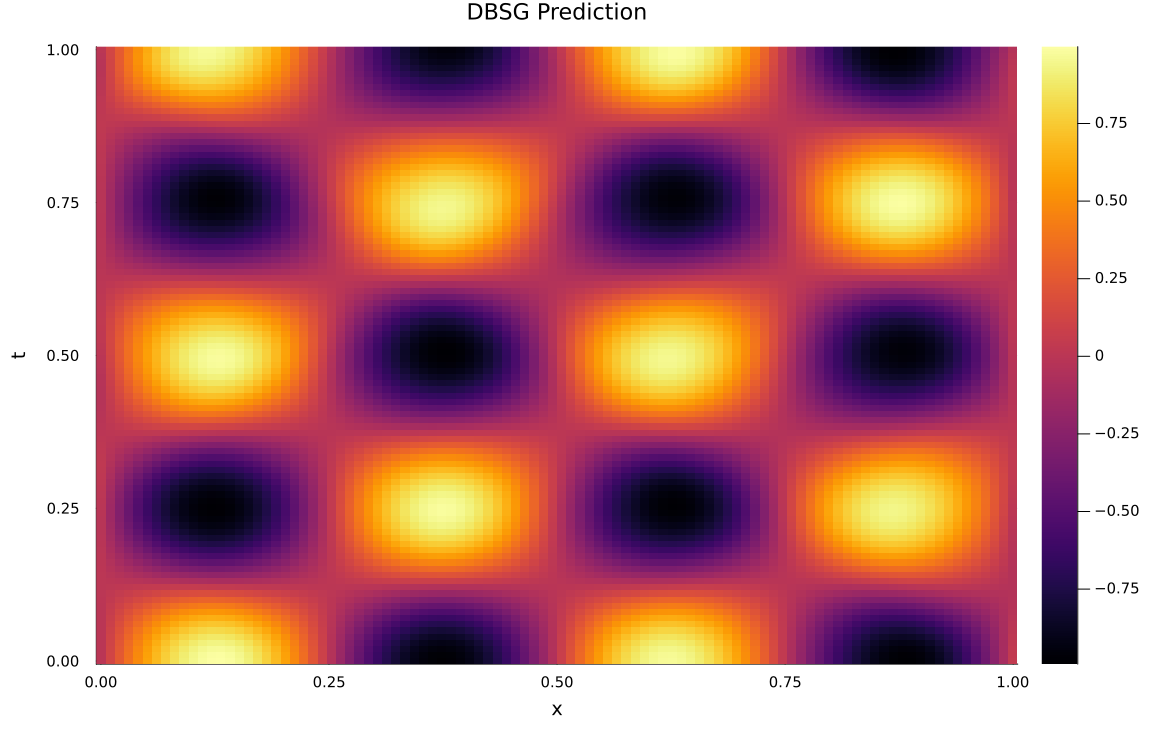}
		\end{subfigure}
		\hfill
		\begin{subfigure}[b]{0.33\textwidth}
			\centering
			\includegraphics[width=\textwidth]{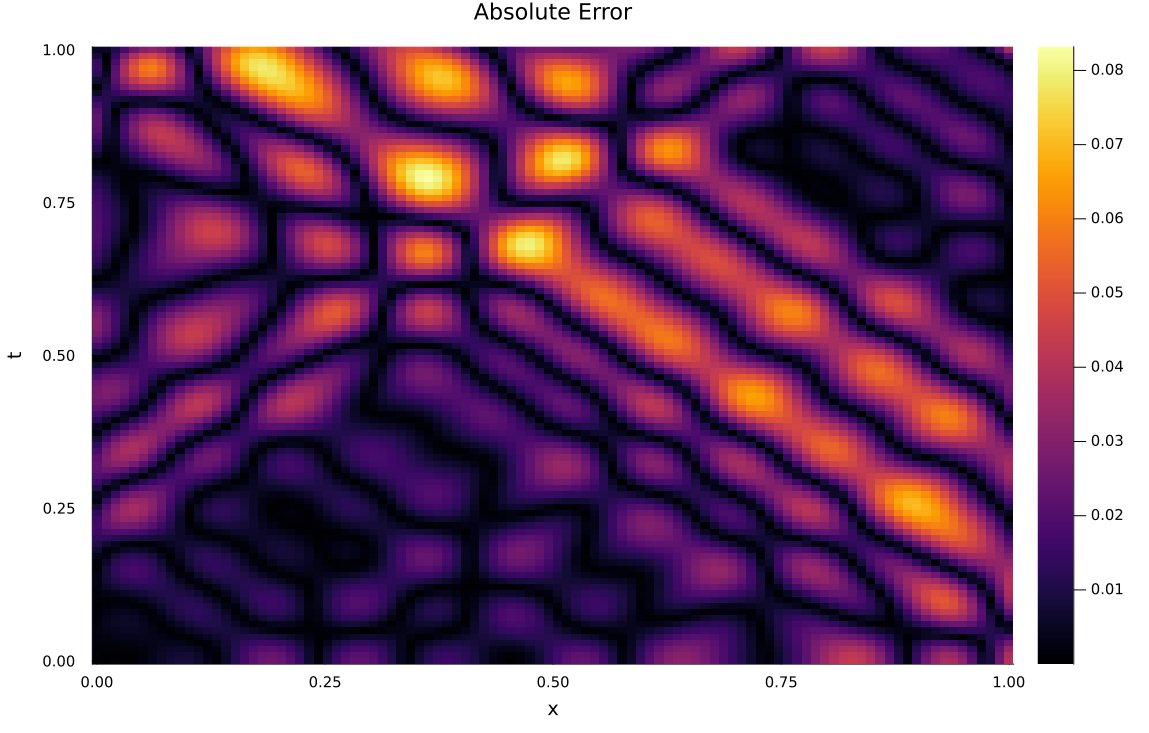}
		\end{subfigure}
		\caption{1D Wave Equation, Full DBSG-PINN: exact solution, prediction, and absolute error.}
		\label{fig:1D_Wave_Equation_Full_DBSG}
	\end{figure}

\begin{figure}[t]
		\centering
		\begin{subfigure}[b]{0.33\textwidth}
			\centering
			\includegraphics[width=\textwidth]{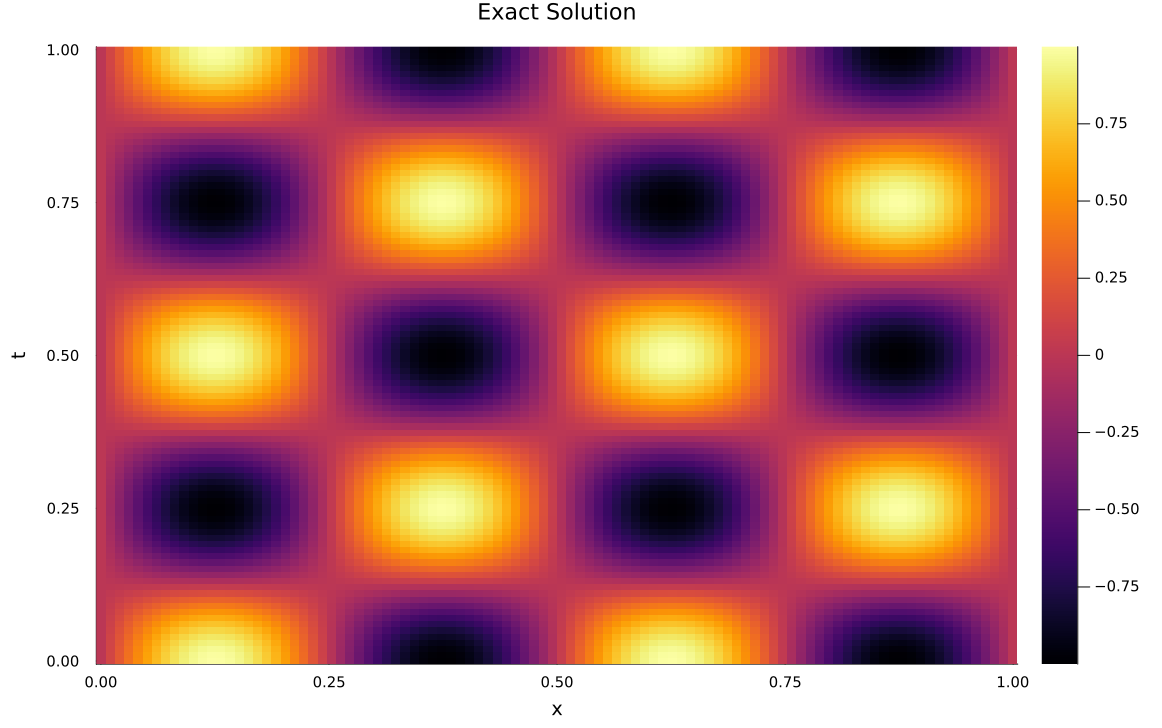}
		\end{subfigure}
		\hfill
		\begin{subfigure}[b]{0.33\textwidth}
			\centering
			\includegraphics[width=\textwidth]{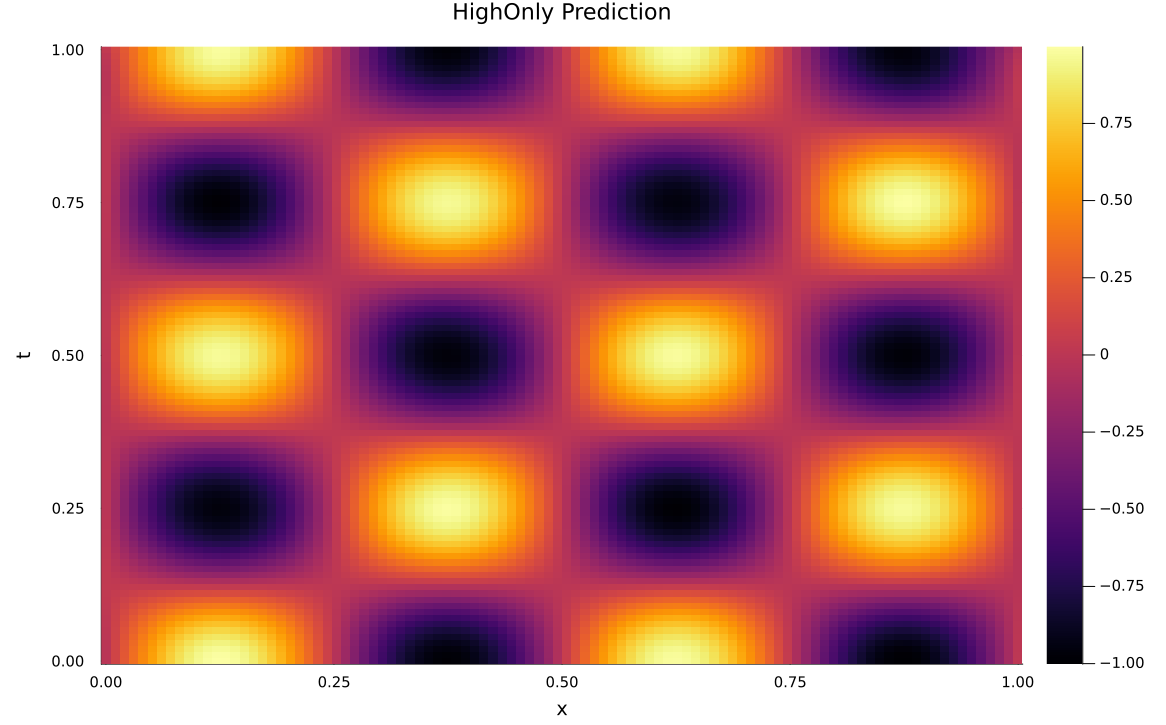}
		\end{subfigure}
		\hfill
		\begin{subfigure}[b]{0.33\textwidth}
			\centering
			\includegraphics[width=\textwidth]{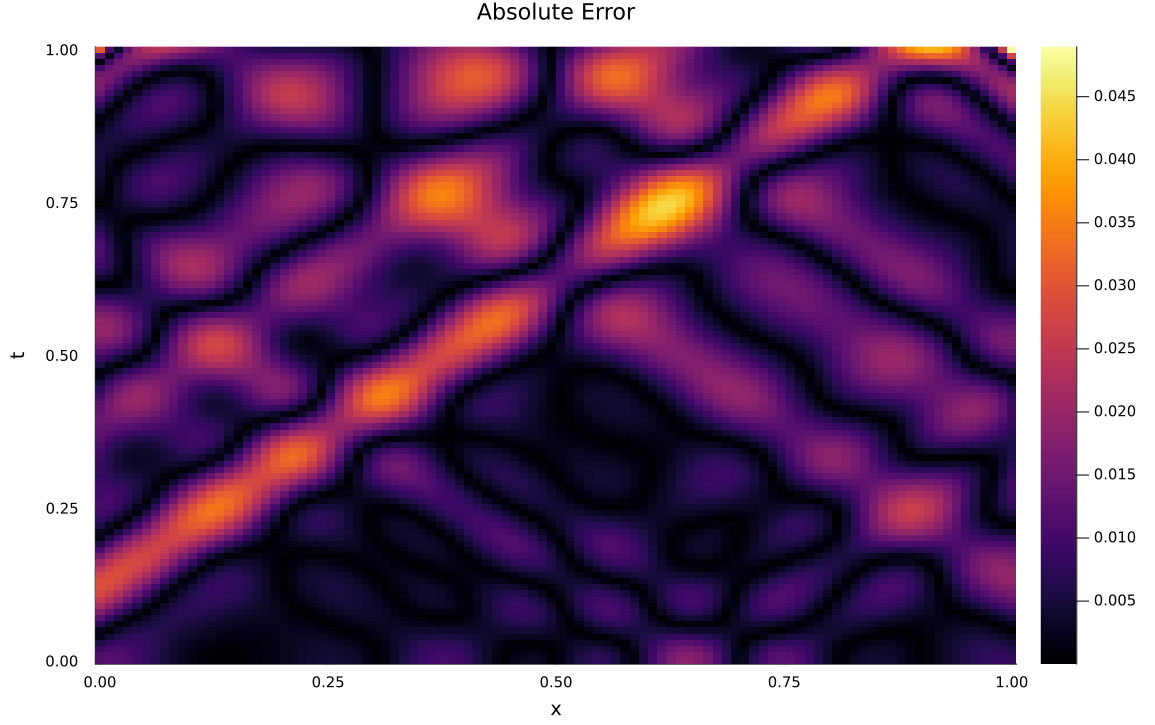}
		\end{subfigure}
		\caption{1D Wave Equation, HighOnly variant.}
		\label{fig:1D_Wave_Equation_HIGH_ONLY}
	\end{figure}

\begin{figure}[t]
		\centering
		\begin{subfigure}[b]{0.33\textwidth}
			\centering
			\includegraphics[width=\textwidth]{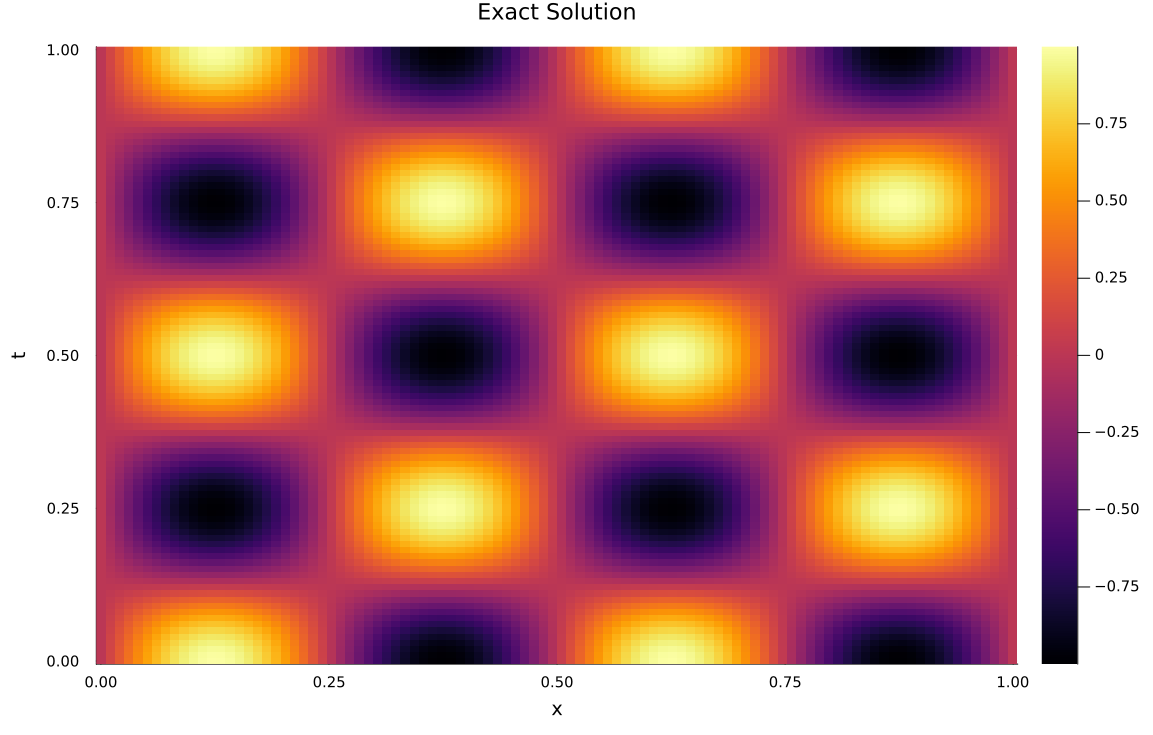}
		\end{subfigure}
		\hfill
		\begin{subfigure}[b]{0.33\textwidth}
			\centering
			\includegraphics[width=\textwidth]{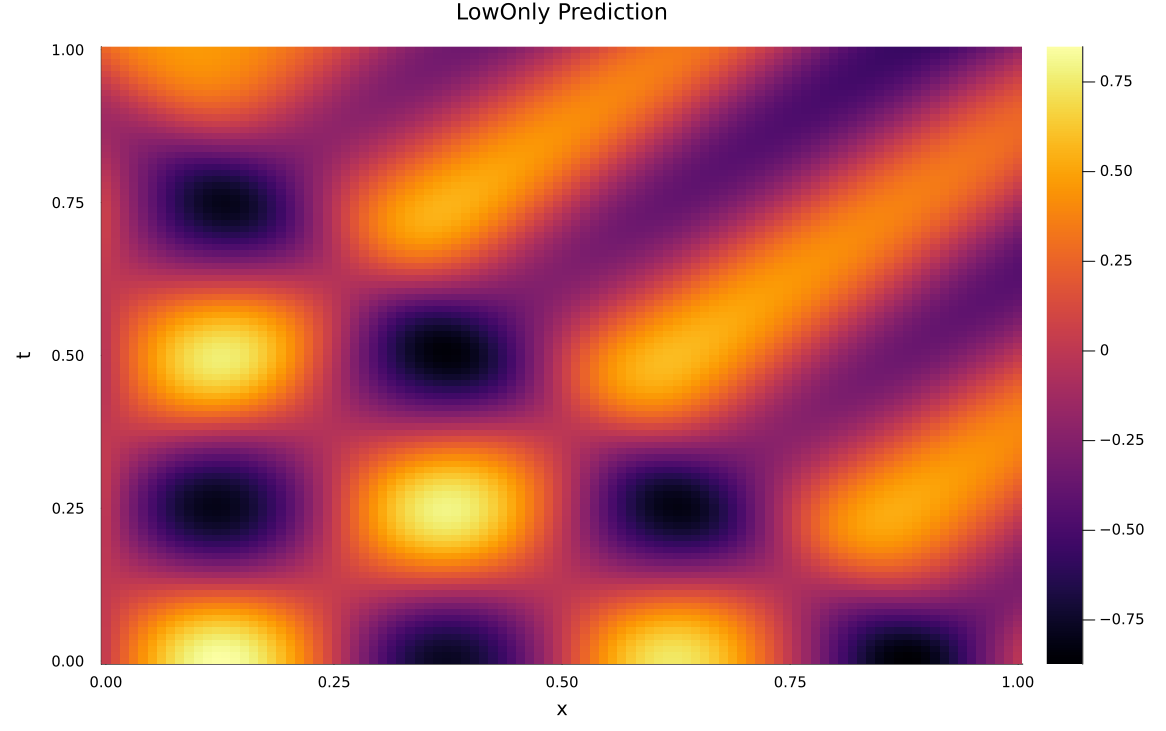}
		\end{subfigure}
		\hfill
		\begin{subfigure}[b]{0.33\textwidth}
			\centering
			\includegraphics[width=\textwidth]{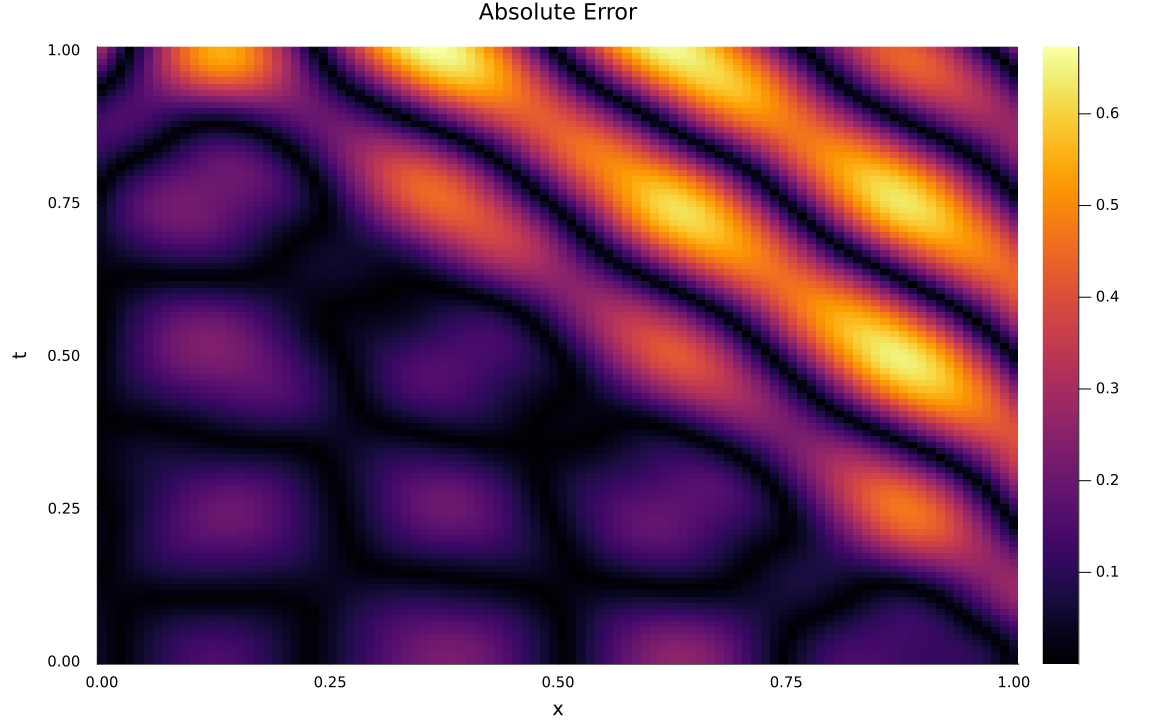}
		\end{subfigure}
		\caption{1D Wave Equation, LowOnly variant.}
		\label{fig:1D_Wave_Equation_LOW_ONLY}
	\end{figure}

\begin{figure}[t]
		\centering
		\begin{subfigure}[b]{0.33\textwidth}
			\centering
			\includegraphics[width=\textwidth]{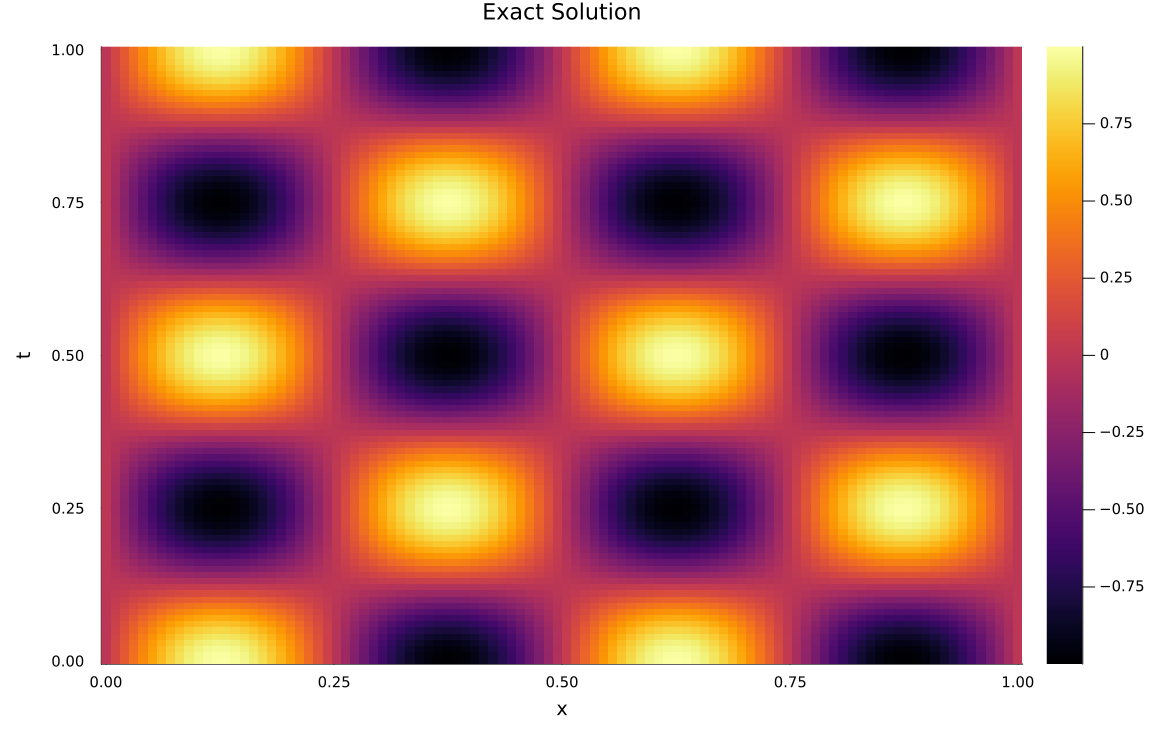}
		\end{subfigure}
		\hfill
		\begin{subfigure}[b]{0.33\textwidth}
			\centering
			\includegraphics[width=\textwidth]{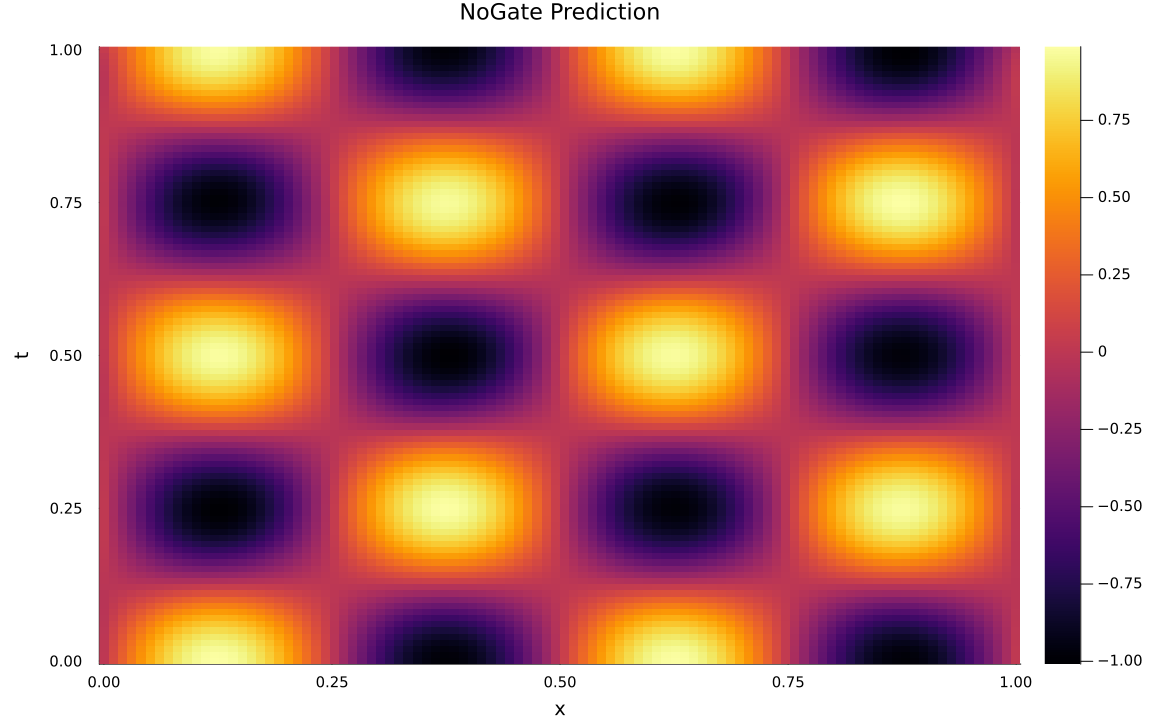}
		\end{subfigure}
		\hfill
		\begin{subfigure}[b]{0.33\textwidth}
			\centering
			\includegraphics[width=\textwidth]{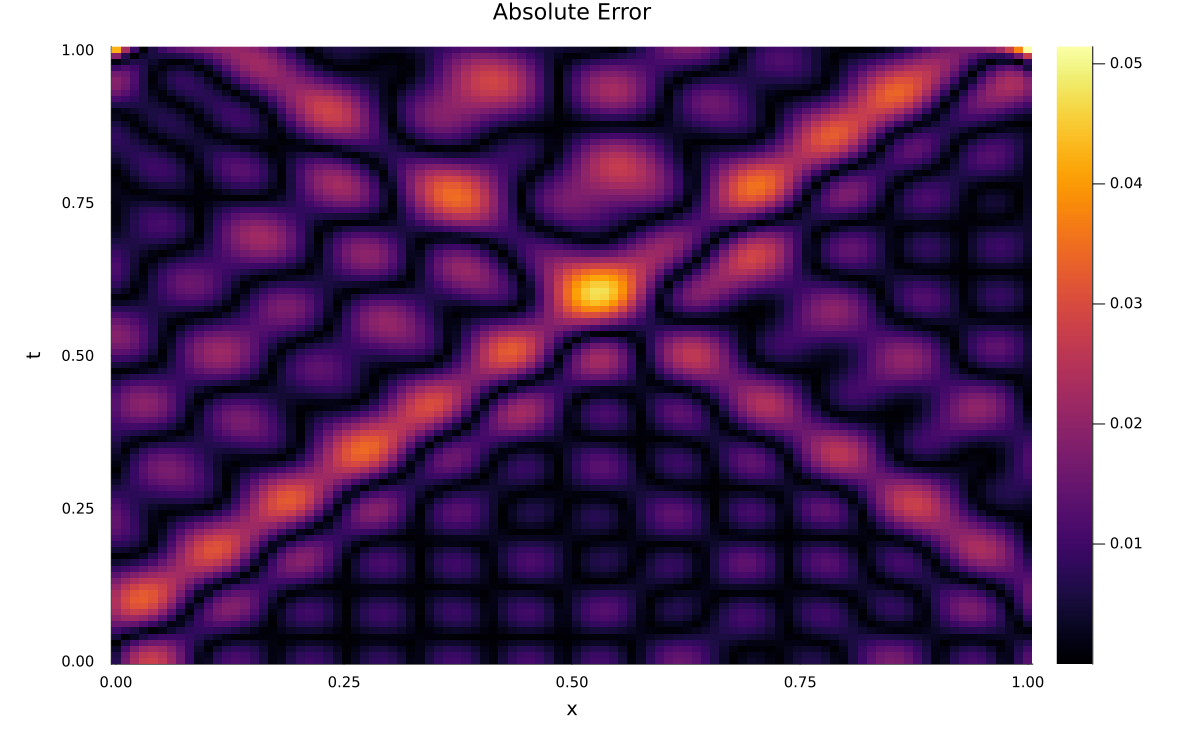}
		\end{subfigure}
		\caption{1D Wave Equation, NoGate variant.}
		\label{fig:1D_Wave_Equation_NO_gate}
	\end{figure}

\begin{figure}[t]
		\centering
		\begin{subfigure}[b]{0.48\textwidth}
			\centering
			\includegraphics[width=\textwidth]{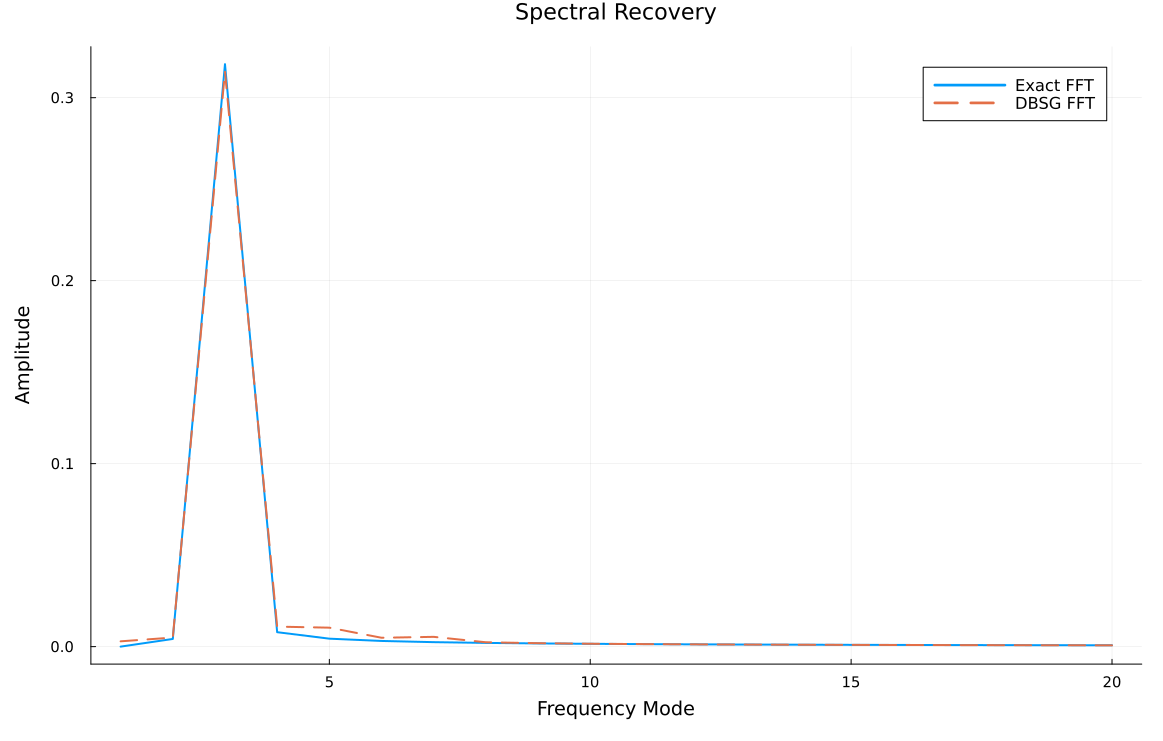}
			\caption{Full (DBSG-PINN)}
		\end{subfigure}
		\hfill
		\begin{subfigure}[b]{0.48\textwidth}
			\centering
			\includegraphics[width=\textwidth]{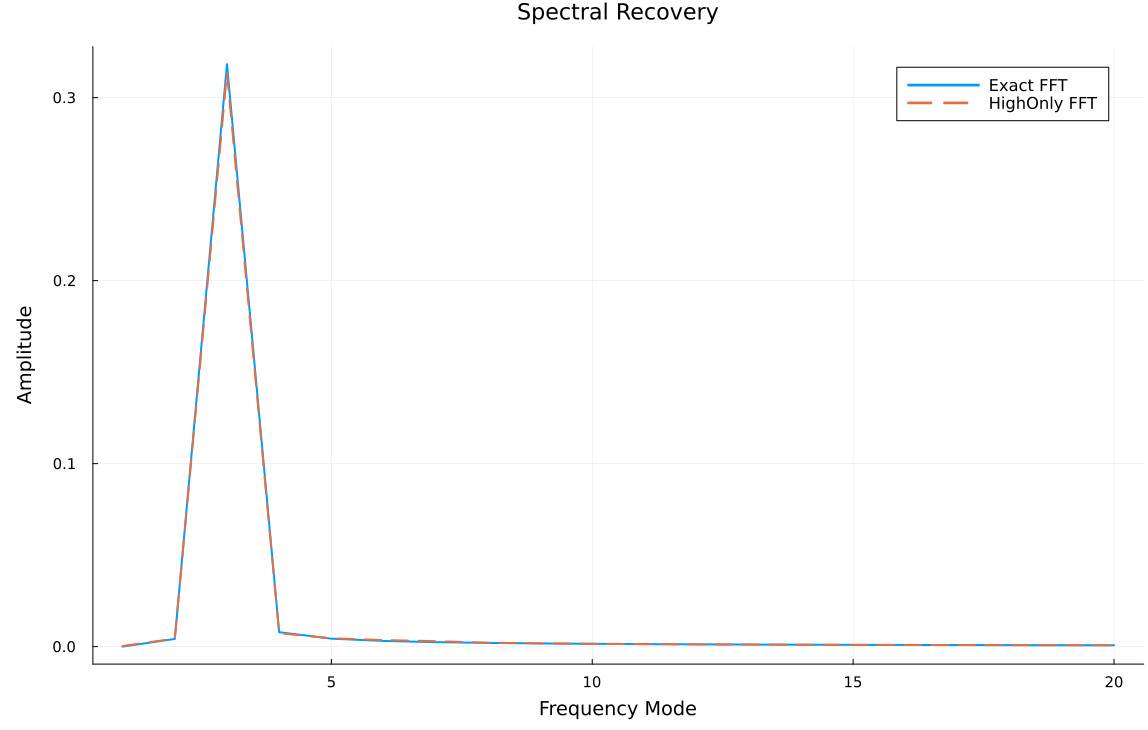}
			\caption{HighOnly}
		\end{subfigure}
		\\[1em]
		\begin{subfigure}[b]{0.48\textwidth}
			\centering
			\includegraphics[width=\textwidth]{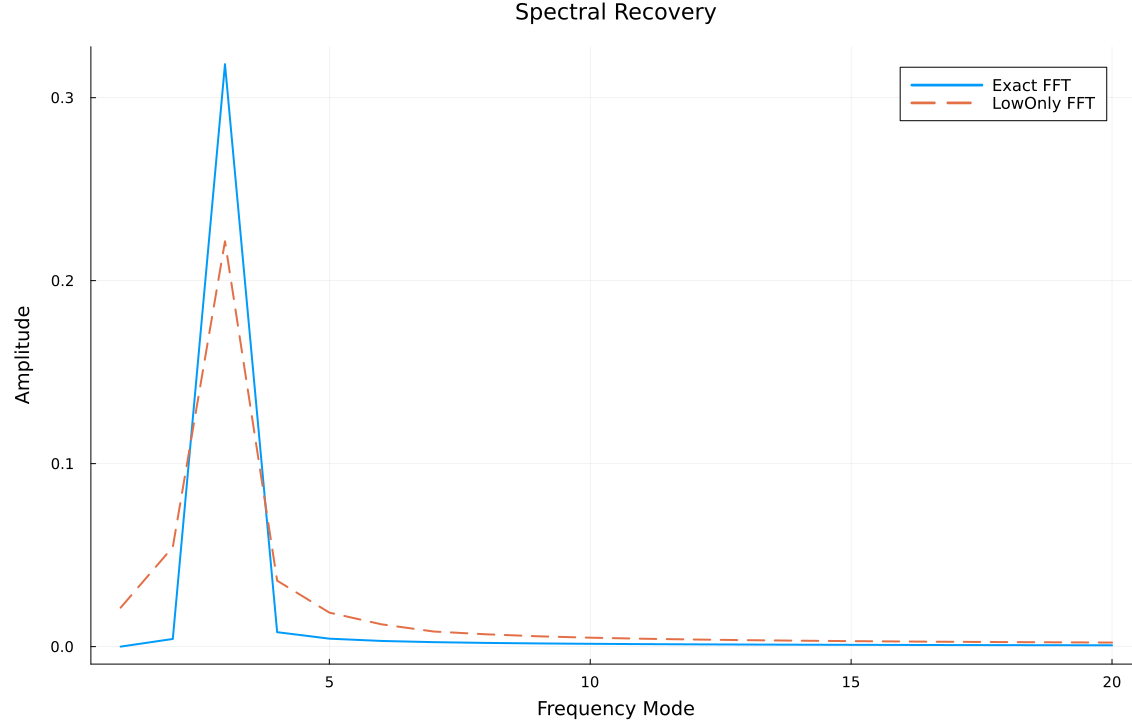}
			\caption{LowOnly}
		\end{subfigure}
		\hfill
		\begin{subfigure}[b]{0.48\textwidth}
			\centering
			\includegraphics[width=\textwidth]{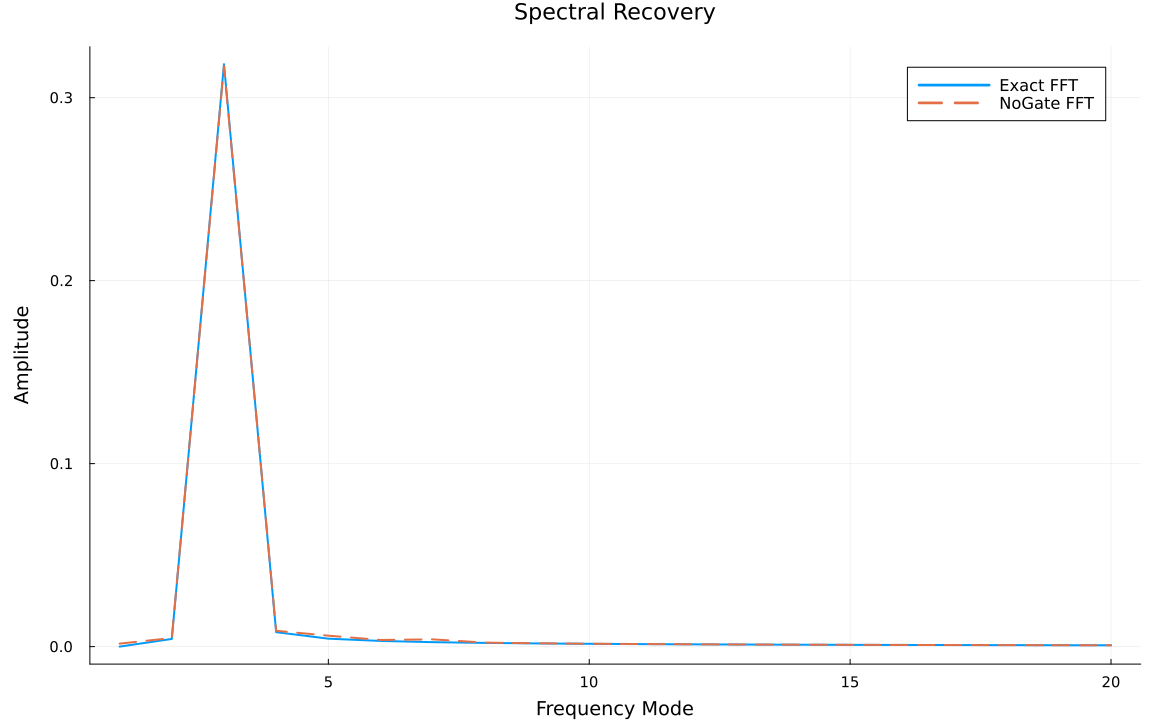}
			\caption{NoGate}
		\end{subfigure}
		\caption{1D Wave Equation spectral recovery across ablation variants (FFT amplitude vs.\ frequency mode, single seed).}
		\label{fig:1D_Wave_Equation_Spectral_Recovery}
	\end{figure}

	Figures~\ref{fig:1D_Wave_Equation_Full_DBSG}--\ref{fig:1D_Wave_Equation_Spectral_Recovery} show the same comparison for the 1D Wave benchmark.
\end{document}